\documentclass{article} 
\usepackage{iclr2023_conference,times}

\usepackage{amsmath,amsfonts,bm}

\def\eqref#1{equation~\ref{#1}}

\def\1{\bm{1}}

\DeclareMathAlphabet{\mathsfit}{\encodingdefault}{\sfdefault}{m}{sl}
\SetMathAlphabet{\mathsfit}{bold}{\encodingdefault}{\sfdefault}{bx}{n}

\usepackage{hyperref}
\usepackage{url}

\usepackage{booktabs}
\usepackage{multirow}
\usepackage{graphicx}
\usepackage{subcaption}
\usepackage{amsmath}
\usepackage{amssymb}
\usepackage{algorithm}
\usepackage{algorithmic}
\usepackage{amsthm}
\usepackage{wrapfig}  
\usepackage{xcolor}
\newtheorem{proposition}{Proposition}
\title{Learning Cut Selection for Mixed-Integer Linear Programming via Hierarchical \\ Sequence Model}

\author{
Zhihai Wang\thanks{Equal contribution.}$\,\,$\textsuperscript{1} ,
Xijun Li\footnotemark[1]$\,\,$\textsuperscript{1,2}, 
Jie Wang\thanks{Corresponding author: jiewangx@ustc.edu.cn}$\,\,$\textsuperscript{1,3} ,
Yufei Kuang\textsuperscript{1},
Mingxuan Yuan\textsuperscript{2}, \\
\textbf{
Jia Zeng\textsuperscript{2},
Yongdong Zhang\textsuperscript{1,3},
Feng Wu\textsuperscript{1,3}
}\\
\textsuperscript{1} CAS Key Laboratory of Technology in GIPAS, 
University of Science and Technology of China, \\
\textsuperscript{2} Noah’s Ark Lab, Huawei Technologies, \\
\textsuperscript{3} Institute of Artificial Intelligence, Hefei Comprehensive National Science Center
}

\iclrfinalcopy 
\begin{document}

\maketitle

\begin{abstract}
    Cutting planes (cuts) are important for solving mixed-integer linear programs (MILPs), which formulate a wide range of important real-world applications. Cut selection---which aims to select a proper subset of the candidate cuts to improve the efficiency of solving MILPs---heavily depends on \textbf{(P1)} which cuts should be preferred, and \textbf{(P2)} how many cuts should be selected. Although many modern MILP solvers tackle \textbf{(P1)-(P2)} by manually designed heuristics, machine learning offers a promising approach to learn more effective heuristics from MILPs collected from specific applications. However, many existing learning-based methods focus on learning which cuts should be preferred, neglecting the importance of learning the number of cuts that should be selected. Moreover, we observe from extensive empirical results that \textbf{(P3)} what \textit{order of selected cuts} should be preferred has a significant impact on the efficiency of solving MILPs as well. To address this challenge, we propose a novel \textbf{h}ierarchical s\textbf{e}quence \textbf{m}odel (HEM) to learn cut selection policies via reinforcement learning. Specifically, HEM consists of a two-level model: (1) a higher-level model to learn the number of cuts that should be selected, (2) and a lower-level model---that formulates the cut selection task as a sequence to sequence learning problem---to learn policies selecting an \textit{ordered subset} with the size determined by the higher-level model. To the best of our knowledge, HEM is \textit{the first} method that can tackle \textbf{(P1)-(P3)} in cut selection simultaneously from a data-driven perspective. Experiments show that HEM significantly improves the efficiency of solving MILPs compared to human-designed and learning-based baselines on both synthetic and large-scale real-world MILPs, including MIPLIB 2017. Moreover, experiments demonstrate that HEM well generalizes to MILPs that are significantly larger than those seen during training.
\end{abstract}
\vspace{-2mm}
\section{Introduction}\label{sec:introduction}
\vspace{-2mm}
Mixed-integer linear programming (MILP) is a general optimization formulation for a wide range of important real-world applications, such as supply chain management \citep{supply_chain_management}, production planning \citep{production_planning}, scheduling \citep{production_scheduling}, 
facility location \citep{facility_location}, 
bin packing \citep{milp_google},
etc.
A standard MILP takes the form of
    \begin{align}
        z^* \triangleq \min_{\textbf{x}} \{ \textbf{c}^{\top} \textbf{x} | \textbf{A}\textbf{x} \leq \textbf{b},\textbf{x}\in \mathbb{R}^n, x_j\in \mathbb{Z} \,\, \text{for all } j \in \mathcal{I}  \}, \label{milp1}
    \end{align}
where $\textbf{c} \in \mathbb{R}^n$, $\textbf{A}\in \mathbb{R}^{m\times n}$, $\textbf{b} \in \mathbb{R}^{m}$, $x_j$ denotes the $j$-th entry of vector $\textbf{x}$, $\mathcal{I} \subseteq \{1,\dots,n\}$ denotes 
the set of indices of integer variables, and $z^*$ denotes the optimal objective value of the problem in (\ref{milp1}).
However, MILPs can be extremely hard to solve as they are $\mathcal{NP}$-hard problems \citep{mip_np_hard}.
To solve MILPs, many modern MILP solvers \citep{gurobi, scip8, xpress} employ a branch-and-bound tree search algorithm \citep{branch_and_bound}, in which a linear programming (LP) relaxation of a MILP (the problem in (\ref{milp1}) or its subproblems) is solved at each node.
To further enhance the performance of the tree search algorithm, cutting planes (cuts) \citep{gomory_cuts} are introduced to tighten the LP relaxations \citep{scip_thesis, bengio_ml4co}.
Existing work on cuts falls into two categories: cut generation and cut selection \citep{adaptive_cut_selection}. Cut generation aims to 
generate cuts, i.e., valid linear inequalities that tighten the LP relaxations \citep{scip_thesis}. However, adding all the generated cuts to the LP relaxations can pose a computational problem \citep{implementing_cutting}. To further improve the efficiency of solving MILPs, cut selection is proposed to select a proper subset of the generated cuts \citep{implementing_cutting}.
In this paper, we focus on \textbf{the cut selection problem}, which has a significant impact on the overall solver performance \citep{scip_thesis, tang_icml20, l2c_lookahead}.

Cut selection heavily depends on \textbf{(P1)} which cuts should be preferred, and \textbf{(P2)} how many cuts should be selected \citep{scip_thesis, theoretical_cuts}. Many modern MILP solvers \citep{gurobi, scip8,xpress} tackle \textbf{(P1)-}\textbf{(P2)} by hard-coded heuristics designed by experts. However, hard-coded heuristics do not take into account underlying patterns among  MILPs collected from certain types of real-world applications, e.g., day-to-day production planning, bin packing, and vehicle routing problems \citep{pochet2006production, vehicle_routing, milp_google}.
To further improve the efficiency of MILP solvers, recent methods \citep{tang_icml20, l2c_lookahead, cut_ranking} propose to learn cut selection policies via machine learning, especially reinforcement learning. They offer promising approaches to learn more effective heuristics by capturing underlying patterns among MILPs from specific applications \citep{bengio_ml4co}.  
However, many existing learning-based methods \citep{tang_icml20, l2c_lookahead, cut_ranking}---which learn a scoring function to measure cut quality and select a fixed ratio/number of cuts with high scores---suffer from two limitations. First, they learn which cuts should be preferred by learning a scoring function, neglecting the importance of learning the number of cuts that should be selected \citep{theoretical_cuts}.
Moreover, we observe from extensive empirical results that \textbf{(P3)} what \textit{order of selected cuts} should be preferred significantly impacts the efficiency of solving MILPs as well (see Section \ref{sec:order}). Second, they do not take into account the interaction among cuts when learning which cuts should be preferred, as they score each cut \textit{independently}. As a result, they struggle to select cuts that complement each other nicely, which could severely hinder the efficiency of solving MILPs \citep{theoretical_cuts}. Indeed, we empirically show that they tend to select many similar cuts with high scores (see Experiment 4 in Section \ref{sec:visu}). 

To address the aforementioned challenges, we propose a novel \textbf{h}ierarchical s\textbf{e}quence \textbf{m}odel (HEM) to learn cut selection policies via reinforcement learning. To the best of our knowledge, HEM is \textit{the first} learning-based method that can tackle \textbf{(P1)-(P3)} simultaneously by proposing a two-level model. Specifically, HEM is comprised of (1) a higher-level model to learn the number of cuts that should be selected, (2) and a lower-level model to learn policies selecting an \textit{ordered subset} with the size
determined by the higher-level model. The lower-level model formulates the cut selection task as a sequence to sequence learning problem, leading to two major advantages. 
First, the sequence model is popular in capturing the underlying order information \citep{order_matters}, which is critical for tackling \textbf{(P3)}.
Second, the sequence model can well capture the \textit{interaction} among cuts, as it models the \textit{joint} conditional probability of the selected cuts given an input sequence of the candidate cuts.
As a result, experiments show that HEM significantly outperforms human-designed and learning-based baselines in terms of solving efficiency on three synthetic MILP problems and seven challenging MILP problems. The challenging MILP problems include some benchmarks from MIPLIB 2017 \citep{miplibs_2017} and large-scale real-world production planning problems. Our results demonstrate the strong ability to enhance modern MILP solvers with our proposed HEM in real-world applications.
Moreover, experiments demonstrate that HEM can well generalize to MILPs that are significantly larger than those seen during training.

We summarize our major contributions as follows. (1) We observe from extensive empirical results that the \textit{order of selected cuts} has a significant impact on the efficiency of solving MILPs (see Section \ref{sec:order}). (2) To the best of our knowledge, our proposed HEM is \textit{the first} method that is able to tackle \textbf{(P1)-(P3)} in cut selection simultaneously from a data-driven perspective. (3) We propose to formulate the cut selection task as a sequence to sequence learning problem, which not only can capture the underlying order information, but also well captures the interaction among cuts to select cuts that complement each other nicely. (4) Experiments demonstrate that HEM achieves significant improvements over competitive baselines on challenging MILP problems, including some benchmarks from MIPLIB 2017 and large-scale real-world production planning problems.      

\vspace{-2mm}
\section{Background}
\vspace{-2mm}


\noindent \textbf{Cutting planes.} 
    Given the MILP problem in (\ref{milp1}), we drop all its integer constraints to obtain its \textit{linear programming (LP) relaxation}, which takes the form of  
    \begin{align}\label{lp_relaxation}
        z_{\text{LP}}^* \triangleq \min_{\textbf{x}} \{\textbf{c}^{\top} \textbf{x} | \textbf{A}\textbf{x} \leq \textbf{b},\textbf{x}\in \mathbb{R}^n\}.
    \end{align}
    Since the problem in (\ref{lp_relaxation}) expands the feasible set of the problem in (\ref{milp1}), we have $z_{\text{LP}}^* \leq z^*$. We denote any lower bound found via an LP relaxation by a \textit{dual bound}.
    Given the LP relaxation in (\ref{lp_relaxation}), cutting planes (cuts) are linear inequalities that are added to the LP relaxation in the attempt to tighten it without removing any integer feasible solutions of the problem in (\ref{milp1}). Cuts generated by MILP solvers are added in successive rounds. Specifically, each round $k$ involves (i) solving the current LP relaxation, (ii) generating a pool of candidate cuts $\mathcal{C}^k$, (iii) selecting a subset $\mathcal{S}^k\subseteq \mathcal{C}^k$, (iv) adding $\mathcal{S}^k$ to the current LP relaxation to obtain the next LP relaxation, (v) and proceeding to the next round.
    Adding all the generated cuts to the LP relaxation would maximally strengthen the LP relaxation and improve the lower bound at each round. However, adding too many cuts could lead to large models, which can increase the computational burden and present numerical instabilities \citep{implementing_cutting, l2c_lookahead}. Therefore, 
    cut selection is proposed to select a proper subset of the candidate cuts, which is significant for improving the efficiency of solving MILPs \citep{theoretical_cuts, tang_icml20}.

\noindent \textbf{Branch-and-cut.} 
    In modern MILP solvers, cutting planes are often combined with the branch-and-bound algorithm \citep{branch_and_bound}, which is known as the branch-and-cut algorithm 
    \citep{branch_and_cut}.
    Branch-and-bound techniques perform implicit enumeration by building a search tree, in which every node represents a subproblem of the original problem in (\ref{milp1}).
    The solving process begins by selecting a leaf node of the tree and solving its LP relaxation. Let $\textbf{x}^*$ be the optimal solution of the LP relaxation. If $\textbf{x}^*$ violates the original integrality constraints, two subproblems (child nodes) of the leaf node are created by \textit{branching}. Specifically, the leaf node is added with constraints $ x_i \leq \lfloor x_i^* \rfloor\,\,\text{and}\,\, x_i \geq \lceil x_i^* \rceil $, respectively, where $x_i$ denotes the $i$-th variable, $x_i^*$ denotes the $i$-th entry of vector $\textbf{x}^*$, and $\lfloor \rfloor$ and $\lceil \rceil$ denote the floor and ceil functions.
    In contrast, if $\textbf{x}^*$ is a (mixed-)integer solution of (\ref{milp1}), then we obtain an upper bound on the optimal objective value of (\ref{milp1}), which we denote by \textit{primal bound}.
    In modern MILP solvers, the addition of cutting planes is alternated with the \textit{branching} phase. That is, cuts are added at search tree nodes before branching to tighten their LP relaxations. 
    Since strengthening the relaxation before starting to branch is decisive to ensure an efficient tree search \citep{implementing_cutting, bengio_ml4co}, we focus on only adding cuts at the root node, which follows \citet{nips19_gcnn, l2c_lookahead}.

\noindent \textbf{Primal-dual gap integral.}
        We keep track of two important bounds when running branch-and-cut, i.e., the global primal and dual bounds, which are the best upper and lower bounds on the optimal objective value of (\ref{milp1}), respectively.
        We define the \textit{primal-dual gap integral} (PD integral) by the \textit{area} between the curve of the solver's global primal bound and the curve of the solver's global dual bound. 
        The PD integral is a 
        widely used metric for evaluating solver performance \citep{nips21_ml4co_competition, cao2022ml4co}.
        We provide more details in Appendix \ref{details_pd_integral}.

\begin{wrapfigure}{r}{0.55\textwidth}
    \centering
    \vspace{-6mm}
    \begin{subfigure}{0.27\textwidth}
        \includegraphics[width=\textwidth,height=0.6\textwidth]{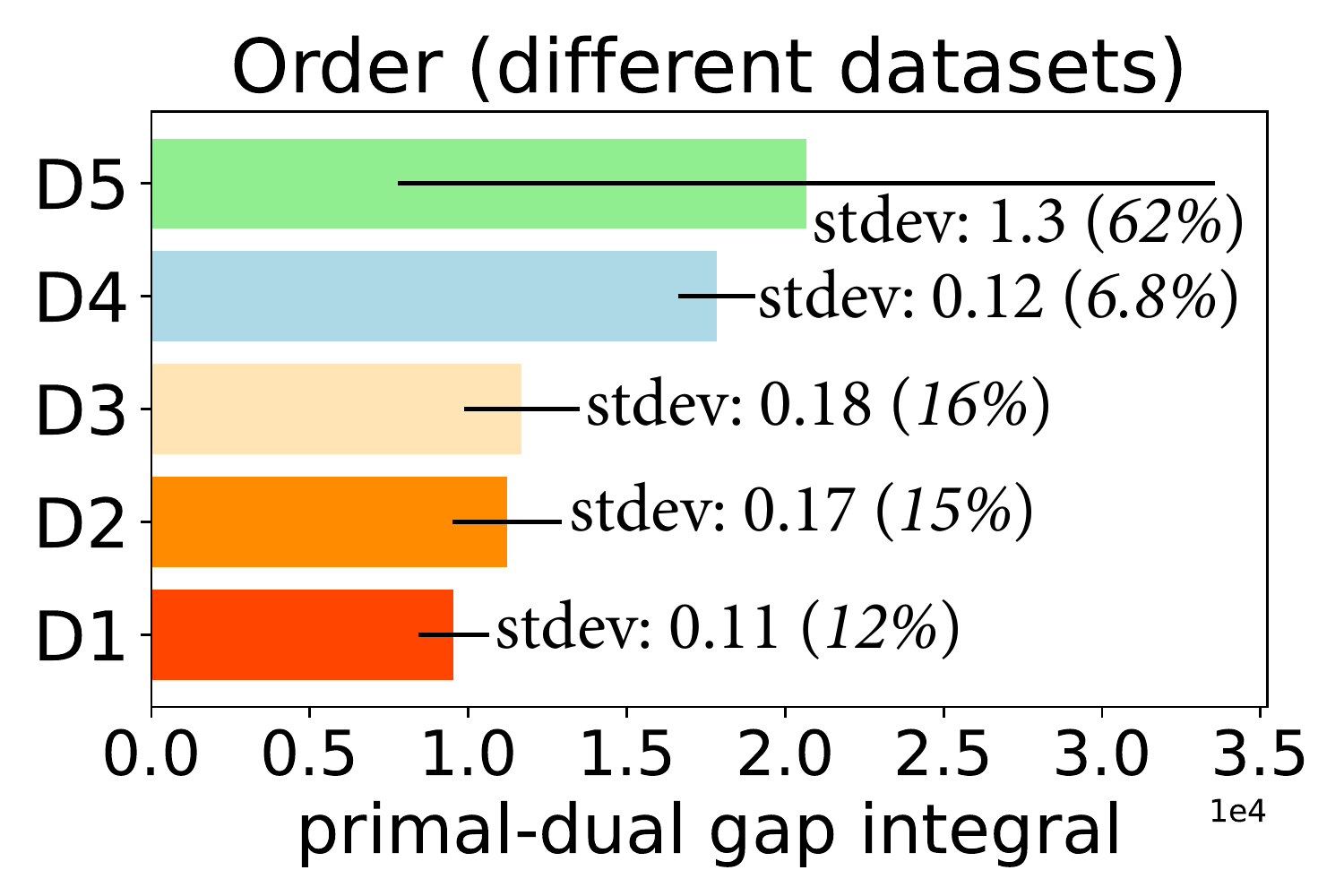}
        \vspace{-5mm}
        \caption{Evaluate RandomAll on five different datasets.}
        \label{fig:order_randomall}
    \end{subfigure}
    \begin{subfigure}{0.27\textwidth}
        \includegraphics[width=\textwidth,height=0.6\textwidth]{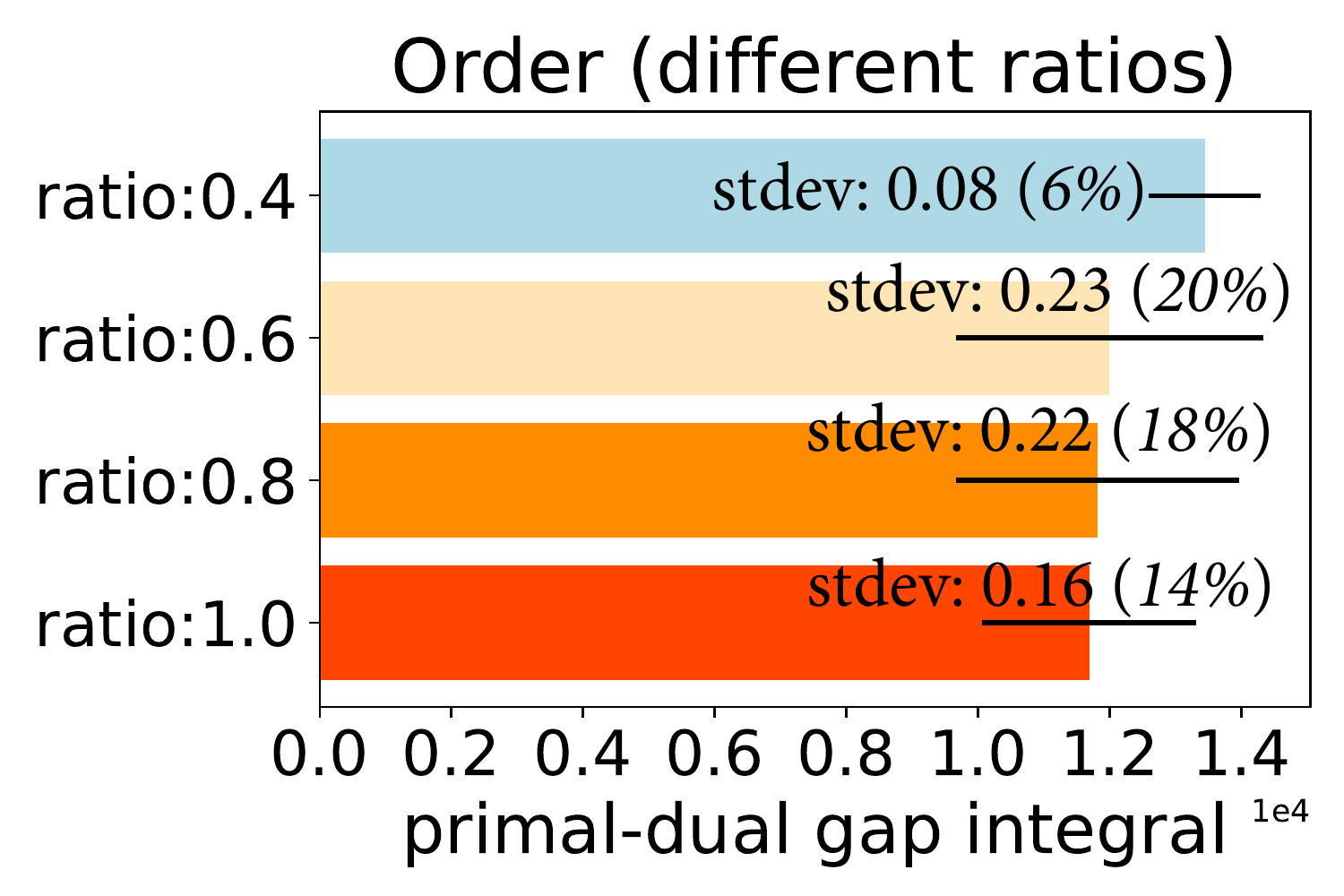}
        \vspace{-5mm}
        \caption{Evaluate RandomNv on MIPLIB mixed neos.}
        \label{fig:order_randomnv}
    \end{subfigure}
    \vspace{-3mm}
    \caption{ 
    We design two cut selection heuristics, namely RandomAll and RandomNV (see Section \ref{sec:order} for details), which both add the same subset of cuts in random order for a given MILP. The results in (a) and (b) show that adding the same selected cuts in different order leads to variable overall solver performance.
    }
    \vspace{-3mm}
    \label{fig:motivating_results_order_size}
\end{wrapfigure}
\vspace{-2mm}
\section{Motivating Results}
\vspace{-2mm}
    We empirically show that
    the \textit{order of selected cuts}, i.e., the selected cuts are added to the LP relaxations in this order, significantly impacts the efficiency of solving MILPs. Moreover, we empirically show that the \textit{ratio of selected cuts} matters significantly when solving MILPs (see Appendix \ref{motivating_results_ratio}). 
    Please see Appendix \ref{datasets_motivating} for details of the datasets used in this section.

\noindent \textbf{Order matters.} \label{sec:order} 
Previous work \citep{implementing_simplex, lp_textbook, Reformulate} has shown that the order of constraints for a given linear program (LP) significantly impacts its constructed initial basis, which is important for solving the LP. As a cut is a linear constraint, adding cuts to the LP relaxations is equivalent to adding constraints to the LP relaxations. Therefore, the order of added cuts could have a significant impact on solving the LP relaxations as well, thus being important for solving MILPs. Indeed, our empirical results show that this is the case.
(1) We design a \textbf{RandomAll} cut selection rule,
which randomly permutes all the candidate cuts, and adds all the cuts to the LP relaxations in the random order. 
We evaluate RandomAll on five challenging datasets, namely D1, D2, D3, D4, and D5. 
We use the SCIP 8.0.0 \citep{scip8} as the backend solver, and evaluate the solver performance by the average PD integral within a time limit. We evaluate RandomAll on each dataset over ten random seeds, and each bar in Figure \ref{fig:order_randomall} shows the mean and standard deviation (stdev) of its performance on each dataset.
As shown in Figure \ref{fig:order_randomall},
the performance of RandomAll on each dataset varies widely with the order of selected cuts.
(2) We further design a \textbf{RandomNV} cut selection rule. RandomNV is different from RandomAll in that it selects a given ratio of the candidate cuts rather than all the cuts. RandomNV first scores each cut using the Normalized Violation \citep{cut_ranking} and selects a given ratio of cuts with high scores. It then randomly permutes the selected cuts. Each bar in Figure \ref{fig:order_randomnv} shows the mean and stdev of the performance of RandomNV with a given ratio on the same dataset.
Figures \ref{fig:order_randomall} and \ref{fig:order_randomnv} show that adding the same selected cuts in different order leads to variable solver performance, which demonstrates that the order of selected cuts is important for solving MILPs.

\vspace{-2mm}
\section{Learning Cut Selection via Hierarchical Sequence Model}\label{sec_methods}
\vspace{-2mm}

In the cut selection task, the optimal subsets that should be selected are inaccessible, but one can assess the quality of selected subsets using a solver and provide the feedbacks to learning algorithms. Therefore, 
we leverage reinforcement learning (RL) to learn cut selection policies. 
In this section, we provide a detailed description of our proposed RL framework for learning cut selection. First, we present our formulation of the cut selection as a Markov decision process (MDP) \citep{rl_sutton}. Then, we present a detailed description of our proposed \textbf{h}ierarchical s\textbf{e}quence \textbf{m}odel (HEM). Finally, we derive a hierarchical policy gradient for training HEM efficiently. 

\begin{wrapfigure}{r}{0.49\textwidth}
    \vspace{-1mm}
    \centering
    \includegraphics[width=0.48\textwidth]{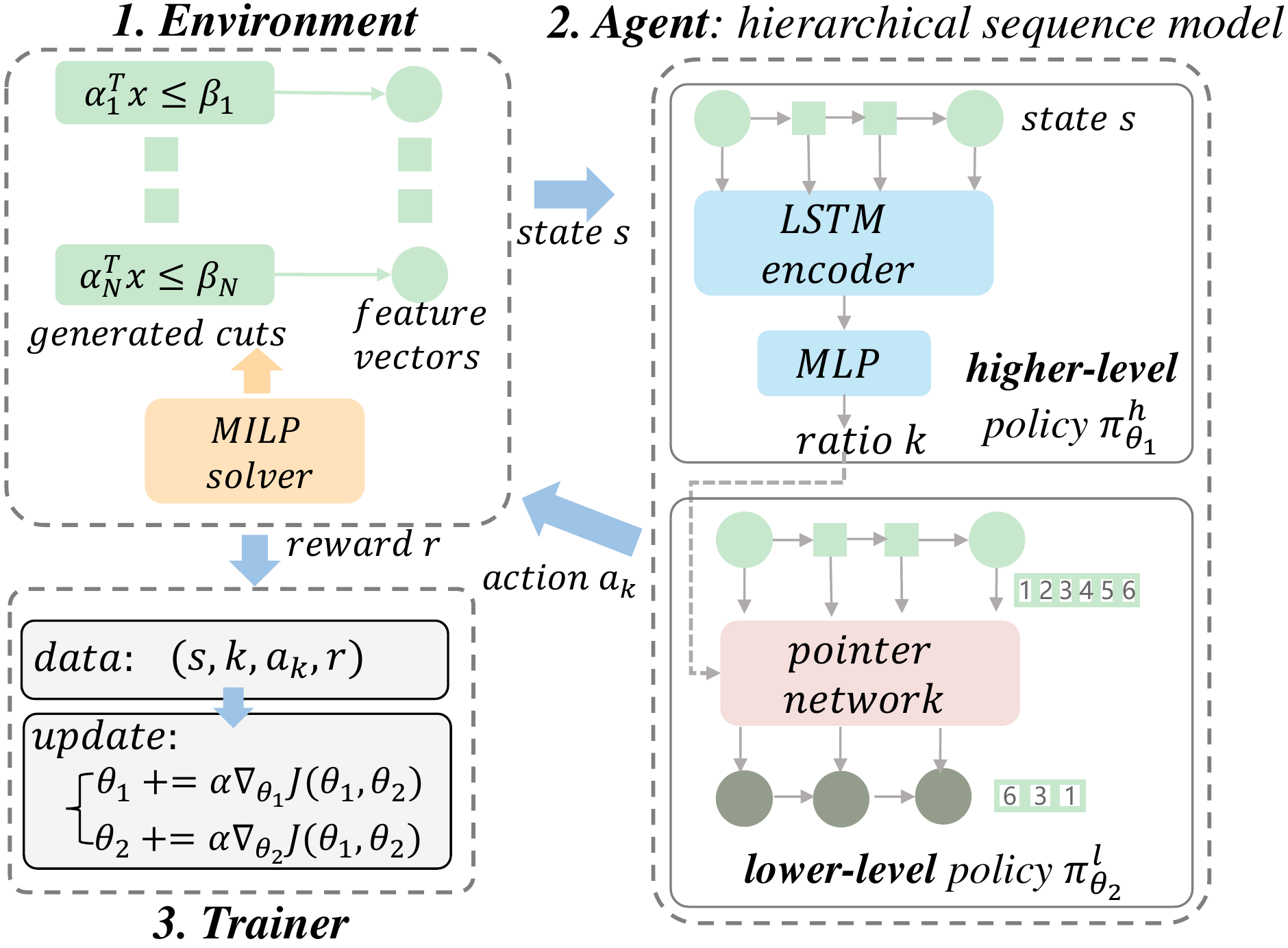}
    \vspace{-2.5mm}
    \caption{Illustration of our proposed RL framework for learning cut selection policies. We formulate a MILP solver as the environment and the HEM as the agent. Moreover, we train HEM via a hierarchical policy gradient algorithm.
    }
    \label{fig:hem_illustration}
    \vspace{-4mm}
\end{wrapfigure}
    \noindent \textbf{Reinforcement Learning Formulation}
    
    As shown in Figure \ref{fig:hem_illustration}, we formulate a MILP solver as the environment and our proposed HEM as the agent. 
    We consider an MDP defined by the tuple
    $(\mathcal{S}, \mathcal{A}, r, f)$.
    Specifically, we specify the state space $\mathcal{S}$, the action space $\mathcal{A}$, the reward function $r:\mathcal{S} \times \mathcal{A} \to \mathbb{R}$, the transition function $f$, and the terminal state in the following. 
    \textbf{(1) The state space $\mathcal{S}$.}  
    Since the current LP relaxation and the generated cuts contain the core information for cut selection, we define a state $s$ by $(M_{\text{LP}}, \mathcal{C}, \textbf{x}_{\text{LP}}^*)$. Here $M_{\text{LP}}$ denotes the mathematical model of the current LP relaxation, $\mathcal{C}$ denotes the set of the candidate cuts, and $\textbf{x}_{\text{LP}}^*$ denotes the optimal solution of the LP relaxation. To encode the state information, we 
    follow \citet{scip_thesis, cut_ranking} to design thirteen features for each candidate cut based on the information of $(M_{\text{LP}}, \mathcal{C}, \textbf{x}_{\text{LP}}^*)$. That is, we actually represent a state $s$ by \textit{a sequence of thirteen-dimensional feature vectors}. We present details of the designed features in Appendix \ref{appendix_cut_features}. 
    \textbf{(2) The action space $\mathcal{A}$.}  
    To take into account the ratio and order of selected cuts, we define the action space by \textit{all the ordered subsets} of the candidate cuts $\mathcal{C}$. 
    It can be challenging to explore the action space efficiently, as the cardinality of the action space can be extremely large due to its combinatorial structure.
    \textbf{(3) The reward function $r$.} 
    To evaluate the impact of the added cuts on solving MILPs, we design the reward function by (i) measures collected at the end of solving LP relaxations such as the dual bound improvement, (ii) or end-of-run statistics, such as the solving time and the primal-dual gap integral. For the first, the reward $r(s,a)$ can be defined as the negative dual bound improvement at each step. For the second, the reward $r(s,a)$ can be defined as zero except for the last step $(s_T,a_T)$ in a trajectory, i.e., $r(s_T,a_T)$ is defined by the negative solving time or the negative primal-dual gap integral. 
    \textbf{(4) The transition function $f$.} The transition function maps the current state $s$ and the action $a$ to the next state $s^{\prime}$, where $s^{\prime}$ represents the next LP relaxation generated by adding the selected cuts at the current LP relaxation. 
    \textbf{(5) The terminal state.}
    There is no standard and unified criterion to determine when to terminate the cut separation procedure \citep{l2c_lookahead}. Suppose we set the cut separation rounds as $T$, then the solver environment terminates the cut separation after $T$ rounds. Under the multiple rounds setting (i.e., $T>1$), we formulate the cut selection as a Markov decision process. Under the one round setting (i.e., $T=1$), the formulation can be simplified as a contextual bandit. 

    \noindent\textbf{Hierarchical Sequence Model}
    

    \noindent \textbf{Motivation.}
    Let $\pi$ denote the cut selection policy $\pi: \mathcal{S} \to \mathcal{P}(\mathcal{A})$, where $\mathcal{P}(\mathcal{A})$ denotes the probability distribution over the action space, and $\pi(\cdot|s)$ denotes the probability distribution over the action space given the state $s$. We emphasize that learning such policies can tackle \textbf{(P1)-(P3)} in cut selection simultaneously. 
    However, directly learning such policies is challenging for the following reasons. First, it is challenging to explore the action space efficiently, as the cardinality of the action space can be extremely large due to its combinatorial structure. Second, the length and max length of actions (i.e., ordered subsets) are variable across different MILPs. 
    However, traditional RL usually deals with problems whose actions have a fixed length. 
    Instead of directly learning the aforementioned policy, many existing learning-based methods \citep{tang_icml20, cut_ranking, l2c_lookahead} learn a scoring function that outputs a score given a cut, and select a fixed ratio/number of cuts with high scores. However, they suffer from two limitations as mentioned in Section \ref{sec:introduction}.

    \noindent \textbf{Policy network architecture.}
    To tackle the aforementioned problems, we propose a novel hierarchical sequence model (HEM) to learn cut selection policies. 
    To promote efficient exploration, HEM leverages the hierarchical structure of the cut selection task to decompose the policy into two sub-policies, i.e., a higher-level policy $\pi^h$ and a lower-level policy $\pi^l$. The policy network architecture of HEM is also illustrated in Figure \ref{fig:hem_illustration}. \textbf{First}, the higher-level policy learns the number of cuts that should be selected by predicting a proper ratio. Suppose the length of the state is $N$ and the predicted ratio is $k$, then the predicted number of cuts that should be selected is $\lfloor N*k \rfloor$, where $\lfloor \cdot \rfloor$ denotes the floor function. We define the higher-level policy by $\pi^h: \mathcal{S} \to \mathcal{P}(\left[0,1\right])$, where  $\pi^h(\cdot|s)$ denotes the probability distribution over $\left[0,1\right]$ given the state $s$.  \textbf{Second}, the lower-level policy learns to select an ordered subset with the size\footnote{We use the terms interchangeably: the \textit{size} of a selected subset and the \textit{number/ratio} of selected cuts.}  determined by the higher-level policy.
    We define the lower-level policy by $\pi^l: \mathcal{S} \times \left[0,1\right] \to \mathcal{P}(\mathcal{A})$, where $\pi^l(\cdot|s,k)$ denotes the probability distribution over the action space given the state $s$ and the ratio $k$. Specifically, 
    we formulate the lower-level policy as a sequence model, which can capture the interaction among cuts. \textbf{Finally}, we derive the cut selection policy via the law of total probability, i.e.,
    $\pi(a_k|s) = \mathbb{E}_{k\sim \pi^h(\cdot|s)}[\pi^l(a_k|s,k)],$
    where $k$ denotes the given ratio and $a_k$ denotes the action. The policy is computed by an expectation, as $a_k$ cannot determine the ratio $k$. For example, suppose that $N=100$ and the length of $a_k$ is $10$, then the ratio $k$ can be any number in the interval $[0.1,0.11)$. Actually, we sample an action from the policy $\pi$ by first sampling a ratio $k$ from $\pi^h$ and then sampling an action from $\pi^l$ given the ratio.

    For the higher-level policy, we first model the higher-level policy as a tanh-Gaussian, i.e., a Gaussian distribution with an invertible squashing function ($\tanh$), which is commonly used in deep reinforcement learning \citep{ppo,sac}. The mean and variance of the Gaussian are given by neural networks. 
    The support of the tanh-Gaussian is $[-1,1]$, but a ratio of selected cuts should belong to $[0,1]$. Thus, we further perform a linear transformation on the tanh-Gaussian. Specifically, we define the parameterized higher-level policy by 
    $\pi^{h}_{\theta_1}(\cdot|s) = 0.5 * \tanh{(K)} + 0.5$, where $K\sim \mathcal{N}(\mu_{\theta_1}(s), \sigma_{\theta_1}(s))$.
    Since the sequence lengths of states are variable across different instances (MILPs), we use a long-short term memory (LSTM) 
    \citep{lstm} network to embed the sequence of candidate cuts. We then use a multi-layer perceptron (MLP) \citep{deeplearning} to predict the mean and variance from the last hidden state of the LSTM. 
    
    For the lower-level policy, we formulate it as a sequence model. That is, its input is a sequence of candidate cuts, and its output is the probability distribution over ordered subsets of candidate cuts with the size determined by the higher-level policy. Specifically, given a state action pair $(s,k,a_k)$, the sequence model computes the conditional probability $\pi^l_{\theta_2}(a_k|s,k)$ using a parametric model to estimate the terms of the probability chain rule, i.e.,
    $\pi^{l}_{\theta_2}(a_k|s,k) = \prod_{i=1}^{m} p_{\theta_2}(a_k^i|a_k^1,\dots,a_k^{i-1},s,k)$.
    Here $s=\{s^1,\dots,s^N\}$ is the input sequence, $m=\lfloor N*k \rfloor$ is the length of the output sequence, and $a_k=\{a_k^1,\dots,a_k^m\}$ is a sequence of $m$ indices, each corresponding a position in the input sequence $s$. Such policy can be parametrized by the vanilla sequence model commonly used in machine translation \citep{seq2seq, transformer}. However, the vanilla sequence model can only be applied to learning on a single instance, as the number of candidate cuts varies on different instances. To generalize across different instances, we use a pointer network \citep{pn, neural_combinatorial}---which uses attention as a pointer to select a member of the input sequence as the output at each decoder step---to parametrize $\pi^l_{\theta_2}$ (see Appendix \ref{appendix_pn_details} for details). To the best of our knowledge, we are the first to formulate the cut selection task as a sequence to sequence learning problem and apply the pointer network to cut selection. This leads to two major advantages: (1) capturing the underlying order information, (2) and the interaction among cuts. This is also illustrated through an example in Appendix \ref{appendix_fig_sequence_model}.

    \noindent \textbf{Training: hierarchical policy gradient}
    
    For the cut selection task, we aim to find $\theta$ that maximizes the expected reward over all trajectories
    \begin{align}\label{eq:obj}
        J(\theta) = \mathbb{E}_{s\sim \mu, a_k \sim \pi_{\theta}(\cdot|s)}[r(s,a_k)],
    \end{align}
    where $\theta=
    \left[\theta_1,\theta_2\right]$ with $\left[\theta_1,\theta_2\right]$ denoting the concatenation of the two vectors, $\pi_{\theta}(a_k|s) = \mathbb{E}_{k\sim \pi^h_{\theta_1}(\cdot|s)}[\pi^l_{\theta_2}(a_k|s,k)]$, and $\mu$ denotes the initial state distribution.  

    To train the policy with a hierarchical structure, we derive a hierarchical policy gradient following the well-known policy gradient theorem \citep{pg_theorem,rl_sutton}.
    \begin{proposition}\label{proof_hpg}
        Given the cut selection policy $\pi_{\theta}(a_k|s) = \mathbb{E}_{k\sim \pi^h_{\theta_1}(\cdot|s)}[\pi^l_{\theta_2}(a_k|s,k)]$ and the training objective (\ref{eq:obj}), the hierarchical policy gradient takes the form of 
        \begin{align*}
            \nabla_{\theta_1}J(\left[\theta_1,\theta_2\right])
            & = \mathbb{E}_{s\sim\mu, k\sim \pi^h_{\theta_1}(\cdot|s)} [\nabla_{\theta_1} \log(\pi^h_{\theta_1}(k|s)) \mathbb{E}_{a_k\sim \pi^l_{\theta2}(\cdot|s,k)}[r(s,a_k)] ], \\
            \nabla_{\theta_2}J(\left[\theta_1,\theta_2\right])
            & = \mathbb{E}_{s\sim\mu, k\sim \pi^h_{\theta_1}(\cdot|s), a_k\sim \pi^l_{\theta_2}(\cdot|s,k)} [\nabla_{\theta_2}\log \pi^l_{\theta_2}(a_k|s,k) r(s,a_k)].
        \end{align*}
        \vspace{-6mm}
    \end{proposition}
    We provide detailed proof in Appendix \ref{appendix_proof}. We use the derived hierarchical policy gradient to update the parameters of the higher-level and lower-level policies. We implement the training algorithm in a parallel manner that is closely related to the asynchronous advantage actor-critic (A3C) \citep{a3c}. Due to limited space, we summarize the procedure of the training algorithm in Appendix  \ref{alg_pseudocode}. Moreover, we discuss some more advantages of HEM (see Appendix \ref{hem_advantages} for details). (1) HEM leverages the hierarchical structure of the cut selection task, which is important for efficient exploration in complex decision-making tasks \citep{smdp_sutton}. (2) We train HEM via gradient-based algorithms, which is sample efficient \citep{rl_sutton}.

\vspace{-2mm}
\section{Experiments}
\vspace{-2mm}

Our experiments have five main parts: \textbf{Experiment 1.} Evaluate our approach on three classical MILP problems and six challenging MILP problem benchmarks from diverse application areas. \textbf{Experiment 2.} Perform carefully designed ablation studies to provide further insight into HEM. \textbf{Experiment 3.} Test whether HEM can generalize to instances significantly larger than those seen during training. \textbf{Experiment 4.} Visualize the cuts selected by our method compared to the baselines. \textbf{Experiment 5.} Deploy our approach to real-world production planning problems.

\noindent \textbf{Benchmarks.}
    We evaluate our approach on nine $\mathcal{NP}$-hard MILP problem benchmarks, which consist of three classical synthetic MILP problems and six challenging MILP problems from diverse application areas. We divide the nine problem benchmarks into three categories according to the difficulty of solving them using the SCIP 8.0.0 solver \citep{scip8}. We call the three categories easy, medium, and hard datasets, respectively. (1) \textit{Easy datasets} comprise three widely used synthetic MILP problem benchmarks: Set Covering \citep{setcover}, Maximum Independent Set \citep{mis}, and Multiple Knapsack \citep{tree_mdp}. We artificially generate instances following \citet{nips19_gcnn, rl_branch}. (2) \textit{Medium datasets} comprise MIK \citep{mik} and CORLAT \citep{corlat}, which are widely used benchmarks for evaluating MILP solvers \citep{learning_to_search, milp_google}. (3) \textit{Hard datasets} include the Load Balancing problem, inspired by real-life applications of large-scale systems, and the Anonymous problem, inspired by a large-scale industrial application \citep{nips21_ml4co_competition}. Moreover, hard datasets contain benchmarks from MIPLIB 2017 (MIPLIB) \citep{miplibs_2017}. Although \citet{adaptive_cut_selection} has shown that directly learning over the full MIPLIB can be extremely challenging, we propose to learn over subsets of MIPLIB. We construct two subsets, called MIPLIB mixed neos and MIPLIB mixed supportcase. Due to limited space, please see Appendix \ref{appendix_main_datasets} for details of these datasets.

\begin{table*}[t]
\caption{Policy evaluation on the easy, medium, and hard datasets. The best performance is marked in bold. Let $m$ denote the average number of constraints and $n$ denote the average number of variables. We report the arithmetic mean (standard deviation) of the Time and PD integral. 
}
\vspace{-3mm}
\label{evaluation_all}
\centering
\resizebox{\textwidth}{!}{
\begin{tabular}{@{}cccccccccc@{}}
\toprule
\toprule
 & \multicolumn{3}{c}{Easy: Set Covering ($n=1000, \,\,m=500$)} & \multicolumn{3}{c}{Easy: Maximum Independent Set ($n=500,\,\,m=1953$)} & \multicolumn{3}{c}{Easy: Multiple Knapsack ($n=720,\,\,m=72$)} \\ \midrule
Method & Time(s) $\downarrow$ & Improvement (Time, \%) $\uparrow$ & PD integral $\downarrow$ & Time(s) $\downarrow$ & Improvement (Time, \%) $\uparrow$ & PD integral $\downarrow$ & Time(s) $\downarrow$ & Improvement (Time, \%) $\uparrow$ & PD integral $\downarrow$\\\cmidrule(r){1-4} \cmidrule(lr){5-7} \cmidrule(l){8-10}
NoCuts & 6.31 (4.61) & NA & 56.99 (38.89) & 8.78 (6.66) & NA & 71.31 (51.74) & 9.88 (22.24) & NA & 16.41 (14.16) \\
Default & 4.41 (5.12) & 29.90 & 55.63 (42.21) & 3.88 (5.04) & 55.80 & 29.44 (35.27) & 9.90 (22.24) & -0.20 & 16.46 (14.25) \\
Random & 5.74 (5.19) & 8.90 & 67.08 (46.58) & 6.50 (7.09) & 26.00 & 52.46 (53.10) & 13.10 (35.51) & -32.60 & 20.00 (25.14) \\
NV & 9.86 (5.43) & -56.50 & 99.77 (53.12) & 7.84 (5.54) & 10.70 & 61.60 (43.95) & 13.04 (36.91) & -32.00 & 21.75 (24.71) \\
Eff & 9.65 (5.45) & -53.20 & 95.66 (51.71) & 7.80 (5.11) & 11.10 & 61.04 (41.88) & 9.99 (19.02) & -1.10 & 20.49 (22.11) \\\midrule
SBP & 1.91 (0.36) & 69.60 & 38.96 (8.66) & 2.43 (5.55) & 72.30 & 21.99 (40.86) & 7.74 (12.36) & 21.60 & 16.45 (16.62) \\
HEM (Ours) & \textbf{1.85 (0.31)} & \textbf{70.60} & \textbf{37.92 (8.46)} & \textbf{1.76 (3.69)} & \textbf{80.00} & \textbf{16.01 (26.21)} & \textbf{6.13 (9.61)} & \textbf{38.00} & \textbf{13.63 (9.63)} \\ \bottomrule
\end{tabular}
}
\newline
\vspace{1mm}
\newline
\resizebox{\textwidth}{!}{
\begin{tabular}{@{}cccccccccc@{}}
\toprule
\toprule
 & \multicolumn{3}{c}{Medium: MIK ($n=413,\,\,m=346$)} & \multicolumn{3}{c}{Medium: Corlat ($n=466,\,\,m=486$)} & \multicolumn{3}{c}{Hard: Load Balancing ($n=61000,\,\,m=64304$)} \\ \midrule
\multirow{2}{*}{Method} & \multirow{2}{*}{Time(s) $\downarrow$} & \multirow{2}{*}{PD integral $\downarrow$} & \multirow{2}{*}{\begin{tabular}[c]{@{}c@{}}Improvement $\uparrow$\\      (PD integral, \%)\end{tabular} } & \multirow{2}{*}{Time(s) $\downarrow$} & \multirow{2}{*}{PD integral $\downarrow$} & \multirow{2}{*}{\begin{tabular}[c]{@{}c@{}}Improvement $\uparrow$\\      (PD integral, \%)\end{tabular} } & \multirow{2}{*}{Time(s) $\downarrow$} & \multirow{2}{*}{PD integral $\downarrow$} & \multirow{2}{*}{\begin{tabular}[c]{@{}c@{}}Improvement $\uparrow$\\      (PD integral, \%)\end{tabular} } \\
 &  &  &  &  &  &  &  &  &  \\\cmidrule(r){1-4} \cmidrule(lr){5-7} \cmidrule(l){8-10}
NoCuts & 300.01 (0.009) & 2355.87 (996.08) & NA & 103.30 (128.14) & 2818.40 (5908.31) & NA & 300.00 (0.12) & 14853.77 (951.42) & NA \\
Default & 179.62 (122.36) & 844.40 (924.30) & 64.10 & 75.20 (120.30) & 2412.09 (5892.88) & 14.40 & 300.00 (0.06) & 9589.19 (1012.95) & 35.40 \\
Random & 289.86 (28.90) & 2036.80 (933.17) & 13.50 & 84.18 (124.34) & 2501.98 (6031.43) & 11.20 & 300.00 (0.09) & 13621.20 (1162.02) & 8.30 \\
NV & 299.76 (1.32) & 2542.67 ( 529.49) & -7.90 & 90.26 (128.33) & 3075.70 (7029.55) & -9.10 & 300.00 (0.05) & 13933.88 (971.10) & 6.20 \\
Eff & 298.48 (5.84) & 2416.57 (642.41) & -2.60 & 104.38 (131.61) & 3155.03 (7039.99) & -11.90 & 300.00 (0.07) & 13913.07 (969.95) & 6.30 \\\midrule
SBP & 286.07 (41.81) & 2053.30 (740.11) & 12.80 & 70.41 (122.17) & 2023.87 (5085.96) & 28.20 & 300.00 (0.10) & 12535.30 (741.43) & 15.60 \\ 
HEM (Ours) & \textbf{176.12 (125.18)} & \textbf{785.04 (790.38)} & \textbf{66.70} & \textbf{58.31 (110.51)} & \textbf{1079.99 (2653.14)} & \textbf{61.68} & \textbf{300.00 (0.04)} & \textbf{9496.42 (1018.35)} & \textbf{36.10} \\ \bottomrule
\end{tabular}
}
\newline
\vspace{1mm}
\newline
\resizebox{\textwidth}{!}{
\begin{tabular}{@{}cccccccccc@{}}
\toprule
\toprule
 & \multicolumn{3}{c}{Hard: Anonymous ($n=37881,\,\,m=49603$)} & \multicolumn{3}{c}{Hard: MIPLIB mixed neos ($n=6958,\,\,m=5660$)} & \multicolumn{3}{c}{Hard: MIPLIB mixed supportcase ($n=19766,\,\,m=19910$)} \\ \midrule
\multirow{2}{*}{Method} & \multirow{2}{*}{Time(s) $\downarrow$} & \multirow{2}{*}{PD integral $\downarrow$} & \multirow{2}{*}{\begin{tabular}[c]{@{}c@{}}Improvement $\uparrow$\\      (PD integral, \%)\end{tabular}} & \multirow{2}{*}{Time(s) $\downarrow$} & \multirow{2}{*}{PD integral $\downarrow$} & \multirow{2}{*}{\begin{tabular}[c]{@{}c@{}}Improvement $\uparrow$\\      (PD integral, \%)\end{tabular}} & \multirow{2}{*}{Time(s) $\downarrow$} & \multirow{2}{*}{PD integral $\downarrow$} & \multirow{2}{*}{\begin{tabular}[c]{@{}c@{}}Improvement $\uparrow$\\      (PD integral, \%)\end{tabular}} \\
 &  &  &  &  &  &  &  &  &  \\\cmidrule(r){1-4} \cmidrule(lr){5-7} \cmidrule(l){8-10}
NoCuts & 246.22 (94.90) & 18297.30 (9769.42) & NA & 253.65 (80.29) & 14652.29 (12523.37) & NA & 170.00 (131.60) & 9927.96 (11334.07) & NA \\
Default & 244.02 (97.72) & 17407.01 (9736.19) & 4.90 & 256.58 (76.05) & 14444.05 (12347.09) & 1.42 & 164.61 (135.82) & 9672.34 (10668.24) & 2.57 \\
Random & 243.49 (98.21) & 16850.89 (10227.87) & 7.80 & 255.88 (76.65) & 14006.48 (12698.76) & 4.41 & 165.88 (134.40) & 10034.70 (11052.73) & -1.07 \\
NV & 242.01 (98.68) & 16873.66 (9711.16) & 7.80 & 263.81 (64.10) & 14379.05 (12306.35) & 1.86 & \textbf{161.67 (131.43)} & 8967.00 (9690.30) & 9.68 \\
Eff & 244.94 (93.47) & 17137.87 (9456.34) & 6.30 & 260.53 (68.54) & 14021.74 (12859.41) & 4.30 & 167.35 (134.99) & 9941.55 (10943.48) & -0.14 \\\midrule
SBP & 245.71 (92.46) & 18188.63 (9651.85) & 0.59 & 256.48 (78.59) & 13531.00 (12898.22) & 7.65 & 165.61 (135.25) & 7408.65 (7903.47) & 25.37 \\ 
HEM (Ours) & \textbf{241.68 (97.23)} & \textbf{16077.15 (9108.21)} & \textbf{12.10} & \textbf{248.66 (89.46)} & \textbf{8678.76 (12337.00)} & \textbf{40.77} & 162.96 (138.21) & \textbf{6874.80 (6729.97)} & \textbf{30.75} \\ \bottomrule
\end{tabular}
}
\end{table*}

\noindent \textbf{Experimental setup.}
    Throughout all experiments, we use SCIP 8.0.0 \citep{scip8} as the backend solver, which is the state-of-the-art open source solver, and is widely used in research of machine learning for combinatorial optimization \citep{nips19_gcnn,cut_ranking,adaptive_cut_selection,milp_google}. Following \citet{nips19_gcnn, cut_ranking, l2c_lookahead}, we only allow cutting plane generation and selection at the root node, and set the cut separation rounds as one. We keep all the other SCIP parameters to default so as to make comparisons as fair and reproducible as possible. We emphasize that all of the SCIP solver's advanced features, such as presolve and heuristics, are open, which ensures that our setup is consistent with the practice setting. Throughout all experiments, we set the solving time limit as 300 seconds. For completeness, we also evaluate HEM with a much longer time limit of three hours. The results are given in Appendix \ref{appendix_results_long_test_time}. We train HEM with ADAM \citep{adam} using the PyTorch \citep{torch}. Additionally, we also provide another implementation using the MindSpore \citep{mindspore}.      
    For simplicity, we split each dataset into the train and test sets with $80\%$ and $20\%$ instances. To further improve HEM, one can construct a valid set for hyperparameters tuning. We train our model on the train set, and select the best model on the train set to evaluate on the test set. 
    Please refer to Appendix \ref{appendix_imple_hyper} for implementation details, hyperparameters, and hardware specification.
    
\noindent \textbf{Baselines.}
    Our baselines include five widely used human-designed cut selection rules and a state-of-the-art (SOTA) learning-based method. Cut selection rules include NoCuts, Random, Normalized Violation (NV), Efficacy (Eff), and Default. NoCuts does not add any cuts. 
    Default denotes the default cut selection rule used in SCIP 8.0.0. For learning-based methods, we implement a \textit{slight} variant of the SOTA learning-based methods \citep{tang_icml20, cut_ranking}, namely score-based policy (SBP).
    Please see Appendix \ref{appendix_imple_baselines} for implementation details of these baselines.

\noindent\textbf{Evaluation metrics.}
    We use two widely used evaluation metrics, i.e., the average solving time (Time, lower is better), and the average primal-dual gap integral (PD integral, lower is better). Additionally, we provide more results in terms of another two metrics, i.e., the average number of nodes and the average primal-dual gap, in Appendix \ref{appendix_results_more_metrics}. Furthermore, to evaluate different cut selection methods compared to pure branch-and-bound without cutting plane separation, we propose an \textit{Improvement} metric. Specifically, we define the metric by 
    $  \text{Im}_{M}(\cdot) = \frac{M(\text{NoCuts}) - M(\cdot)}{M(\text{NoCuts})},
    $
    where $M(\text{NoCuts})$ represents the performance of NoCuts, and $M(\cdot)$ represents a mapping from a method to its performance. The improvement metric represents the improvement of a given method compared to NoCuts. \textit{We mainly focus on the Time metric on the easy datasets}, as the solver can solve all instances to optimality within the given time limit. However, HEM and the baselines cannot solve all instances to optimality within the time limit on the medium and hard datasets. As a result, the average solving time of those unsolved instances is the same, which makes it difficult to distinguish the performance of different cut selection methods using the Time metric. Therefore, \textit{we mainly focus on the PD integral metric on the medium and hard datasets}. The PD integral is also a well-recognized metric for evaluating the solver performance \citep{nips21_ml4co_competition, cao2022ml4co}.

\noindent \textbf{Experiment 1. Comparative evaluation} 
    The results in Table \ref{evaluation_all} suggest the following.
    (1) \textbf{Easy datasets.} HEM significantly outperforms all the baselines on the easy datasets, especially on Maximum Independent Set and Multiple Knapsack. SBP achieves much better performance than all the rule-based baselines, demonstrating that our implemented SBP is a strong baseline. Compared to SBP, HEM improves the Time by up to $16.4\%$ on the three datasets,  demonstrating the superiority of our method over the SOTA learning-based method. 
    (2) \textbf{Medium datasets.} On MIK and CORLAT, HEM still outperforms all the baselines. Especially on CORLAT, HEM achieves at least $33.48\%$ improvement in terms of the PD integral compared to the baselines. 
    (3) \textbf{Hard datasets.}
    HEM significantly outperforms the baselines in terms of the PD integral on several problems in the hard datasets. HEM achieves outstanding performance on two challenging datasets from MIPLIB 2017 and real-world problems (Load Balancing and Anonymous), 
    demonstrating the powerful ability to enhance MILP solvers with HEM in large-scale real-world applications. Moreover, 
    SBP performs extremely poorly on several medium and hard datasets, which implies that it can be difficult to learn good cut selection policies on challenging MILP problems.

\begin{table*}[t]
\centering
\caption{Comparison between HEM and HEM without the higher-level model. 
}
\vspace{-3mm}
\label{ablation_each_component}
\resizebox{\textwidth}{!}{
\begin{tabular}{@{}cccccccccc@{}}
\toprule
\toprule
 & \multicolumn{3}{c}{Easy: Maximum Independent Set ($n=500,\,\,m=1953$)} & \multicolumn{3}{c}{Medium: Corlat ($n=466,\,\,m=486$)} & \multicolumn{3}{c}{Hard: MIPLIB mixed neos ($n=6958,\,\,m=5660$)} \\ \midrule
\multirow{2}{*}{Method} & \multirow{2}{*}{Time(s) $\downarrow$} & \multirow{2}{*}{\begin{tabular}[c]{@{}c@{}}Improvement $\uparrow$\\      (Time, \%)\end{tabular}} & \multirow{2}{*}{PD integral $\downarrow$} & \multirow{2}{*}{Time(s) $\downarrow$} & \multirow{2}{*}{PD integral $\downarrow$} & \multirow{2}{*}{\begin{tabular}[c]{@{}c@{}}Improvement $\uparrow$\\      (PD integral, \%)\end{tabular}} & \multirow{2}{*}{Time(s) $\downarrow$} & \multirow{2}{*}{PD integral $\downarrow$} & \multirow{2}{*}{\begin{tabular}[c]{@{}c@{}}Improvement $\uparrow$\\      (PD integral, \%)\end{tabular}} \\
 &  &  &  &  &  &  &  &  &  \\ \cmidrule(r){1-4} \cmidrule(lr){5-7} \cmidrule(l){8-10}
NoCuts & 8.78 (6.66) & NA & 71.31 (51.74) & 103.30 (128.14) & 2818.40 (5908.31) & NA & 253.65 (80.29) & 14652.29 (12523.37) & NA \\
Default & 3.88 (5.04) & 55.81 & 29.44 (35.27) & 75.20 (120.30) & 2412.09 (5892.88) & 14.42 & 256.58 (76.05) & 14444.05 (12347.09) & 1.42 \\
SBP & 2.43 (5.55) & 72.32 & 21.99 (40.86) & 70.41 (122.17) & 2023.87 (5085.96) & 28.19 & 256.48 (78.59) & 13531.00 (12898.22) & 7.65 \\ \midrule
HEM w/o H & 1.88 (4.20) & 78.59 & 16.70 (28.15) & 63.14 (115.26) & 1939.08 (5484.83) & 31.20 & 249.21 (88.09) & 13614.29 (12914.76) & 7.08 \\
HEM (Ours) & \textbf{1.76 (3.69)} & \textbf{79.95} & \textbf{16.01 (26.21)} & \textbf{58.31 (110.51)} & \textbf{1079.99 (2653.14)} & \textbf{61.68} & \textbf{248.66 (89.46)} & \textbf{8678.76 (12337.00)} & \textbf{40.77} \\ \bottomrule
\end{tabular}
}
\end{table*}

\begin{table*}[t]
\centering
\caption{Comparison between HEM, HEM-ratio, and HEM-ratio-order. 
}
\vspace{-3mm}
\label{ablation_each_factor}
\resizebox{\textwidth}{!}{
\begin{tabular}{@{}cccccccccc@{}}
\toprule
\toprule
 & \multicolumn{3}{c}{Easy: Maximum Independent Set ($n=500,\,\,m=1953$)} & \multicolumn{3}{c}{Medium: Corlat ($n=466,\,\,m=486$)} & \multicolumn{3}{c}{Hard: MIPLIB mixed neos ($n=6958,\,\,m=5660$)} \\ \midrule
\multirow{2}{*}{Method} & \multirow{2}{*}{Time(s) $\downarrow$} & \multirow{2}{*}{\begin{tabular}[c]{@{}c@{}}Improvement $\uparrow$\\      (Time, \%)\end{tabular}} & \multirow{2}{*}{PD integral $\downarrow$} & \multirow{2}{*}{Time(s) $\downarrow$} & \multirow{2}{*}{PD integral $\downarrow$} & \multirow{2}{*}{\begin{tabular}[c]{@{}c@{}}Improvement $\uparrow$\\      (PD integral, \%)\end{tabular}} & \multirow{2}{*}{Time(s) $\downarrow$} & \multirow{2}{*}{PD integral $\downarrow$} & \multirow{2}{*}{\begin{tabular}[c]{@{}c@{}}Improvement $\uparrow$\\      (PD integral, \%)\end{tabular}} \\
 &  &  &  &  &  &  &  &  &  \\ \cmidrule(r){1-4} \cmidrule(lr){5-7} \cmidrule(l){8-10}
NoCuts & 8.78 (6.66) & NA & 71.31 (51.74) & 103.30 (128.14) & 2818.40 (5908.31) & NA & 253.65 (80.29) & 14652.29 (12523.37) & NA \\
Default & 3.88 (5.04) & 55.81 & 29.44 (35.27) & 75.20 (120.30) & 2412.09 (5892.88) & 14.42 & 256.58 (76.05) & 14444.05 (12347.09) & 1.42 \\
SBP & 2.43 (5.55) & 72.32 & 21.99 (40.86) & 70.41 (122.17) & 2023.87 (5085.96) & 28.19 & 256.48 (78.59) & 13531.00 (12898.22) & 7.65 \\\midrule
HEM-ratio-order & 2.30 (5.18) & 73.80 & 21.19 (38.52) & 70.94 (122.93) & 1416.66 (3380.10) & 49.74 & 245.99 (93.67) & 14026.75 (12683.60) & 4.27 \\
HEM-ratio & 2.26 (5.06) & 74.26 & 20.82 (37.81) & 67.27 (117.01) & 1251.60 (2869.87) & 55.59 & \textbf{244.87 (95.56)} & 13659.93 (12900.59) & 6.77 \\
HEM (Ours) & \textbf{1.76 (3.69)} & \textbf{79.95} & \textbf{16.01 (26.21)} & \textbf{58.31 (110.51)} & \textbf{1079.99 (2653.14)} & \textbf{61.68} & 248.66 (89.46) & \textbf{8678.76 (12337.00)} & \textbf{40.77} \\ \bottomrule
\end{tabular}
}
\end{table*}

\noindent \textbf{Experiment 2. Ablation study}
We present ablation studies on Maximum Independent Set (MIS), CORLAT, and MIPLIB mixed neos, which are representative datasets from the easy, medium, and hard datasets. We provide more results on the other datasets in Appendix \ref{appendix_results_ablation_study}. 

\noindent \textbf{Contribution of each component.}
    We perform ablation studies to understand the contribution of each component in HEM. 
    We report the performance of HEM and HEM without the higher-level model (HEM w/o H) in Table \ref{ablation_each_component}. HEM w/o H is essentially a pointer network. Note that it can still implicitly predicts the number of cuts that should be selected by predicting an end token as used in language tasks \citep{seq2seq}. Please see Appendix \ref{appendix_imple_rlk} for details.  \textbf{First},
    the results in Table \ref{ablation_each_component} show that HEM w/o H outperforms all the baselines on MIS and CORLAT, demonstrating the advantages of the lower-level model. Although HEM w/o H outperforms Default on MIPLIB mixed neos, HEM w/o H performs on par with SBP. A possible reason is that it is difficult for HEM w/o H to explore the action space efficiently, and thus HEM w/o H tends to be trapped to the local optimum. 
    \textbf{Second}, the results in Table \ref{ablation_each_component} show that HEM significantly outperforms HEM w/o H and the baselines on the three datasets. The results demonstrate that the higher-level model is important for efficient exploration in complex tasks, thus significantly improving the solving efficiency.
    
\noindent \textbf{The importance of tackling \textbf{(P1)-(P3)}.}
    We perform ablation studies to understand the 
    importance of tackling \textbf{(P1)-(P3)} in cut selection.
    (1) \textbf{HEM.} HEM tackles \textbf{(P1)-(P3)} in cut selection simultaneously. (2) \textbf{HEM-ratio.} In order not to learn how many cuts should be selected, we remove the higher-level model of HEM and \textit{force the lower-level model to select a fixed ratio of cuts}. We denote it by HEM-ratio. Note that HEM-ratio is different from HEM w/o H (see Appendix \ref{appendix_imple_rlk}). HEM-ratio tackles \textbf{(P1)} and \textbf{(P3)} in cut selection. (3) \textbf{HEM-ratio-order.} To further mute the effect of the order of selected cuts, we reorder the selected cuts given by HEM-ratio with the original index of the generated cuts, which we denote by HEM-ratio-order. HEM-ratio-order mainly tackles \textbf{(P1)} in cut selection.
    The results in Table \ref{ablation_each_factor} suggest the following. HEM-ratio-order significantly outperforms Default and NoCuts, demonstrating that tackling \textbf{(P1)} by data-driven methods is crucial. HEM significantly outperforms HEM-ratio in terms of the PD integral, demonstrating the significance of tackling \textbf{(P2)}. HEM-ratio outperforms HEM-ratio-order in terms of the Time and the PD integral, which demonstrates the importance of tackling \textbf{(P3)}. Moreover, HEM-ratio and HEM-ratio-order perform better than SBP on MIS and CORLAT, demonstrating the advantages of using the sequence model to learn cut selection over SBP. HEM-ratio and HEM-ratio-order perform on par with SBP on MIPLIB mixed neos. We provide possible reasons in Appendix \ref{appendix_analysis_hem_ratio}.

\begin{table}[t]
\caption{\textbf{Left}: The generalization ability of HEM. \textbf{Right}: Test on Production Planning problems.}
\begin{subtable}{0.54\textwidth}
    \centering
    \vspace{-3mm}
    \resizebox{0.99\textwidth}{0.05\textheight}{
    \begin{tabular}{@{}ccccccc@{}}
    \toprule
    \toprule
     & \multicolumn{3}{c}{Maximum Independent Set $(n=1000,\,\, m=3946,\,\,4\times)$} & \multicolumn{3}{c}{Maximum Independent Set $(n=5940,\,\, m=1500,\,\, 9\times)$} \\ \midrule
    \multirow{2}{*}{Method} & \multirow{2}{*}{Time(s) $\downarrow$} & \multirow{2}{*}{\begin{tabular}[c]{@{}c@{}}Improvement $\uparrow$\\      (Time, \%)\end{tabular}} & \multirow{2}{*}{PD integral $\downarrow$} & \multirow{2}{*}{Time(s) $\downarrow$} & \multirow{2}{*}{\begin{tabular}[c]{@{}c@{}}Improvement $\uparrow$\\      (Time, \%)\end{tabular}} & \multirow{2}{*}{PD integral $\downarrow$} \\
     &  &  &  &  &  &  \\\cmidrule(r){1-4} \cmidrule(l){5-7}
    NoCuts & 170.06 (100.61) & NA & 874.45 (522.29) & 300.00 (0) & NA & 2019.93 (353.27) \\
    Default & 42.40 (76.00) & 48.72 & 198.61 (331.20) & 111.18 (144.13) & 60.91 & 616.46 (798.94) \\
    Random & 118.25 (109.05) & -43.00 & 574.33 (516.11) & 245.13 (115.80) & 13.82 & 1562.20 (793.09) \\
    NV & 160.30 (101.41) & -93.86 & 784.98 (493.24) & 299.97 (0.49) & -5.46 & 1922.52 (349.67) \\
    Eff & 158.75 (100.40) & -91.98 & 779.63 (493.05) & 299.45 (3.77) & -5.28 & 1921.61 (361.26) \\\midrule
    SBP & 50.55 (89.14) & 38.87 & 253.81 (426.94) & 108.42 (143.68) & 61.88 & 680.41 (903.88) \\
    HEM (Ours) & \textbf{35.34 (67.91)} & \textbf{57.26} & \textbf{160.56 (282.03)} & \textbf{108.02 (143.02)} & \textbf{62.02} & \textbf{570.48 (760.65)} \\ \bottomrule
    \end{tabular}
    }
\end{subtable}
\begin{subtable}{0.45\textwidth}
    \centering
    \vspace{-3mm}
    \resizebox{0.99\textwidth}{0.05\textheight}{
        \begin{tabular}{@{}ccccc@{}}
        \toprule
        \toprule
         & \multicolumn{4}{c}{Production Planning ($n=3582.25,\,\,m=5040.42$)} \\ \midrule
        \multirow{2}{*}{Method} & \multirow{2}{*}{Time (s) $\downarrow$} & \multirow{2}{*}{\begin{tabular}[c]{@{}c@{}}Improvement $\uparrow$ \\      (Time, \%)\end{tabular}} & \multirow{2}{*}{PD integral $\downarrow$} & \multirow{2}{*}{\begin{tabular}[c]{@{}c@{}}Improvement $\uparrow$\\      (PD integral, \%)\end{tabular}} \\
         &  &  &  &  \\ \midrule
        NoCuts & 278.79 (231.02) & NA & 17866.01 (21309.85) & NA \\
        Default & 296.12 (246.25) & -6.22 & 17703.39 (21330.40) & 0.91 \\
        Random & 280.18 (237.09) & -0.50 & 18120.21 (21660.01) & -1.42 \\
        NV & 259.48 (227.81) & 6.93 & 17295.18 (21860.07) & 3.20 \\
        Eff & 263.60 (229.24) & 5.45 & 16636.52 (21322.89) & 6.88 \\ \midrule
        SBP & 276.61 (235.84) & 0.78 & 16952.85 (21386.07) & 5.11 \\
        HEM (Ours) & \textbf{241.77 (229.97)} & \textbf{13.28} & \textbf{15751.08 (20683.53)} & \textbf{11.84} \\ \bottomrule
        \end{tabular}
    }
\end{subtable}
\label{generalization_setcovering_mis}
\end{table}

\noindent \textbf{Experiment 3. Generalization}
    We evaluate the ability of HEM to generalize across different sizes of MILPs. Let $n\times m$ denote the size of MILP instances. Following \citet{nips19_gcnn, rl_branch}, we test the generalization ability of HEM on Set Covering and Maximum Independent Set (MIS), as we can artificially generate instances with arbitrary sizes for synthetic MILP problems. On MIS, we test HEM on four times and nine times larger instances than those seen during training. 
    The results in Table \ref{generalization_setcovering_mis} (\textbf{Left}) show that HEM significantly outperforms the baselines in terms of the Time and the PD integral on $4\times$ and $9\times$ MIS, demonstrating the superiority of HEM in terms of the generalization ability.
    Interestingly, SBP also generalizes well to large instances, demonstrating that SBP is a strong baseline. We provide more results on Set Covering in Appendix \ref{appendix_gener}.

\begin{wrapfigure}{r}{0.63\textwidth}
    \vspace{-1.5mm}
    \centering
    \includegraphics[width=0.31\textwidth]{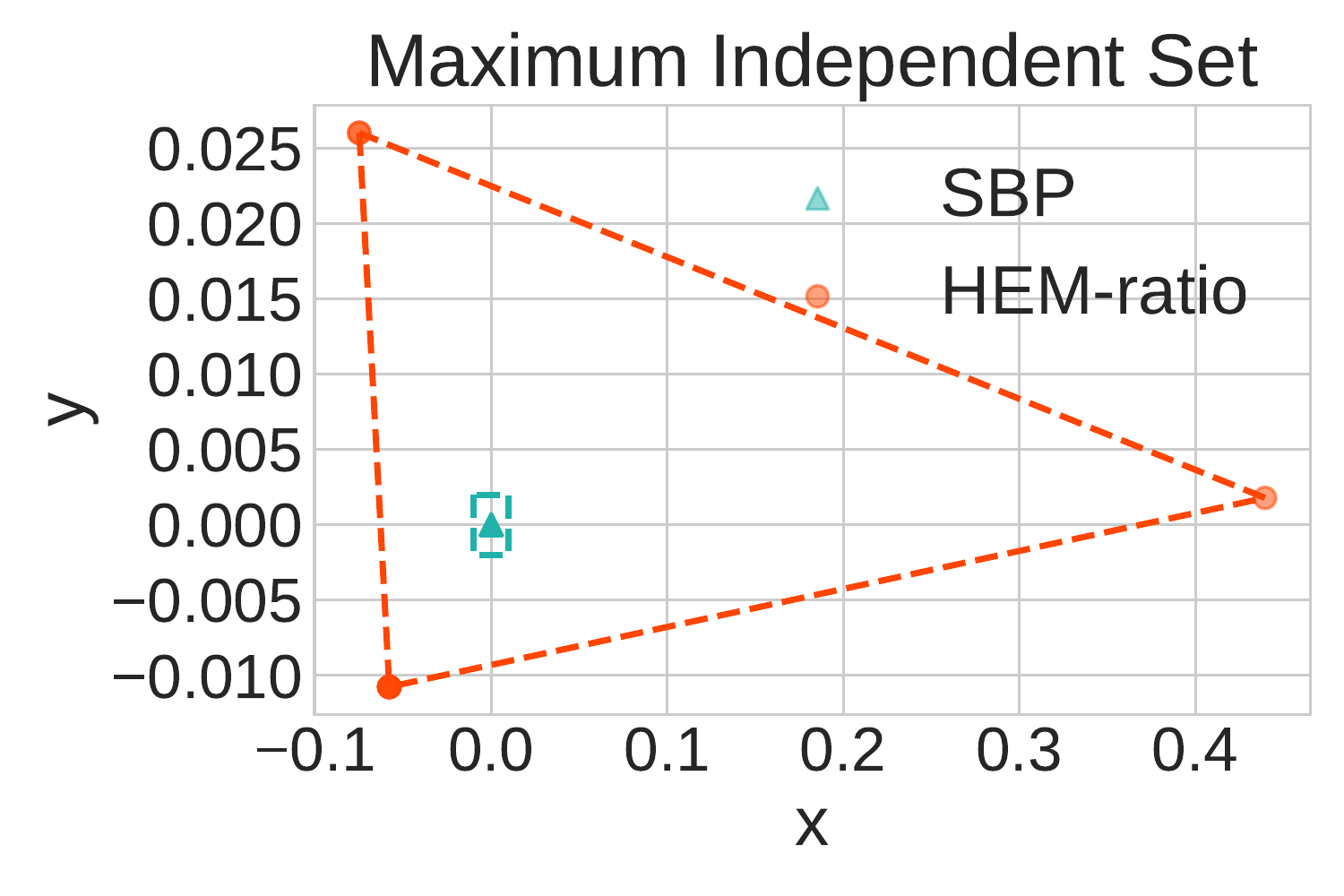}
    \includegraphics[width=0.31\textwidth]{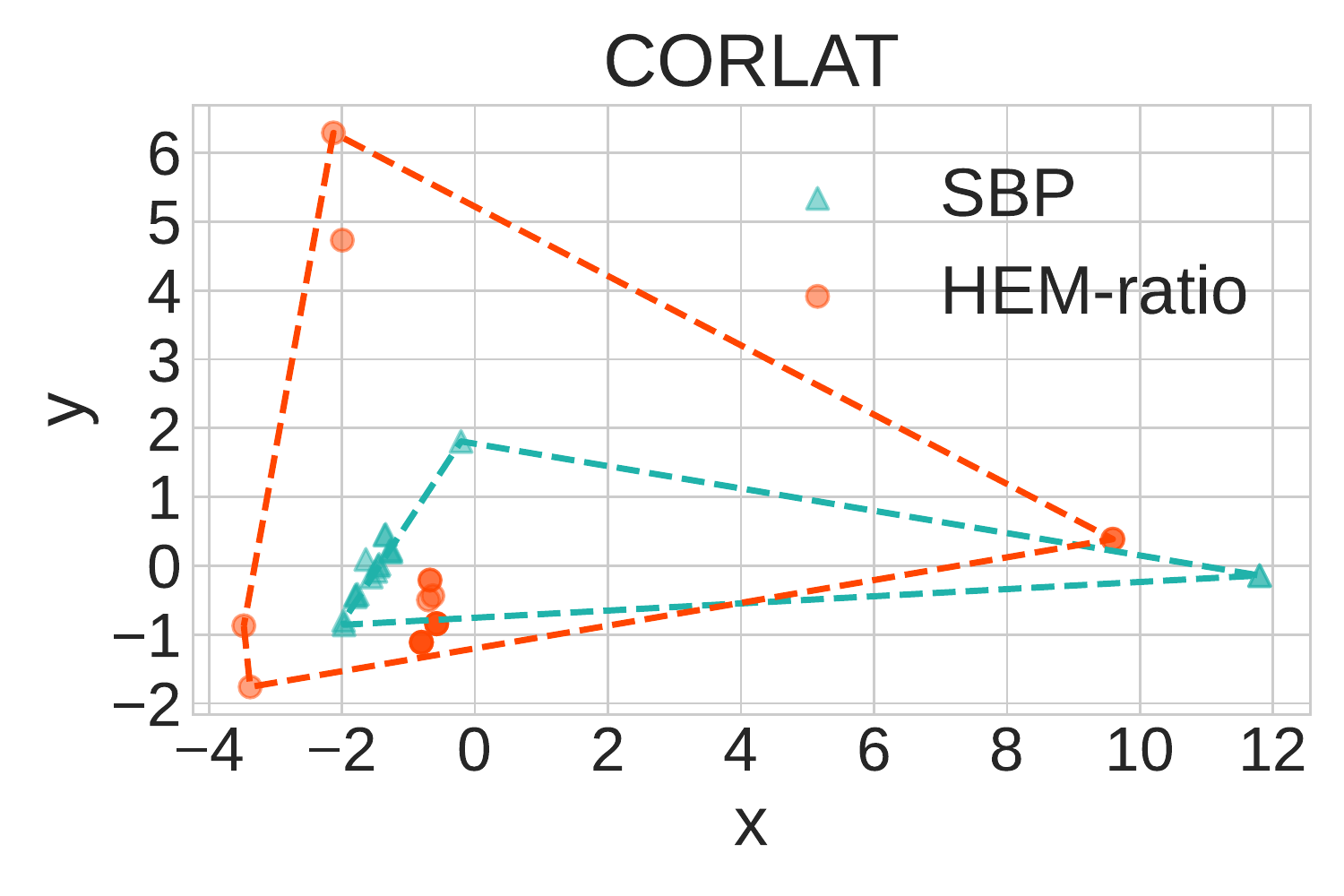}
    \vspace{-7mm}
    \caption{We perform principal component analysis on the cuts selected by HEM-ratio and SBP. Colored points illustrate the reduced cut features. The area covered by the dashed lines represents the diversity of selected cuts. The results show that HEM-ratio selects much more diverse cuts than SBP.}
    \label{fig:visualization}
    \vspace{-1mm}
\end{wrapfigure}

\noindent \textbf{Experiment 4. \mbox{Visualization} of selected cuts}\label{sec:visu} 
    We visualize the diversity of selected cuts, an important metric for evaluating whether the selected cuts complement each other nicely \citep{theoretical_cuts}. We visualize the cuts selected by HEM-ratio and SBP on a randomly sampled instance from Maximum Independent Set and CORLAT, respectively. We \mbox{evaluate} HEM-ratio rather than HEM, as HEM-ratio selects the same number of cuts as SBP. Furthermore, we perform principal component analysis on the \mbox{selected} cuts to reduce the cut features to two-dimensional space. Colored points illustrate the reduced cut features. To visualize the diversity of selected cuts, we use dashed lines to connect the points with the smallest and largest x,y coordinates. That is, \textit{the area covered by the dashed lines represents the diversity}.
    Figure \ref{fig:visualization} shows that SBP tends to select many similar cuts that are possibly redundant, especially on Maximum Independent Set. In contrast, HEM-ratio selects much more diverse cuts that can well complement each other. Please refer to Appendix \ref{appendix_results_visualization} for results on more datasets.


\noindent \textbf{Experiment 5. Deployment in real-world production planning problems}
To further evaluate the effectiveness of our proposed HEM, we deploy HEM to large-scale real-world production planning problems\footnote{We will release the dataset and our code once the paper is accepted to be published.} at an anonymous enterprise, which is one of the largest global commercial technology enterprises.  
Please refer to Appendix \ref{appendix_datasets_sc} for more details of the problems.
The results in Table \ref{generalization_setcovering_mis} (\textbf{Right}) show that HEM significantly outperforms all the baselines in terms of the Time and PD integral. 
The results demonstrate the strong ability to enhance modern MILP solvers with our proposed HEM in real-world applications. Interestingly, Default performs poorer than NoCuts, which implies that an improper cut selection policy could significantly degrade the performance of MILP solvers. 
\vspace{-2mm}
\section{Conclusion}
\vspace{-2mm}
    In this paper, we observe from extensive empirical results that the order of selected cuts has a significant impact on the efficiency of solving MILPs. We propose a novel \textbf{h}ierarchical s\textbf{e}quence \textbf{m}odel (HEM) to learn cut selection policies via reinforcement learning. Specifically, HEM consists of a two-level model: (1) a higher-level model to learn the number of cuts that should be selected, (2) and a lower-level model---that formulates the cut selection task as a sequence to sequence learning problem---to learn policies selecting an ordered subset with the size determined by the higher-level model. Experiments show that HEM significantly improves the efficiency of solving MILPs compared to human-designed and learning-based baselines on both synthetic and large-scale real-world MILPs. We believe that our proposed approach brings new insights into learning cut selection.

\section{Ethics Statement}
    As our work is fundamental research in solving mixed-integer linear programs (MILPs), there are no direct ethical risks. Modern MILP solvers with expert-designed cut selection heuristics aim to solve general MILPs, while modern MILP solvers with our proposed HEM can learn more effective cut selection heuristics from MILPs collected from certain types of real-world applications.

\section{Reproducibility Statement}
    To ensure reproducibility, we describe our proposed HEM in Section \ref{sec_methods} and provide more implementation details in Appendix \ref{appendix_implementation}. Moreover, we provide the pseudocode of our proposed HEM in Appendix \ref{alg_pseudocode}. We will release all the source codes publicly once the paper is accepted. 

    All datasets except the real-world production planning problems used in the experiments are publicly available. Please refer to Appendix \ref{appendix_dataset} for details of these datasets. We will also publicly release the dataset from the real-world production planning problems once the paper is accepted.

\bibliography{iclr2023_conference}
\bibliographystyle{iclr2023_conference}

\newpage
\appendix

\section{Proof}\label{appendix_proof}
    \subsection{Proof of Proposition \ref{proof_hpg}}
    
    \begin{proof}
        The optimization objective takes the form of 
    \begin{align*}
        J(\theta) & = \mathbb{E}_{s\sim \mu, a_k \sim \pi_{\theta}(\cdot|s)}[r(s,a_k)] \\
        & = \mathbb{E}_{s\sim \mu}[ \sum_{a_k} \int_{k=0}^{1} \pi^h_{\theta_1}(k|s) \pi^{l}_{\theta_2}(a_k|s,k) r(s,a_k) dk] \\
        & =  \mathbb{E}_{s\sim \mu}[  \int_{k=0}^{1} \sum_{a_k} \pi^h_{\theta_1}(k|s) \pi^{l}_{\theta_2}(a_k|s,k) r(s,a_k) dk] ,
    \end{align*}
    where $\theta=
    \left[\theta_1,\theta_2\right]$ with $\left[\theta_1,\theta_2\right]$ denoting the concatenation of the two vectors, $\pi_{\theta}(a_k|s) = \mathbb{E}_{k\sim \pi^h_{\theta_1}(\cdot|s)}[\pi^h_{\theta_2}(a_k|s,k)]$, and $\mu$ denotes the initial state distribution. 
    
    We first compute the policy gradient for $\theta_1$:
    \begin{align*}
        & \nabla_{\theta_1} J(\left[\theta_1,\theta_2\right]) \\
        = & \nabla_{\theta_1} (\mathbb{E}_{s\sim \mu}[  \int_{k=0}^{1} \sum_{a_k} \pi^h_{\theta_1}(k|s) \pi^{l}_{\theta_2}(a_k|s,k) r(s,a_k) dk]) \\
        = & \mathbb{E}_{s\sim \mu} [ \nabla_{\theta_1} [\int_{k=0}^{1} \pi_{\theta_1}^h(k|s) \sum_{a_k} \pi_{\theta_2}^l(a_k|s,k) r(s,a_k) dk   ] ].
    \end{align*}
    Let 
    \begin{align*}
        r^h(s,k,\theta_2) & = \sum_{a_k} \pi_{\theta_2}^l(a_k|s,k) r(s,a_k) \\
        & = \mathbb{E}_{a_k\sim \pi_{\theta_2}^l(\cdot|s,k)}[r(s,a_k)],
    \end{align*} then we have that 
    \begin{align*}
        \nabla_{\theta_1} J(\left[\theta_1,\theta_2\right])& = \mathbb{E}_{s\sim\mu}[ \nabla_{\theta_1} [\int_{k=0}^{1} \pi_{\theta_1}^h(k|s) r(s,k,\theta_2) dk] ] \\
        & = \mathbb{E}_{s\sim\mu,k\sim\pi_{\theta_1}^h(\cdot|s)}[\nabla_{\theta_1}\log \pi_{\theta_1}^h(k|s) r(s,k,\theta_2)].
    \end{align*} 
    Therefore, we have that 
    \begin{align*}
        & \nabla_{\theta_1}J(\left[\theta_1,\theta_2\right]) \\
        & = \mathbb{E}_{s\sim\mu, k\sim \pi^h_{\theta_1}(\cdot|s)} [\nabla_{\theta_1} \log(\pi^h_{\theta_1}(k|s)) \mathbb{E}_{a_k \sim \pi^l_{\theta_2}(\cdot|s,k)}[r(s,a_k)] ].
    \end{align*}
    
    We then compute the policy gradient for $\theta_2$:
    \begin{align*}
        & \nabla_{\theta_2}J(\left[\theta_1,\theta_2\right]) \\
        = & \nabla_{\theta_2} (\mathbb{E}_{s\sim \mu}[  \int_{k=0}^{1} \sum_{a_k} \pi^h_{\theta_1}(k|s) \pi^{l}_{\theta_2}(a_k|s,k) r(s,a_k) dk]) \\
        = & \mathbb{E}_{s\sim\mu,k\sim\pi_{\theta_1}^{h}(\cdot|s)}[\nabla_{\theta_2} [\sum_{a_k}\pi_{\theta_2}^l(a_k|s,k)r(s,a_k)] ] \\
        = & \mathbb{E}_{s\sim\mu,k\sim\pi_{\theta_1}^{h}(\cdot|s),a_k\sim\pi_{\theta_2}^l(\cdot|s,k)} [\nabla_{\theta_2}\log \pi_{\theta_2}^{l}(a_k|s,k) r(s,a_k)],
    \end{align*}
    which completes the proof.
    \end{proof}

\section{Related Work}\label{related_work}
\noindent\textbf{Machine learning for MILP.}
    The use of machine learning methods to help improve the MILP solver performance has been an active topic of significant interest in recent years \citep{bengio_ml4co, lodi2017learning, nips21_ml4co_competition, nips19_gcnn}. During the solving process of the solvers, many crucial decisions that significantly impact the solver performance are based on heuristics \citep{scip_thesis}. Recent methods propose to replace these hand-crafted heuristics with machine learning models \citep{bengio_ml4co}. This line of research has shown significant improvement on the solver performance, including cut selection \citep{tang_icml20,l2c_lookahead, adaptive_cut_selection,baltean2019scoring}, variable selection \citep{Khalil_learn_to_branch,nips19_gcnn,pmlr-v80-balcan18a, Parameterizing_branch}, node selection \citep{learning_to_search, node_uct}, column generation \citep{morabit2021machine}, and primal heuristics selection \citep{khalil2017learning,hendel2019adaptive}. In this paper, we focus on cut selection, which plays a significant role in modern MILP solvers \citep{dey2018theoretical, tang_icml20}.
    
    For cut selection, many existing learning-based methods \citep{tang_icml20, l2c_lookahead, cut_ranking} focus on learning which cuts should be preferred by learning a scoring function to measure cut quality. Specifically, \citep{tang_icml20} proposes a reinforcement learning approach to learn to score Gomory cuts \citep{gomory_cuts} and select a Gomory cut with the best scores. Furthermore, \citep{l2c_lookahead} designs a lookahead selection rule which selects a cut that yields the best dual bound improvement, and proposes to learn the expert rule via imitation learning. Instead of selecting the best cut, \citep{cut_ranking} frames cut selection as multiple instance learning to learn a scoring function and selects a fixed ratio of cuts with high scores. However, they neglect the importance of learning how many cuts should be selected. Moreover, we empirically show that the \textit{order of selected cuts} has a large impact on the efficiency of solving MILPs (see Section \ref{sec:order}).

    Moreover, \citep{adaptive_cut_selection} proposes to learn the weightings of four existing scoring rules designed by experts. For the theoretical analysis, \citep{balcan2021sample} provides some provable guarantees for learning cut selection policies. 

\noindent\textbf{Sequence model.}
    Sequence model such as long-short term memory and Transformer has achieved outstanding performance in language tasks such as machine translation 
    \citep{lstm, seq2seq, transformer}. For combinatorial optimization, recent works \citep{pn,neural_combinatorial} propose a variant of the traditional sequence model, namely pointer network, which is applied to directly finding solutions for specific combinatorial optimization problems, such as the Travelling Salesman Problem \citep{tsp}. Instead of finding solutions directly, we propose to use the pointer network for cut selection in modern MILP solvers. To the best of our knowledge, we are the first to apply the pointer network to cut selection, which not only captures the order of selected cuts, but also can well capture the interaction among cuts to select cuts that complement each other nicely.

\section{More Details of Background}
\subsection{More details of the primal-dual gap integral}\label{details_pd_integral}
        We keep track of two important bounds when running branch-and-cut, including the global primal and dual bound.
        The global primal bound corresponds to the value of the best feasible solution found so far, which is the best upper bound of the problem in (\ref{milp1}). The global dual bound corresponds to the minimum dual bound across all leaves of the search tree, which is the best lower bound of the  problem in (\ref{milp1}).
        We define the \textit{primal-dual gap integral} by the area between the curve of the solver's global primal bound and the curve of the solver's global dual bound. 
        With a time limit $T$, we define the primal-dual gap integral by
        \begin{align*}
            \int_{t=0}^{T}( \textbf{c}^T \textbf{x}_t^* - \textbf{z}_t^* ) \mathrm{d}t,
        \end{align*}
        where $\textbf{c}$ is the objective coefficient vector as in (\ref{milp1}), $\textbf{x}_t^*$ is the best feasible solution found at time $t$, $\textbf{z}_t^*$ is the best dual bound at time $t$. 
        We define the \textit{primal-dual gap} by the difference between the global primal bound and the global dual bound. 
        In SCIP 8.0.0 \citep{scip8}, the initial value of the primal-dual gap is set to a constant 100.
        The primal-dual gap integral is a 
        well-recognized metric for evaluating solver performance. For example, the primal-dual gal integral is a primary evaluation metric in the NeurIPS 2021 ML4CO competition \citep{nips21_ml4co_competition}.

\section{Details of the Datasets Used in this Paper}\label{appendix_dataset}
    
    \subsection{The datasets used in the main evaluation}\label{appendix_main_datasets}

    \begin{algorithm}[t]
        \caption{Pseudo code for constructing MIPLIB datasets}
        \label{alg:bfs}
        \begin{algorithmic}[1]
            \STATE \textbf{Input} the initial instance $I_0$, the set of full MIPLIB $\mathcal{M}$, an empty set $\mathcal{M}^{\prime}$, an empty queue $Q$.
            \STATE Initialize $Q$ with the instance $I_0$, $I_0 \to Q$
            \WHILE{Q is not empty}
                \STATE n$=$Q.size()
                \FOR{$i=1,\dots,n$}
                    \STATE Pull an element from $Q$, namely $I^{\prime}$
                    \STATE Compute the similarity scores between each instance in $\mathcal{M}$ except $I^{\prime}$ and $I^{\prime}$ 
                    \STATE Select five instances with the best similarity scores $\mathcal{M}_i$
                    \FOR{$I$ in $\mathcal{M}_i$}
                        \IF{$I$ not in $\mathcal{M}^{\prime}$}
                            \STATE Push $I$ to $\mathcal{M}^{\prime}$; Push $I$ to $Q$
                        \ENDIF
                    \ENDFOR
                \ENDFOR
            \ENDWHILE
            \STATE \textbf{Return} $\mathcal{M}^{\prime}$
        \end{algorithmic}
    \end{algorithm}

    \noindent\textbf{Easy datasets.} The SCIP 8.0.0 solver needs one minute to solve the MILP instances in the easy datasets to optimality. Easy datasets are comprised of three synthetic MILP problems: Set Covering \citep{setcover}, Maximum Independent Set \citep{mis}, and Multiple Knapsack \citep{tree_mdp}.  We choose these three classes of problems for the following reasons. First, they are widely used benchmarks for evaluating MILP solvers \citep{nips19_gcnn, cut_ranking,rl_branch,lookback}. Second, they represent a wide collection of MILP problems encountered in practice. Third, for each class of these problems, the average number of generated cuts is at least twenty, which ensures that proper cut selection strategies are significant for improving the solver performance. Similarly to \citep{nips19_gcnn, tree_mdp, rl_branch, lookback}, we generate set covering instances with 500 rows and 1000 columns, Maximum Independent Set instances with graphs of 500 nodes and affinity set to 4, multiple knapsack instances with 60 items and 12 knapsacks. For each benchmark, we generate a training set of 10,000 instances, and a test set of 100 instances that are never seen during training. Specifically, readers can refer to \url{https://github.com/ds4dm/learn2branch} or  \url{https://github.com/lascavana/rl2branch} for code to generate the easy datasets. We will also release our code once the paper is accepted to be published. 
    
    \noindent\textbf{Medium datasets.} The SCIP 8.0.0 solver needs at least five minutes to solve the instances in the medium datasets to optimality. Following \citet{learning_to_search, mik_corlat, milp_google}, medium datasets comprise MIK \citep{mik}, a set of MILP problems with knapsack constraints, and CORLAT \citep{corlat}, a real dataset used for the construction of a wildlife corridor for grizzly bears in the Northern Rockies region. Each problem set is split into training and test sets with $80\%$ and $20\%$ of the instances. Readers can refer to  \url{https://atamturk.ieor.berkeley.edu/data/mixed.integer.knapsack/} for MIK. Readers can refer to \url{https://bitbucket.org/mlindauer/aclib2/src/master/} for CORLAT.

\begin{wraptable}{r}{0.65\textwidth}
    \centering
    \caption{Criteria for removing instances from MIPLIB 2017.}
    \resizebox{0.64\textwidth}{!}{
    \begin{tabular}{c|c}
    \toprule
         Criteria & $\%$ of instances removed \\ \midrule
         Tags: \textit{feasibility}, \textit{numerics}, \textit{infeasible}, \textit{no solution} & $4.5\%$, $17.4\%$, $2.8\%$, $0.9\%$ \\ \midrule
         Presolve longer than 300 seconds under default conditions & $4.8\%$ \\ \midrule
         Solved to optimality at root & $9.9\%$ \\
         \bottomrule
    \end{tabular}
    }
    \label{tab:Criteria}
\end{wraptable}
    \noindent\textbf{Hard datasets.} The SCIP 8.0.0 solver needs at least one hour to solve the instances in the hard datasets to optimality.
    
    \noindent\textbf{(1) Benchmarks from MIPLIB 2017.} 
        Note that MIPLIB 2017 (MIPLIB) \citep{miplibs_2017} contains instances of MILPs across many different application areas and has been used as a long-standing standard benchmark for MILP solvers \citep{milp_google, adaptive_cut_selection, miplibs_2017}. Previous work \citep{adaptive_cut_selection} has shown that directly learning over the full MIPLIB can be extremely challenging, as these instances are heterogeneous but machine learning has difficulty in learning from heterogeneous datasets. Despite this challenge, we take the first step towards learning over 
        subsets of MIPLIB. Specifically, we construct two subsets by selecting similar instances from MIPLIB. We measure the similarity between instances by 100 human-designed instance features \citep{miplibs_2017}. 
        Following \citet{adaptive_cut_selection}, we first discard instances from MILLIB that satisfy any of the criteria in Table \ref{tab:Criteria}. This ensures that a good cut selection policy can significantly improve the dual bound on the remaining instances. Note that we only use three of seven criteria that are used in \citep{adaptive_cut_selection} to preserve as many instances as possible. 
        
        To select similar instances from MIPLIB 2017, we first choose a representative instance with knapsack constraints (neos-1456979), and a representative instance with set covering constraints (supportcase40). 
        Then we construct the dataset MIPLIB mixed neos following the procedure in Algorithm \ref{alg:bfs} with the initial instance neos-1456979. We construct the dataset MIPLIB mixed supportcase following the procedure in Algorithm \ref{alg:bfs} with the initial instance supportcase40. Note that We measure the similarity between instances by 100 human-designed instance features \citep{miplibs_2017}. Each dataset is split into training and test sets with $80\%$ and $20\%$ of the instances.
        
        Specifically, MIPLIB mixed neos contains 20 instances: neos-1456979, 
        ic97\_tension, icir97\_tension, l2p12, lectsched-4-obj, lectsched-5-obj, loopha13, neos-686190, neos-2294525-abba, neos-3009394-lami, neos-3046601-motu, neos-3046615-murg, neos-3610173-itata, neos-4338804-snowy, neos-5221106-oparau, neos-5260764-orauea, neos-5261882-treska, neos-5266653-tugela, neos16, and timtab1CUTS. 
        
        Moreover, MIPLIB mixed supportcase contains 40 instances: supportcase40,  30\_70\_45\_05\_100, 30\_70\_45\_095\_100, acc-tight2, acc-tight4, acc-tight5, comp07-2idx, comp08-2idx, comp12-2idx, comp21-2idx, decomp1, decomp2, gus-sch, istanbul-no-cutoff, mkc, mkc1, neos-555343, neos-555424, neos-738098, neos-872648, neos-933562, neos-933638, neos-933966, neos-935234, neos-935769, neos-983171, neos-1330346, neos-1337307, neos-1396125, neos-3209462-rhin, neos-3755335-nizao, neos-3759587-noosa, neos-4300652-rahue,
        neos18, physiciansched6-1, physiciansched6-2, piperout-d27, qiu, reblock354, and supportcase37.

    \begin{table*}[t]
    \caption{The statistical description of used datasets. In all datasets, $m$ denotes the average number of constraints and $n$ denotes the average number of variables. Inference Time denotes the inference time of our proposed HEM given the average number of candidate cuts.}
    \vspace{-2mm}
    \label{datasets}
    \centering
    \resizebox{\textwidth}{!}{
    \begin{tabular}{@{}cccccccccc@{}}
    \toprule
    \toprule
    Datasets & Set Covering & Maximum Independent Set & Multiple Knapsack & MIK & CORLAT & Load Balancing & Anonymous & MIPLIB mixed neos & MIPLIB mixed supportcase \\ \midrule
    $m$ & 500 & 1953 & 72 & 346 & 486 & 64304 & 49603 & 5660 & 19910 \\
    $n$ & 1000 & 500 & 720 & 413 & 466 & 61000 & 37881 & 6958 & 19766 \\
    Avg. Candidate Cuts & 780.51 $\pm$ 289.92 & 57.04 $\pm$ 15.53 & 45.00 $\pm$ 12.71 & 62.00 $\pm$ 13.1 & 60.00 $\pm$ 33.29 & 392.53 $\pm$ 32.92 & 79.40 $\pm$ 72.64 & 239.00 $\pm$ 154 & 173.25 $\pm$ 267.27 \\ \midrule
    Inference Time (s) & 1.58 & 0.11 & 0.09 & 0.12 & 0.12 & 0.77 & 0.15 & 0.47 & 0.34 \\ \bottomrule
    \end{tabular}
    }
    \end{table*}

        \textbf{(2) Benchmarks used in NeurIPS 2021 ML4CO competition} The Load Balancing and Anonymous problems used in the main text are from the NeurIPS 2021 ML4CO competition \citep{nips21_ml4co_competition}. 
        Readers can refer to \url{https://www.ecole.ai/2021/ml4co-competition/} for details of the competition. The competition releases three challenging datasets, but we only use two of the three datasets. The major reason is that the average number of the candidate cuts on the instances from the third dataset (Item Placement) is less than five, which makes cut selection has little impact on the overall solver performance.  
    
    \subsubsection{Detailed description of the aforementioned datasets}
        In this part, we provide detailed description of the aforementioned datasets. 
        Note that all datasets we use except MIPLIB 2017 are application-specific, i.e., they contain instances from only a single application. We summarize the statistical description of the used datasets in this paper in Table \ref{datasets}. Let $n,m$ denote the average number of variables and constraints in the MILPs. Let $m\times n$ denote the size of the MILPs. We emphasize that the largest size of our used datasets is up to two orders of magnitude larger than that used in previous learning-based cut selection methods \citep{tang_icml20, l2c_lookahead}, which demonstrates the superiority of our proposed HEM. Moreover, we test the inference time of our proposed HEM given the average number of candidate cuts. The results in Table \ref{datasets} show that the computational overhead of the HEM is very low.

    \subsection{Datasets used in Section \ref{sec:order} in the main text}\label{datasets_motivating}
    In Figure \ref{fig:order_randomall} in the main text, we use five challenging datasets, namely D1, D2, D3, D4, and D5, respectively. Specifically, D1 represents MIPLIB mixed supportcase, D2 represents the single instance neos-1456979 from MIPLIB 2017, D3 represents MIPLIB mixed neos, D4 represents Anonymous, and D5 represents the single instance lectsched-5-obj from MIPLIB 2017. In Figure \ref{fig:order_randomnv} in the main text, we use the dataset MIPLIB mixed neos.

    \subsection{Large-scale real-world production planning problems}\label{appendix_datasets_sc}
    
    The production planning problem aims to find the optimal production planning for thousands of factories according to the daily order demand. The constraints include the production capacity for each production line in each factory, transportation limit, the order rate, etc. The optimization objective is to minimize the production cost and lead time simultaneously. We split the dataset into training and test sets with $80\%$ and $20\%$ of the instances. The average size of the production planning problems is approximately equal to $3500\times 5000=1.75\times10^8$, which are large-scale real-world problems. To promote the machine learning community for MILP, we will release the dataset once the paper is accepted to be published. 

\begin{figure}[t]
    \centering
    \includegraphics[width=0.95\columnwidth]{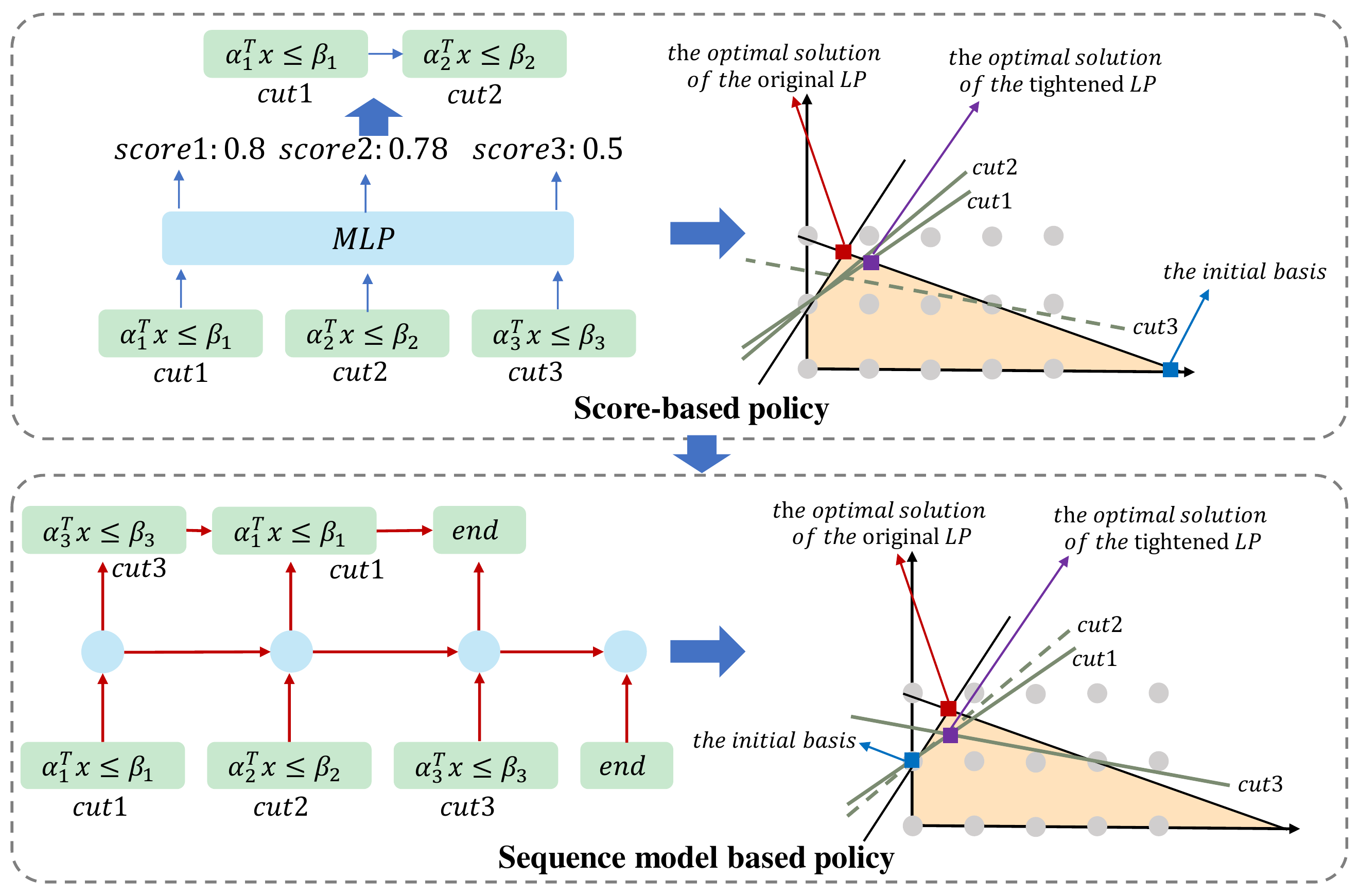}
    \vspace{-3mm}
    \caption{Illustration of selecting cuts using a sequence to sequence model compared to using a scoring function. The sequence model has two main advantages. 
    First, it captures the interaction among cuts by selecting cuts one by one. Consequently, it selects cut3 and cut1 that complement each other nicely, leading to more tightened LP relaxation. Second, it naturally captures the order of selected cuts. Better order of selected cuts may lead to a better initial basis, thus solving the LP relaxation faster \citep{Reformulate} (see Section \ref{sec:order}).}
    \label{fig:illustration_sequece_model}
\end{figure}

 \section{Illustration of Advantages of Using a Sequence Model}\label{appendix_fig_sequence_model}
    Figure \ref{fig:illustration_sequece_model} illustrate two major advantages of using the sequence model to learn cut selection. First, the sequence model takes into account the order of selected cuts by modeling the selected cuts as an output sequence. As shown in Figure \ref{fig:illustration_sequece_model}, the order of cuts determined by the sequence model is better than the score-based method, thus leading to a better initial basis for solving the LP relaxation faster.  
    Second, the sequence model captures the interaction among cuts, as it models the \textit{joint} conditional probability of the selected cuts given an input sequence of the candidate cuts. As shown in Figure \ref{fig:illustration_sequece_model}, the sequence model selects cuts that complement each other nicely, thus leading to a more tightened LP relaxation and speeding up solving the MILP.

\section{Algorithm Implementation and Experimental Settings}\label{appendix_implementation}

\subsection{Designed cut features}\label{appendix_cut_features}

\begin{wraptable}{r}{0.68\textwidth}
    \centering
    \caption{The designed cut features of a generated cut $\boldsymbol{\alpha}^T \textbf{x} \leq \beta$. (Suppose $\textbf{c}$ denotes the objective coefficient.)}
    \resizebox{0.67\textwidth}{!}{
    \begin{tabular}{c|c|c}
    \toprule
         Feature & Description & Number  \\ \midrule
         cut coefficients & the mean, max, min, std of cut coefficients & 4 \\ \midrule
         objective coefficients & the mean, max, min, std of the objective coefficients & 4 \\
         \midrule
         parallelism & the parallelism between the objective and the cut $\frac{\textbf{c}^T \boldsymbol{\alpha}}{|\textbf{c}| | \boldsymbol{\alpha} |}$ & 1 \\\midrule
         efficacy & the Euclidean distance of the cut hyperplane to the current LP solution & 1 \\ \midrule
         support & the proportion of non-zero coefficients of the cut & 1 \\ \midrule
         integral support & the proportion of non-zero coefficients with respect to integer variables of the cut & 1 \\ \midrule
         normalized violation & the violation of the cut to the current LP solution $\max \{0, \frac{\boldsymbol{\alpha}^T\textbf{x}^*_{\text{LP}} - \beta}{|\beta|}\}$ & 1 \\
         \bottomrule
    \end{tabular}
    }
    \label{tab:cut_features}
\end{wraptable}
Following \citet{cut_ranking, implementing_cutting,theoretical_cuts,scip_thesis}, we design thirteen cut features for the cut selection task, such as the extent to which a cut is violated by the current LP solution and the proportion of non-zero coefficients of a cut. We present a detailed description of the designed cut features in Table \ref{tab:cut_features}. We emphasize that we do not tune the cut features. Therefore, it is promising to further improve our method by designing better cut features or using graph neural networks to learn better features in future work.

\subsection{Implementation details of the baselines}\label{appendix_imple_baselines}

In this part, we present a detailed description of all the baselines used in this paper. We denote a cut by $\boldsymbol{\alpha}^T \textbf{x} \leq \beta$ and the optimal solution of the current LP relaxation by $\textbf{x}^*$. Throughout all experiments, we set the ratio of selected cuts as $0.2$ for all score-based rules and learning baselines.

\noindent \textbf{Random.} Random selects a fixed ratio of the candidate cuts stochastically. The ratio is set as $0.2$ in this paper. 
    
\noindent \textbf{Normalized Violation (NV).} NV is a score-based rule. It scores each cut based on the normalized violation of the cut to the current LP solution, and selects a fixed ratio of cuts with high scores. The normalized violation is defined by $\max \{0, \frac{\boldsymbol{\alpha}^T\textbf{x}^*_{\text{LP}} - \beta}{|\beta|}\}$. The ratio is set as $0.2$ in this paper. 

\noindent \textbf{Efficacy (Eff).} Eff is a score-based rule. It scores each cut based on the Euclidean distance of the cut hyperplane to the current LP solution, and selects a fixed ratio of cuts with high scores. The ratio is set as $0.2$ in this paper. 

\noindent \textbf{Default.} Default is the default cut selection rule used in SCIP 8.0.0 \citep{scip8}. Please refer to \citep{scip_thesis} for a detailed description of the SCIP's default cut selection rule. Note that Default tackles the two problems: (1) which cuts should be preferred, and (2) how many cuts should be selected, in cut selection by human-designed heuristics. That is, Default selects variable ratios of cuts rather than a fixed ratio.

\noindent \textbf{Score-based policy (SBP).} Since the state-of-the-art (SOTA) reinforcement learning based method for cut selection \citep{tang_icml20} is designed for the setting that selects the best cut in each round, we implement a slight variant of the SOTA to adapt to our setting that selects a subset of cuts in each round, namely SBP. Specifically, the core idea of SBP is learning a scoring function to measure cut quality as \citet{tang_icml20, cut_ranking, l2c_lookahead} do. For a fair comparison, SBP uses the same cut features as HEM and we train SBP via reinforcement learning as well. Our implemented SBP is also a slight variant of the method proposed in \citet{cut_ranking}. We emphasize that experiments in the main text show that our implemented SBP is a strong baseline. Specifically, we implement the scoring function with a multi-layer perceptron that predicts the score of a given cut. That is, the scoring function predicts a cut's score based on the features of the cut. The MLP network contains two hidden layers with 128 units. Moreover, we train the scoring function via evolutionary strategies as \citep{tang_icml20} does. We will also release the code for implementing SBP once the paper is accepted to be published.  

\subsection{Implementation details and hyperparameters }\label{appendix_imple_hyper}
\subsubsection{Hardware specification}
    Throughout all experiments, we use a single machine that contains eight GPU devices (NVidia GeForce GTX 3090 Ti) and two Intel Gold 6246R CPUs. 

\subsubsection{Solver setup}
    For reproducibility, we emphasize that all results in the main text are obtained by averaging results over the SCIP random seeds $\{1,2,3\}$.

\begin{wraptable}{r}{0.6\textwidth}
\centering
\caption{Evaluation on real-world production planning problems with rewards being the negative PD integral. The results show that HEM still significantly outperforms all the baselines.}
\label{tab:pp_pd_integral}
\resizebox{0.59\textwidth}{!}{
\begin{tabular}{@{}ccccc@{}}
\toprule
\toprule
 & \multicolumn{4}{c}{Production planning} \\ \midrule
\multirow{2}{*}{Method} & \multirow{2}{*}{Time (s)} & \multirow{2}{*}{Improvement (Time, \%)} & \multirow{2}{*}{PD integral} & \multirow{2}{*}{\begin{tabular}[c]{@{}c@{}}Improvement\\      (PD integral, \%)\end{tabular}} \\
 &  &  &  &  \\ \midrule
NoCuts & 278.79 & NA & 17866.01 & NA \\
Default & 296.12 & -6.22 & 17703.39 & 0.91 \\
Random & 280.18 & -0.50 & 18120.21 & -1.42 \\
NV & 259.48 & 6.93 & 17295.18 & 3.20 \\
Eff & 263.60 & 5.45 & 16636.52 & 6.88 \\ \midrule
SBP & 276.61 & 0.78 & 16952.85 & 5.11 \\
HEM (Ours) & \textbf{251.64} & \textbf{9.74} & \textbf{16533.05} & \textbf{7.46} \\ \bottomrule
\end{tabular}
}
\end{wraptable}
\subsubsection{Reward function}
    On the easy datasets, we set the reward as the negative solving time. On the medium and hard datasets, we set the reward as the negative primal-dual gap integral within a time limit of 300 seconds. 
    
    For the real-world production planning problems, we set the reward as the negative primal-dual gap integral within a time limit of 600 seconds or the negative dual bound improvement. The results reported in the main text are achieved by HEM with the negative dual bound improvement reward. We provide the performance of HEM that uses the negative primal-dual gap integral in Table \ref{tab:pp_pd_integral}. The results still show that HEM significantly outperforms all the baselines in terms of the Time and PD integral. 

    We emphasize that we can set the reward according to our objective in real-world problems. For example, suppose we aim to minimize the primal-dual gap within a time limit, then we can set the reward as the primal-dual gap within the time limit.

\subsubsection{Policy network architecture}
    The higher-level model contains an LSTM encoder and an MLP. The LSTM network encodes variable-sized inputs
    into hidden vectors with dimension 128. The MLP network contains two hidden layers with 128 units. The lower-level model is essentially a pointer network. We keep the hyperparameters of the pointer network as that used in \citep{neural_combinatorial}. 

\begin{algorithm}[t]
    \caption{Pseudo code for training the HEM}.
    \label{alg:hem}
    \begin{algorithmic}[1]
        \STATE \textbf{Initialize} Hierarchical sequence model $\pi_{[\theta_1,\theta_2]}$, MILP instances $\mathcal{D}$, training dataset $\mathcal{D}_{\text{train}}$, batch size $N_b$, training epochs $N_e$, policy learning rate $\alpha$
        \FOR{$N_e$ epochs}
            \STATE Empty the training dataset $\mathcal{D}_{\text{train}}$
            \FOR{$N_b$ steps}
            \STATE Randomly sample a MILP $s_0$ from $\mathcal{D}$
            \STATE Take action $k$ and $a_k$ at state $s_0$ with the policy $\pi$
            \STATE Receive reward $r$ and add $(s_0,k,a_k,r)$ to  $\mathcal{D}_{\text{train}}$
            \ENDFOR
            \STATE Compute hierarchical policy gradient using $\mathcal{D}_{\text{train}}$ as in proposition \ref{proof_hpg}
            \STATE Update the parameters, $\theta_1 = \theta_1 + \alpha \nabla_{\theta_1}J(\left[\theta_1,\theta_2\right])$, $\theta_2 = \theta_2 + \alpha \nabla_{\theta_2}J(\left[\theta_1,\theta_2\right])$
        \ENDFOR
    \end{algorithmic}
\end{algorithm}

\subsubsection{Optimization}
    Throughout all experiments, we apply Adam optimizer with learning rate $\alpha_1=1\times10^{-4}$ to optimize the lower-level model, and learning rate $\alpha_2=5\times10^{-4}$ to optimize the higher-level model. For each epoch, we collect 32 samples for training, and we set the total epochs as 100. It is surprising that learning a good cut selection policy does not need too much data as shown in \citet{tang_icml20}. 
    For training stability, we delay the higher-level policy update. This creates a two-timescale algorithm, as often required for convergence in the linear setting \citep{td3, konda2003actor}. We set the delay update freq as two. That is, we first train the lower-level policy twice, then train the higher-level policy once. Additionally, the results in Appendix \ref{appendix_delay} show that the convergence performance of HEM is insensitive to the hyperparameter delay update freq.

\subsubsection{The training algorithm}\label{alg_pseudocode}
    We provide the procedure of the training algorithm of HEM in Algorithm \ref{alg:hem}. 

\subsection{More details of HEM}

\begin{figure}[t]
    \centering
    \includegraphics[width=0.8\textwidth]{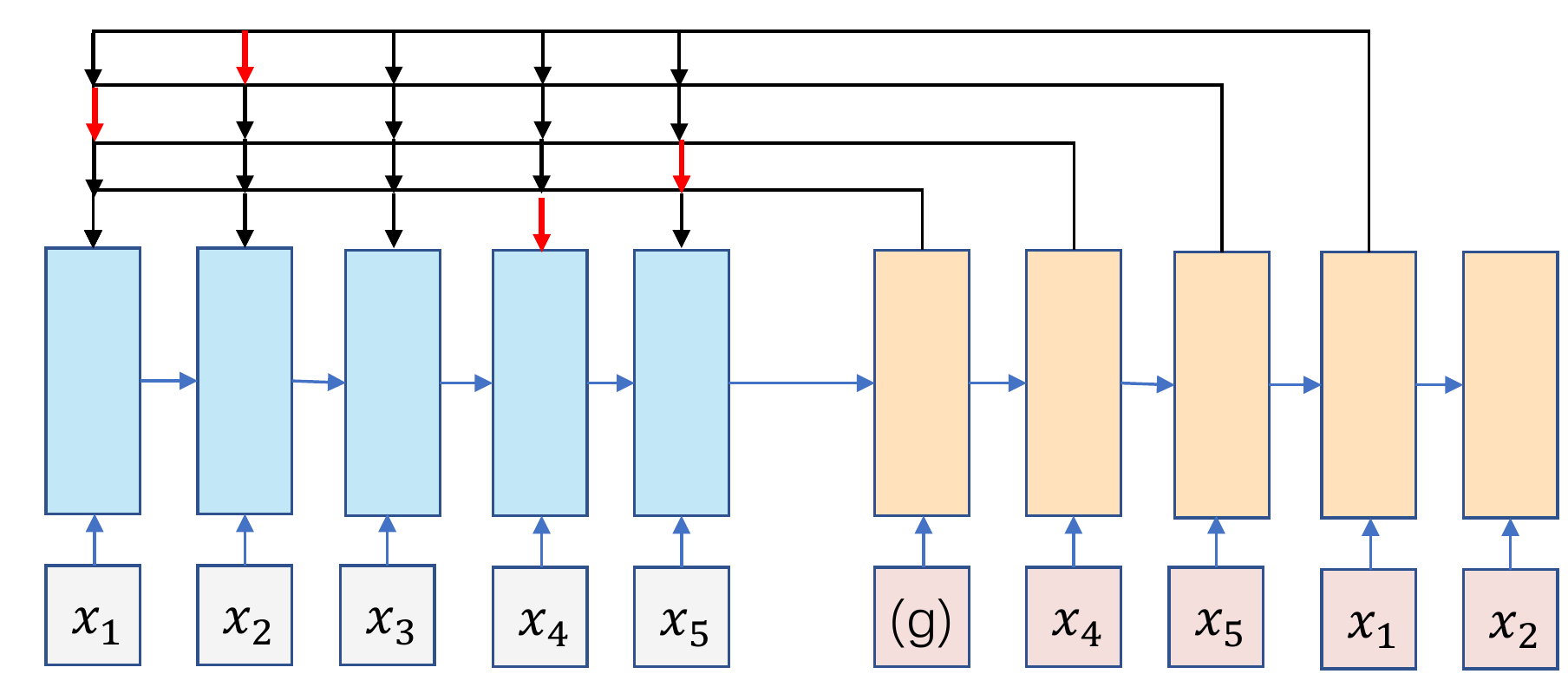}
    \caption{Illustration of the pointer network architecture introduced by \citet{pn}. }
    \label{fig:pn}
\end{figure}

    \subsubsection{Details of the pointer network}\label{appendix_pn_details}
        The pointer network is first introduced by \citep{pn} for directly finding solutions of specific combinatorial optimization problems, such as the Travelling Salesman Problems. The pointer network architecture is illustrated in Figure \ref{fig:pn}.
        The pointer network consists of a Long Short-Term Memory encoder, a Long Short-Term Memory decoder, and an attention that is used as a pointer to select a member of the input sequence as the output \citep{pn}. Specifically, we implement the pointer network following \citet{neural_combinatorial}. Please refer to \citep{neural_combinatorial} for implementation details of the pointer network. 
        
        The major difference between our used pointer network and the pointer network used in \citep{neural_combinatorial} is that we use the pointer network to select ordered subsets of input sequences, but \citep{neural_combinatorial} use the pointer network to output permutations of input sequences.

    \subsubsection{Difference between HEM-ratio and HEM w/o H}\label{appendix_imple_rlk}
        \noindent \textbf{Details of HEM w/o H} To implement HEM w/o H, we augment each input sequence with an end token, i.e., a thirteen-dimensional tensor with values all being one. The end token is at the end position of the input sequence. Once the decoder of HEM w/o H outputs the end token, then the decoding ends. That is, HEM w/o H can implicitly predict the number of cuts that should be selected by predicting whether to decode the end token at the current decoding step. 
            
        The policy network of HEM-ratio and HEM w/o H are both essentially a pointer network \citep{pn}, a variant of the sequence model. We present the major difference between HEM-ratio and HEM w/o H in the following. HEM w/o H predicts an end token as used in language tasks \citep{seq2seq,transformer} to determine the number of cuts that should be selected implicitly. In contrast, HEM-ratio always selects a fixed ratio of cuts, i.e, it always ends decoding at a pre-determined position. Therefore, HEM w/o H can learn the number of cuts that should be selected, but HEM-ratio cannot. 
        
    \subsubsection{More discussion of HEM}\label{hem_advantages}
        In this part, we provide details of some more advantages of HEM. (1) Inspired by hierarchical reinforcement learning \citep{smdp_sutton, hiro}, HEM leverages the hierarchical structure of the cut selection task, which is important for efficient exploration in complex decision-making tasks. (2) Previous methods  \citep{tang_icml20, cut_ranking} usually train cut selection policies via black-box optimization methods such as evolution strategies \citep{es}. In contrast, HEM is differentiable and we train the HEM via gradient-based algorithms, which is more sample efficient than black-box optimization methods \citep{rl_sutton,trpo}. Although we can offline generate training samples as much as possible using a MILP solver, high sample efficiency is significant as generating samples can be extremely time-consuming in practice. 

\section{More Results}\label{appendix_results}

\begin{table*}[t]
\caption{Policy evaluation on easy, medium, and hard datasets in terms of the total number of nodes and the primal-dual gap. The best performance are marked in bold. }
\vspace{-2mm}
\label{evaluation_all_more_metrics}
\centering
\resizebox{0.96\textwidth}{!}{
\begin{tabular}{@{}ccccccc@{}}
\toprule
\toprule
 & \multicolumn{2}{c}{Set Covering} & \multicolumn{2}{c}{Maximum Independent Set} & \multicolumn{2}{c}{Multiple Knapsack} \\ \midrule
Method & Nodes & PD gap & Nodes & PD gap & Nodes & PD gap \\ \cmidrule(r){1-3} \cmidrule(lr){4-5} \cmidrule(l){6-7}
NoCuts & 189.44 (423.68) & 0.00 (0) & 2170.66 (4054.09) & 0.00 (0) & 16945.58 (41242.04) & 0.00 (0.000128) \\
Default & 116.77 (420.09) & 0.00 (0) & 588.23 (1916.99) & 0.00 (0) & 16949.88 (41297.50) & 0.00 (0.000128) \\
Random & 95.70 (285.05) & 0.00 (0) & 1416.96 (3820.01) & 0.00 (0) & 21463.16 (59411.07) & 0.00 (0.000361) \\
NV & 199.70 (436.25) & 0.00 (0) & 1618.42 (3089.12) & 0.00 (0) & 20673.32 (62526.13) & 0.00 (0.00022) \\
Eff & 194.17 (439.35) & 0.00 (0) & 1575.88 (2742.66) & 0.00 (0) & 14909.93 (28575.12) & 0.00 (0) \\ \midrule
SBP & 1.16 (2.04) & 0.00 (0) & 698.41 (2869.65) & 0.00 (0) & 13537.59 (22527.45) & 0.00 (0) \\
HEM (Ours) & \textbf{1.11 (1.46)} & \textbf{0.00 (0)} & \textbf{311.64 (1309.94)} & \textbf{0.00 (0)} & \textbf{10463.85 (18491.25)} & \textbf{0.00 (0)} \\ \bottomrule
\end{tabular}
}
\newline
\vspace{2mm}
\newline
\resizebox{0.96\textwidth}{!}{
\begin{tabular}{@{}ccccccc@{}}
\toprule
\toprule
 & \multicolumn{2}{c}{MIK} & \multicolumn{2}{c}{CORLAT} & \multicolumn{2}{c}{Load Balancing} \\ \midrule
Method & Nodes & PD gap & Nodes & PD gap & Nodes & PD gap \\ \cmidrule(r){1-3} \cmidrule(lr){4-5} \cmidrule(l){6-7}
NoCuts & 395290.17 (120526.21) & 0.09 (0.018) & 97098.72 (119287.56) & 2.000000e+18 (1.4E+19) & 108.20 (56.28) & 0.94 (0.12) \\
Default & 224746.60 (179450.80) & 0.02 (0.033) & 70215.01 (113515.62) & 2.666667e+18 (1.61E+19) & \textbf{65.22 (55.34)} & 0.43 (0.073) \\
Random & 403618.10 (115802.29) & 0.07 (0.035) & 74149.90 (109043.28) & 2.000000e+18 (1.4E+19) & 107.05 (68.05) & 0.79 (0.133) \\
NV & 397025.47 (106676.73) & 0.08 (0.028) & 77943.73 (111355.03) & 4.666667e+18 (2.11E+19) & 81.91 (38.55) & 0.82 (0.11) \\
Eff & 406832.83 (113363.27) & 0.07 (0.033) & 95956.53 (120858.92) & 5.333333e+18 (2.25E+19) & 86.22 (45.35) & 0.82 (0.11) \\ \midrule
SBP & 417070.37 (133702.50) & 0.07 (0.030) & 62297.99 (109664.21) & 6.666667e+17 (8.14E+18) & 138.37 (53.79) & 0.65 (0.06) \\
HEM(Ours) & \textbf{220547.93 (172537.94)} & \textbf{0.02 (0.028)} & \textbf{51929.71 (100973.74)} & \textbf{0.02 (0.051)} & 74.47 (60.98) & \textbf{0.42 (0.072)} \\ \bottomrule
\end{tabular}
}
\newline
\vspace{2mm}
\newline
\resizebox{0.96\textwidth}{!}{
\begin{tabular}{@{}ccccccc@{}}
\toprule
\toprule
 & \multicolumn{2}{c}{Anonymous} & \multicolumn{2}{c}{MIPLIB mixed neos} & \multicolumn{2}{c}{MIPLIB mixed   supportcase} \\ \midrule
Method & Nodes & PD gap & Nodes & PD gap & Nodes & PD gap \\ \cmidrule(r){1-3} \cmidrule(lr){4-5} \cmidrule(l){6-7}
NoCuts & 20110.58 (14167.51) & 8.333333e+18 (2.76E+19) & 169348.83 (117745.23) & 2.500000e+19 (4.33E+19) & 5971.00 (8679.98) & 9.19 (19.29) \\
Default & 20025.78 (13056.50) & 8.333333e+18 (2.76E+19) & 161366.25 (107049.96) & 2.500000e+19 (4.33E+19) & \textbf{4055.04 (7404.61)} & 11.73 (31.52) \\
Random & 19824.48 (12366.55) & 1.000000e+19 (3E+19) & \textbf{143930.17 (95296.41)} & 2.500000e+19 (4.33E+19) & 4656.13 (7745.46) & 6.05 (14.69) \\
NV & \textbf{19313.33 (12391.94)} & 8.333333e+18 (2.76E+19) & 150046.50 (94801.62) & 2.500000e+19 (4.33E+19) & 4986.21 (9024.19) & 2.78 (10.08) \\
Eff & 19526.23 (12116.80) & 6.666667e+18 (2.49E+19) & 144128.83 (93404.76) & 2.500000e+19 (4.33E+19) & 4450.88 (7845.48) & 5.58 (15.40) \\ \midrule
SBP & 19351.67 (12337.81) & 6.666667e+18 (2.49E+19) & 177736.25 (122005.81) & 2.500000e+19 (4.33E+19) & 5618.08 (9765.43) & 0.14 (0.26) \\
HEM(Ours) & 20191.28 (13219.21) & \textbf{1.666667e+18 (1.28E+19)} & 177735.83 (129020.42) & \textbf{2.500000e+19 (4.33E+19)} & 4844.88 (9996.16) & \textbf{0.14 (0.25)} \\ \bottomrule
\end{tabular}
}
\end{table*}

\begin{figure}
    \centering
    \begin{subfigure}{0.4\textwidth}
        \includegraphics[width=\textwidth]{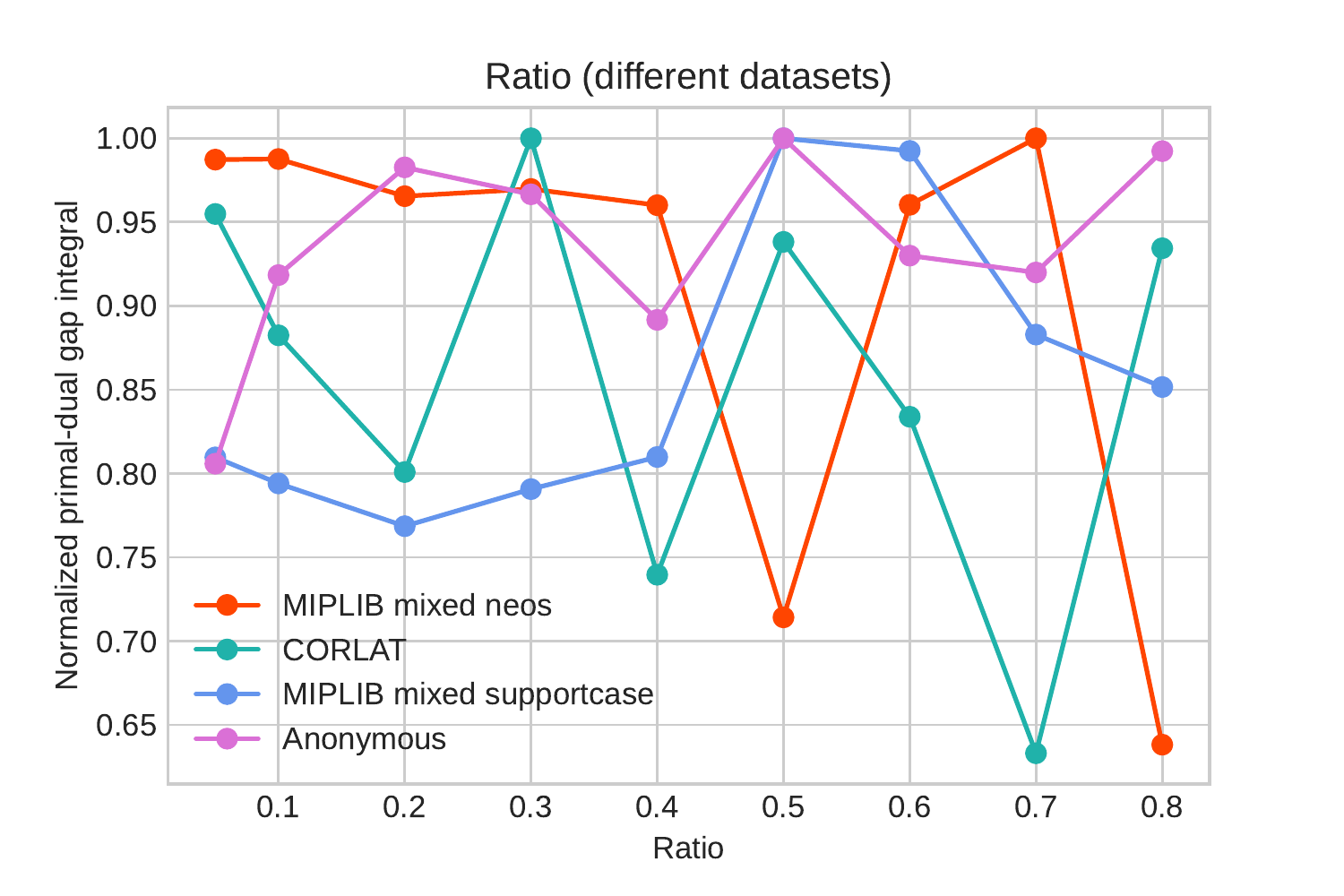}
        \vspace{-5mm}
        \caption{NV with different ratios on four datasets.}
        \label{fig:size_diff_datasets}
    \end{subfigure}
    \begin{subfigure}{0.4\textwidth}
        \includegraphics[width=\textwidth]{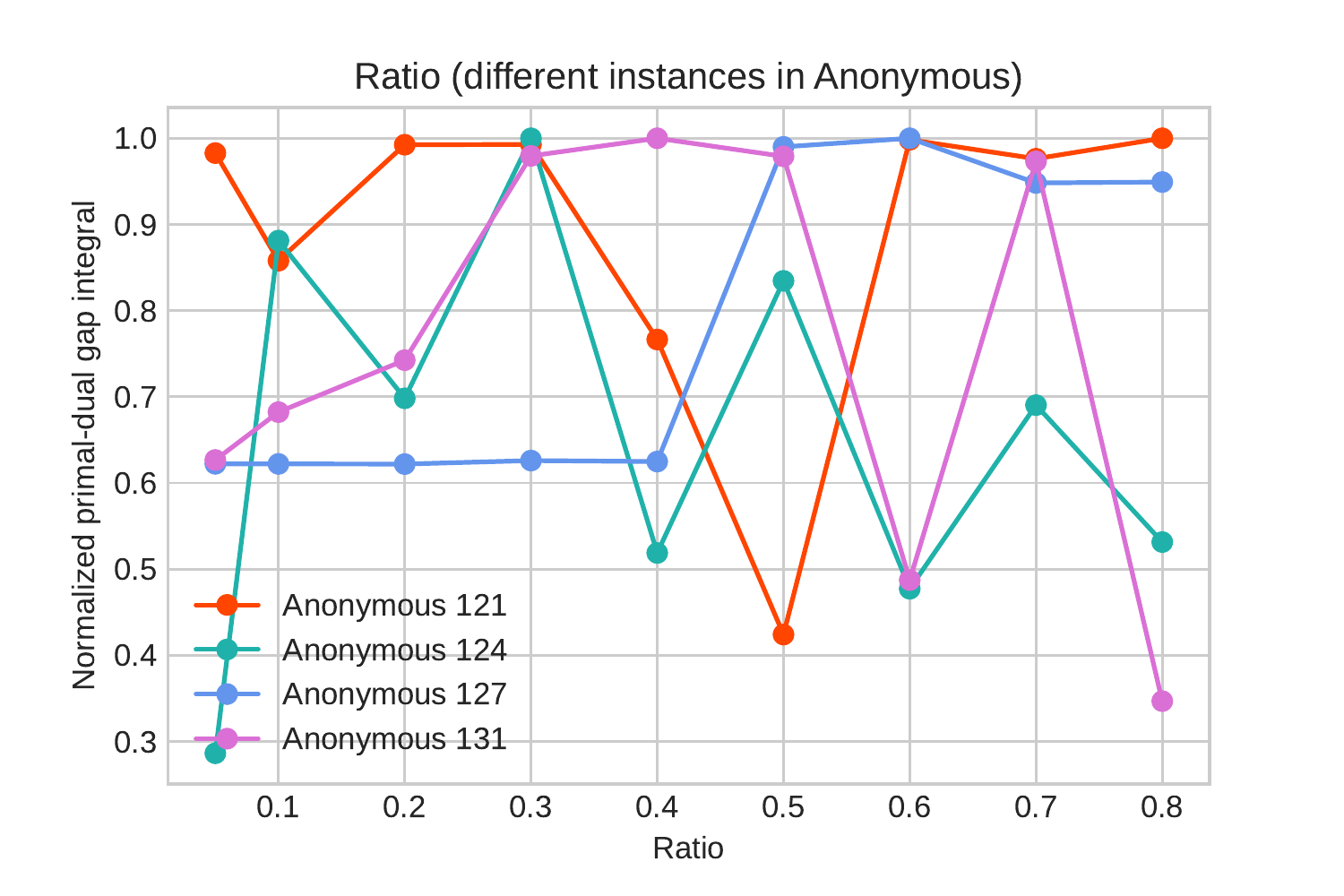}
        \vspace{-5mm}
        \caption{NV with different ratios on Anonymous.}
        \label{fig:size_diff_instances}
    \end{subfigure}
    \vspace{-3mm}
    \caption{ 
        We use the Normalized Violation (NV) rule \citep{cut_ranking}. The results in (c) show that NV with different given ratios achieve variable normalized PD integral on four datasets. The results in (d) show that NV with different ratios achieve variable normalized PD integral on different instances from the same dataset.
    }
    \label{fig:motivating_results_size}
\end{figure}

\subsection{More motivating results}\label{motivating_results_ratio}
\noindent \textbf{Ratio matters.} \label{sec:ratio}
To evaluate the effect of the ratio of selected cuts on solving MILPs, we focus on the Normalized Violation (NV) cut selection method with different ratios of selected cuts. (1) We first evaluate the NV methods that select $5\%, 10\%, 20\%, 30\%, 40\%, 50\%, 60\%, 70\%$, and $80\%$ of candidate cuts, respectively, on four datasets. The results in Figure \ref{fig:size_diff_datasets} show that the NV achieves better solver performance with larger ratios on CORLAT and MIPLIB mixed neos. The results demonstrate that the ratio that leads to better solver performance is variable over different datasets, which implies that learning dataset-dependent ratios is important.
(2) We then evaluate the NV methods that select $5\%, 10\%, 20\%, 30\%, 40\%, 50\%, 60\%, 70\%$, and $80\%$ of candidate cuts, respectively, on four instances from the Anonymous dataset. The results in Figure \ref{fig:size_diff_instances} show that NV achieves better solver performance with larger ratios on Anonymous 121 and Anonymous 131. The results demonstrate that the ratio that leads to better solver performance is variable over different instances from the same dataset, which implies that learning instance-dependent ratios is important as well.     

\subsection{More results of main evaluation}\label{appendix_results_more_metrics}

    In this section, we provide more results of the main evaluation. The results in Table \ref{evaluation_all_more_metrics} show the performance of HEM and the baselines in terms of the total number of nodes (Nodes) and the primal-dual gap (PD gap). (1) \textbf{Easy datasets.} On the easy datasets, HEM and most baselines find the optimal solution within the time limit, as the PD gap converges to zero. Additionally, HEM significantly outperforms all the baselines in terms of 
    the Nodes on the easy datasets. (2) \textbf{Medium and hard datasets.} In terms of the PD gap, HEM outperforms all the baselines on medium and hard datasets, especially on CORLAT. However, the Nodes metric cannot well distinguish the performance of different cut selection methods on the medium and hard datasets for the following two reasons. First, the solving time of the LP relaxation on each node is different, and thus the Nodes cannot directly determine the solving time \citep{cut_ranking}. Second, on those unsolved instances within the time limit, the Nodes metric is not a proper metric, as the Nodes cannot evaluate the quality of the solving process. 

\subsection{More results of ablation study}\label{appendix_results_ablation_study}

In this section, we provide more results of ablation studies in the main text. 

\subsubsection{In-depth analysis of HEM-ratio and SBP}\label{appendix_analysis_hem_ratio}
    We provide possible reasons for HEM-ratio performing poorer than SBP on several challenging MILP problem benchmarks. 
    Fundamentally, HEM-ratio formulates the cut selection task as a sequence modeling problem, which has two main advantages over SBP. That is, the sequence model can not only capture the underlying order information, but also capture the interaction among cuts. However, training a sequence model is more difficult than training a scoring function, as the sequence model aims to learn a much more complex task. Specifically, the scoring function aims to learn to score each cut, while the sequence model aims to model the joint probability of the selected cuts. The latter is a more challenging learning task. Moreover, we follow the reinforcement learning paradigm instead of supervised learning to train the model, making the training process more unstable. Therefore, the sequence model may suffer from inefficient exploration and be trapped to a local optimum. As a result, HEM-ratio can perform poorer than SBP, especially on challenging MILP problem benchmarks. 

\subsubsection{Contribution of each component}
    To understand the contribution of each component of HEM, we provide more results of HEM and HEM without the higher-level model on Set Covering, Multiple Knapsack, MIK, Load Balancing, Anonymous, and MIPLIB mixed supportcase. The results in Table \ref{ablation_component_more_datasets} show that HEM outperforms HEM w/o H in terms of the solving time, the primal-dual gap, and the primal-dual gap integral on several challenging datasets, demonstrating the importance of our proposed higher-level model. Moreover, HEM w/o H significantly outperforms SBP in terms of the solving time, the primal-dual gap, and the primal-dual gap integral on several challenging datasets, demonstrating the significance of our proposed lower-level model.  

\begin{table*}[t]
\caption{Comparsion between HEM, HEM without the higher-level model on more datasets.}
\vspace{-2mm}
\label{ablation_component_more_datasets}
\centering
\resizebox{0.96\textwidth}{!}{
\begin{tabular}{@{}ccccccccc@{}}
\toprule
\toprule
 & \multicolumn{4}{c}{Set Covering} & \multicolumn{4}{c}{Multiple Knapsack} \\ \midrule
Method & Time (s) & Nodes & PD gap & PD integral & Time (s) & Nodes & PD gap & PD integral \\ \cmidrule(r){1-5}  \cmidrule(l){6-9}
NoCuts & 6.31 (4.61) & 189.44 (423.68) & 0.00 (0) & 56.99 (38.89) & 9.88 (22.24) & 16945.58 (41242.04) & 0.00 (0.000128) & 16.41 (14.16) \\
Default & 4.41 (5.12) & 116.77 (420.08) & 0.00 (0) & 55.63 (42.21) & 9.90 (22.24) & 16949.87 (41297.49) & 0.00 (0.000128) & 16.46 (14.25) \\
SBP & 1.91 (0.36) & 1.16 (2.04) & 0.00 (0) & 38.96 (8.66) & 7.74 (12.36) & 13537.58 (22527.45) & 0.00 (0) & 16.45 (16.62) \\ \midrule
HEM w/o H & \textbf{1.84 (0.31)} & 1.18 (2.03) & 0.00 (0) & \textbf{37.69 (8.33)} & 7.36 (12.90) & 12418.85 (23256.47) & 0.00 (0) & 14.01 (10.94) \\
HEM (Ours) & 1.85 (0.31) & \textbf{1.09 (1.46)} & \textbf{0.00 (0)} & 37.92 (8.46) & \textbf{6.13 (9.61)} & \textbf{10463.84 (18491.25)} & \textbf{0.00 (0)} & \textbf{13.63 (9.63)} \\ \bottomrule
\end{tabular}
}
\newline
\vspace{2mm}
\newline
\resizebox{0.96\textwidth}{!}{
\begin{tabular}{@{}ccccccccc@{}}
\toprule
\toprule
 & \multicolumn{4}{c}{MIK} & \multicolumn{4}{c}{Load Balancing} \\ \midrule
Method & Time (s) & Nodes & PD gap & PD integral & Time (s) & Nodes & PD gap & PD integral \\ \cmidrule(r){1-5}  \cmidrule(l){6-9}
NoCuts & 300.01 (0.009) & 395290.16 (120526.21) & 0.09 (0.018) & 2355.87 (996.08) & 300.00 (0.12) & 108.20 (56.28) & 0.94 (0.12) & 14853.77 (951.42) \\
Default & 179.62 (122.36) & 224746.60 (179450.80) & 0.02 (0.033) & 844.40 (924.30) & 300.00 (0.06) & \textbf{65.22 (55.34)} & 0.43 (0.07) & 9589.19 (1012.95) \\
SBP & 286.07 (41.81) & 417070.36 (133702.49) & 0.07 (0.0295) & 2053.30 (740.11) & 300.00 (0.10) & 138.37 (53.79) & 0.65 (0.06) & 12535.30 (741.43) \\ \midrule
HEM w/o H & 218.87 (115.97) & 297058.97 (189410.43) & 0.04 (0.039) & 1321.63 (1165.23) & 300.03 (0.04) & 70.93 (60.36) & \textbf{0.42 (0.07)} & \textbf{9475.16 (1005.81)} \\
HEM (Ours) & \textbf{176.12 (125.18)} & \textbf{220547.93 (172537.94)} & \textbf{0.02 (0.028)} & \textbf{785.04 (790.38)} & \textbf{300.00 (0.04)} & 74.47 (60.98) & \textbf{0.42 (0.07)} & \textbf{9496.42 (1018.35)} \\ \bottomrule
\end{tabular}
}
\newline
\vspace{2mm}
\newline
\resizebox{0.96\textwidth}{!}{
\begin{tabular}{@{}ccccccccc@{}}
\toprule
\toprule
 & \multicolumn{4}{c}{Anonymous} & \multicolumn{4}{c}{MIPLIB mixed   supportcase} \\ \midrule
Method & Time (s) & Nodes & PD gap & PD integral & Time (s) & Nodes & PD gap & PD integral \\ \cmidrule(r){1-5}  \cmidrule(l){6-9}
NoCuts & 246.22 (94.90) & 20110.58 (14167.51) & 8.333333e+18 (2.76E+19) & 18297.30 (9769.42) & 170.00 (131.60) & 5971.00 (8679.98) & 9.19 (19.29) & 9927.96 (11334.07) \\
Default & 244.02 (97.72) & 20025.78 (13056.50) & 8.333333e+18 (2.76E+19) & 17407.01 (9736.19) & 164.61 (135.82) & 4055.04 (7404.61) & 11.73 (31.52) & 9672.34 (10668.24) \\
SBP & 245.71 (92.46) & \textbf{19351.66 (12337.81)} & 6.666667e+18 (2.49E+19) & 18188.63 (9651.85) & 165.61 (135.25) & 5618.08 (9765.43) & 0.14 (0.26) & 7408.65 (7903.47) \\ \midrule
HEM w/o H & 251.00 (88.45) & 21192.40 (16436.32) & 5.000000e+18 (2.18E+19) & 17226.84 (9553.90) & \textbf{156.96 (133.35)} & \textbf{3779.83 (7465.75)} & 0.56 (1.43) & 7709.19 (8655.48) \\
HEM (Ours) & \textbf{241.68 (97.23)} & 20191.28 (13219.21) & \textbf{1.666667e+18 (1.28E+19)} & \textbf{16077.15 (9108.21)} & 162.96 (138.21) & 4844.87 (9996.15) & \textbf{0.14 (0.25)} & \textbf{6874.80 (6729.97)} \\ \bottomrule
\end{tabular}
}
\end{table*}

\begin{table*}[t]
\caption{Comparsion between HEM, HEM-ratio, and HEM-ratio-order on more datasets.}
\vspace{-2mm}
\label{ablation_factor_more_datasets}
\centering
\resizebox{0.96\textwidth}{!}{
\begin{tabular}{@{}ccccccccc@{}}
\toprule
\toprule
 & \multicolumn{4}{c}{Set Covering} & \multicolumn{4}{c}{Multiple Knapsack} \\ \midrule
Method & Time (s) & Nodes & PD gap & PD integral & Time (s) & Nodes & PD gap & PD integral \\ \cmidrule(r){1-5}  \cmidrule(l){6-9}
NoCuts & 6.31 (4.61) & 189.44 (423.68) & 0.00 (0) & 56.99 (38.89) & 9.88 (22.24) & 16945.58 (41242.04) & 0.00 (0.000128) & 16.41 (14.16) \\
Default & 4.41 (5.12) & 116.77 (420.08) & 0.00 (0) & 55.63 (42.21) & 9.90 (22.24) & 16949.87 (41297.49) & 0.00 (0.000128) & 16.46 (14.25) \\
SBP & 1.91 (0.36) & 1.16 (2.04) & 0.00 (0) & 38.96 (8.66) & 7.74 (12.36) & 13537.58 (22527.45) & 0.00 (0) & 16.45 (16.62) \\ \midrule
HEM-ratio-order & 2.11 (0.38) & 1.11 (1.29) & 0.00 (0) & 42.01 (9.88) & 10.46 (29.74) & 18434.20 (59254.10) & 0.00 (0.000128) & 16.92 (18.19) \\
HEM-ratio & 2.10 (0.40) & 1.11 (1.18) & 0.00 (0) & 41.95 (9.82) & 7.63 (12.64) & 13307.73 (24339.3) & 0.00 (0) & 16.19 (15.16) \\
HEM (Ours) & \textbf{1.85 (0.31)} & \textbf{1.09 (1.46)} & \textbf{0.00 (0)} & \textbf{37.92 (8.46)} & \textbf{6.13 (9.61)} & \textbf{10463.84 (18491.25)} & \textbf{0.00 (0)} & \textbf{13.63 (9.63)} \\ \bottomrule
\end{tabular}
}
\newline
\vspace{2mm}
\newline
\resizebox{0.96\textwidth}{!}{
\begin{tabular}{@{}ccccccccc@{}}
\toprule
\toprule
 & \multicolumn{4}{c}{MIK} & \multicolumn{4}{c}{Load Balancing} \\ \midrule
Method & Time (s) & Nodes & PD gap & PD integral & Time (s) & Nodes & PD gap & PD integral \\ \cmidrule(r){1-5}  \cmidrule(l){6-9}
NoCuts & 300.01 (0.009) & 395290.16 (120526.21) & 0.09 (0.018) & 2355.87 (996.08) & 300.00 (0.12) & 108.20 (56.28) & 0.94 (0.12) & 14853.77 (951.42) \\
Default & 179.62 (122.36) & 224746.60 (179450.80) & 0.02 (0.033) & 844.40 (924.30) & 300.00 (0.06) & \textbf{65.22 (55.34)} & 0.43 (0.07) & 9589.19 (1012.95) \\
SBP & 286.07 (41.81) & 417070.36 (133702.49) & 0.07 (0.0295) & 2053.30 (740.11) & 300.00 (0.10) & 138.37 (53.79) & 0.65 (0.06) & 12535.30 (741.43) \\ \midrule
HEM-ratio-order & 282.92 (42.07) & 417397.10 (132077.3) & 0.06 (0.034) & 2072.69 (849.16) & 300.11 (0.11) & 145.18 (56.52) & 0.65 (0.061) & 12368.12 (726.77) \\
HEM-ratio & 283.75 (39.87) & 401540.97 (131295.8) & 0.07 (0.034) & 1869.66 (978.85) & 300.10 (0.098) & 148.92 (59.12) & 0.65 (0.058) & 12410.84 (715.44) \\
HEM (Ours) & \textbf{176.12 (125.18)} & \textbf{220547.93 (172537.94)} & \textbf{0.02 (0.028)} & \textbf{785.04 (790.38)} & \textbf{300.00 (0.04)} & 74.47 (60.98) & \textbf{0.42 (0.07)} & \textbf{9496.42 (1018.35)} \\ \bottomrule
\end{tabular}
}
\newline
\vspace{2mm}
\newline
\resizebox{0.96\textwidth}{!}{
\begin{tabular}{@{}ccccccccc@{}}
\toprule
\toprule
 & \multicolumn{4}{c}{Anonymous} & \multicolumn{4}{c}{MIPLIB mixed   supportcase} \\ \midrule
Method & Time (s) & Nodes & PD gap & PD integral & Time (s) & Nodes & PD gap & PD integral \\ \cmidrule(r){1-5}  \cmidrule(l){6-9}
NoCuts & 246.22 (94.90) & 20110.58 (14167.51) & 8.33e+18 (2.76E+19) & 18297.30 (9769.42) & 170.00 (131.60) & 5971.00 (8679.98) & 9.19 (19.29) & 9927.96 (11334.07) \\
Default & 244.02 (97.72) & 20025.78 (13056.50) & 8.33e+18 (2.76E+19) & 17407.01 (9736.19) & 164.61 (135.82) & \textbf{4055.04 (7404.61)} & 11.73 (31.52) & 9672.34 (10668.24) \\
SBP & 245.71 (92.46) & \textbf{19351.66 (12337.81)} & 6.67e+18 (2.49E+19) & 18188.63 (9651.85) & 165.61 (135.25) & 5618.08 (9765.43) & 0.14 (0.26) & 7408.65 (7903.47) \\ \midrule
HEM-ratio-order & 245.45 (94.99) & 20495.80 (12472.44) & 5.00e+18 (2.18E+19) & 16496.06 (9282.15) & 169.45 (132.55) & 6252.63 (9827.98) & 4.25 (15.97) & 9226.95 (9995.94) \\
HEM-ratio & 245.17 (95.21) & 20942.07 (13379.46) & 5.00e+18 (2.18E+19) & 16148.82 (9247.48) & 163.03 (137.16) & 5551.29 (10708.46) & 7.54 (21.85) & 9979.35 (11048.11) \\
HEM (Ours) & \textbf{241.68 (97.23)} & 20191.28 (13219.21) & \textbf{1.67e+18 (1.28E+19)} & \textbf{16077.15 (9108.21)} & \textbf{162.96 (138.21)} & 4844.87 (9996.15) & \textbf{0.14 (0.25)} & \textbf{6874.80 (6729.97)} \\ \bottomrule
\end{tabular}
}
\end{table*}

\begin{table}[t]
\caption{Sensitivity analysis of HEM to the hyperparameter dealy update freq $d$.}
\vspace{-2mm}
\label{ablation_sensitivity}
\resizebox{0.96\textwidth}{!}{
\begin{tabular}{@{}cccccccccc@{}}
\toprule
\toprule
 & \multicolumn{3}{c}{Maximum Independent Set} & \multicolumn{3}{c}{Corlat} & \multicolumn{3}{c}{MIPLIBS Mixed (neos)} \\ \midrule
\multirow{2}{*}{Method} & \multirow{2}{*}{Time(s)} & \multirow{2}{*}{PD integral} & \multirow{2}{*}{\begin{tabular}[c]{@{}c@{}}Improvement\\      (PD integral, \%)\end{tabular}} & \multirow{2}{*}{Time(s)} & \multirow{2}{*}{PD integral} & \multirow{2}{*}{\begin{tabular}[c]{@{}c@{}}Improvement\\      (PD integral, \%)\end{tabular}} & \multirow{2}{*}{Time(s)} & \multirow{2}{*}{PD integral} & \multirow{2}{*}{\begin{tabular}[c]{@{}c@{}}Improvement\\      (PD integral, \%)\end{tabular}} \\
 &  &  &  &  &  &  &  &  &  \\ \cmidrule(r){1-4} \cmidrule(lr){5-7} \cmidrule(l){8-10}
NoCuts & 8.78 (6.66) & 71.31 (51.74) & NA & 103.30 (128.14) & 2818.40 (5908.31) & NA & 253.65 (80.29) & 14652.29 (12523.37) & NA \\
HEM (delay=1) & 1.79 (3.65) & 16.56 (26.6) & 76.78 & 57.39 (111.76) & 1260.83 (3518.07) & 55.26 & 254.48 (78.84) & 9273.01 (12031.44) & 36.71 \\
HEM (delay=2) & 1.76 (3.69) & 16.01 (26.21) & 77.55 & 58.31 (110.51) & 1079.99 (2653.14) & 61.68 & 248.66 (89.46) & 8678.76 (12337.00) & 40.77 \\
HEM (delay=3) & 1.80 (3.82) & 16.91 (28.49) & 76.29 & 58.50 (108.88) & 1003.52 (2264.91) & 64.39 & 247.62 (90.73) & 8408.16 (12467.02) & 42.62 \\
HEM (delay=4) & 1.89 (3.90) & 17.34 (28.46) & 75.68 & 69.60 (115.42) & 1309.33 (3336.59) & 53.54 & 246.93 (91.93) & 8368.46 (12489.68) & 42.89 \\ \bottomrule
\end{tabular}
}
\end{table}

\begin{table}[t]
\centering
\caption{Evaluate the generalization ability of HEM on Set Covering.}
\vspace{-2mm}
\label{generalization_setcovering}
\resizebox{0.8\textwidth}{!}{
\begin{tabular}{@{}ccccccc@{}}
\toprule
 & \multicolumn{3}{c}{Set Covering ($2 \times$)} & \multicolumn{3}{c}{Set Covering ($4 \times$)} \\ \midrule
\multirow{2}{*}{Method} & \multirow{2}{*}{Time(s) $\downarrow$} & \multirow{2}{*}{\begin{tabular}[c]{@{}c@{}}Improvement $\uparrow$\\      (time, \%)\end{tabular}} & \multirow{2}{*}{PD integral $\downarrow$} & \multirow{2}{*}{Time(s) $\downarrow$} & \multirow{2}{*}{\begin{tabular}[c]{@{}c@{}}Improvement $\uparrow$ \\      (time, \%)\end{tabular}} & \multirow{2}{*}{PD integral $\downarrow$} \\
 &  &  &  &  &  &  \\ \cmidrule(r){1-4} \cmidrule(l){5-7}
NoCuts & 82.69 (78.27) & NA & 609.43 (524.92) & 284.44 (48.70) & NA & 3215.34 (1019.47) \\
Default & 61.01 (78.12) & 26.22 & 494.63 (545.76) & 149.69 ( 141.92) & 47.37 & 1776.22 (1651.15) \\
Random & 64.44 (73.98) & 22.07 & 520.84 (489.52) & 208.12 (131.52) & 26.53 & 2528.36 (1678.66) \\
NV & 92.05 (80.11) & -11.32 & 725.53 (541.68) & 286.10 (45.47) & -0.58 & 3422.46 (1024.19) \\
Eff & 92.32 (79.33) & -11.64 & 733.72 (538.60) & 286.20 (45.04) & -0.62 & 3437.06 (1043.44) \\\midrule
SBP & 3.52 (1.36) & 95.74 & 92.89 (25.83) & 7.62 (6.46) & 97.32 & 256.79 (145.92) \\
HEM (Ours) & \textbf{3.33 (0.47)} & \textbf{95.97} & \textbf{89.24 (14.26)} & \textbf{7.40 (5.03)} & \textbf{97.40} & \textbf{250.83 (131.43)} \\ \bottomrule
\end{tabular}
}
\end{table}

\subsubsection{The importance of tackling \textbf{P1-P3} in cut selection}
    To understand the importance of tackling \textbf{P1-P3} in cut selection, we provide more results of HEM, HEM-ratio, and HEM-ratio-order on Set Covering, Multiple Knapsack, MIK, Load Balancing, Anonymous, and MIPLIB mixed supportcase. Here we refresh what HEM, HEM-ratio, HEM-ratio-order mean. (1) \textbf{HEM}. HEM tackles \textbf{P1-P3} in cut selection simultaneously. (2) \textbf{HEM-ratio}. In order not to learn how many cuts should be selected, we remove the higher-level model of HEM and force the lower-level model to select a fixed ratio of cuts. We denote it by HEM-ratio. Note that HEM-ratio is different from HEM w/o H (see Appendix \ref{appendix_implementation}). HEM-ratio tackles \textbf{P1} and \textbf{P3} in cut selection. (3) \textbf{HEM-ratio-order.} To further mute the effect of the order of selected cuts, we reorder the selected cuts given by HEM-ratio with the original index of the generated cuts, which we denote by HEM-ratio-order. HEM-ratio-order mainly tackles \textbf{P1}.
    
    The results in Table \ref{ablation_factor_more_datasets} suggest the following. HEM-ratio-order outperforms Default and NoCuts on several datasets, demonstrating that tackling \textbf{P1} by data-driven methods is crucial. HEM significantly outperforms HEM-ratio in terms of the primal-dual gap integral, demonstrating the significance of tackling \textbf{P2}. HEM-ratio outperforms HEM-ratio-order  on several datasets, which demonstrates the importance of tackling \textbf{P3}. Moreover, HEM-ratio performs better than SBP in terms of the solving time and the primal-dual gap integral on all six datasets except Set Covering and MIPLIB mixed supportcase, which shows the superiority of formulating the cut selection as a sequence to sequence learning problem over formulating it as a scoring task. However, HEM-ratio and HEM-ratio-order perform a little worse than SBP on Set Covering and MIPLIB mixed supportcase. A possible reason is that it is more difficult to train a sequence model than to train a multi-layer perceptron and thus the sequence model may suffer from inefficient exploration and be trapped to the local optimum. Please refer to Appendix \ref{appendix_analysis_hem_ratio} for a detailed analysis.  

\subsubsection{Sensitivity analysis}\label{appendix_delay}
    Additionally, we perform ablation studies to test the sensitivity of HEM to the hyperparameter delay update freq $d$. The results in Table \ref{ablation_sensitivity} show that there is a wide range of $d$ for HEM to achieve comparable performance on Maximum Independent Set, CORLAT, and MIPLIB mixed neos. Moreover, we emphasize that we do not tune the hyperparameter $d$. As the results shown in Table \ref{ablation_sensitivity}, $d=3$ and $d=4$ performs the best on CORLAT and MIPLIB mixed neos, respectively. However, we simply set $d=2$ throughout all experiments in the main text.

\subsection{More results of generalization}\label{appendix_gener}
        Here we provide more results of the generalization experiments on Set Covering. 
        On Set Covering, we test HEM on two times and four times larger instances than those seen during training. 
        The results in Table \ref{generalization_setcovering} show that HEM generalizes well 
        to instances that are significantly larger than seen during training. 
        In particular, HEM achieves at least $70\%$ improvement in terms of the Time compared to all the rule-based baselines. Moreover, SBP also generalizes well to large instances, demonstrating that SBP is a strong baseline.

    \begin{figure}[t]
        \centering
        \includegraphics[width=0.4\textwidth]{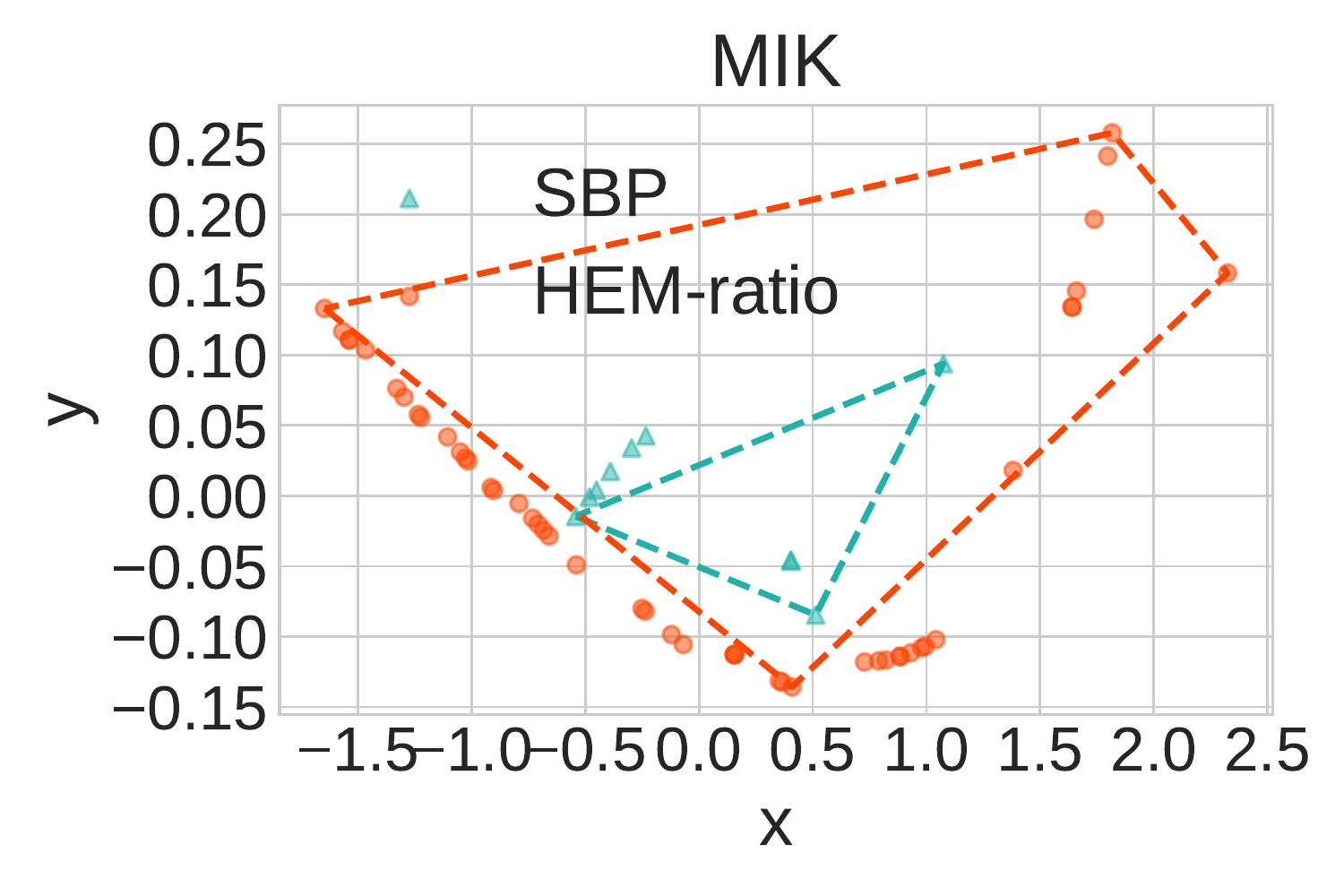}
        \includegraphics[width=0.4\textwidth]{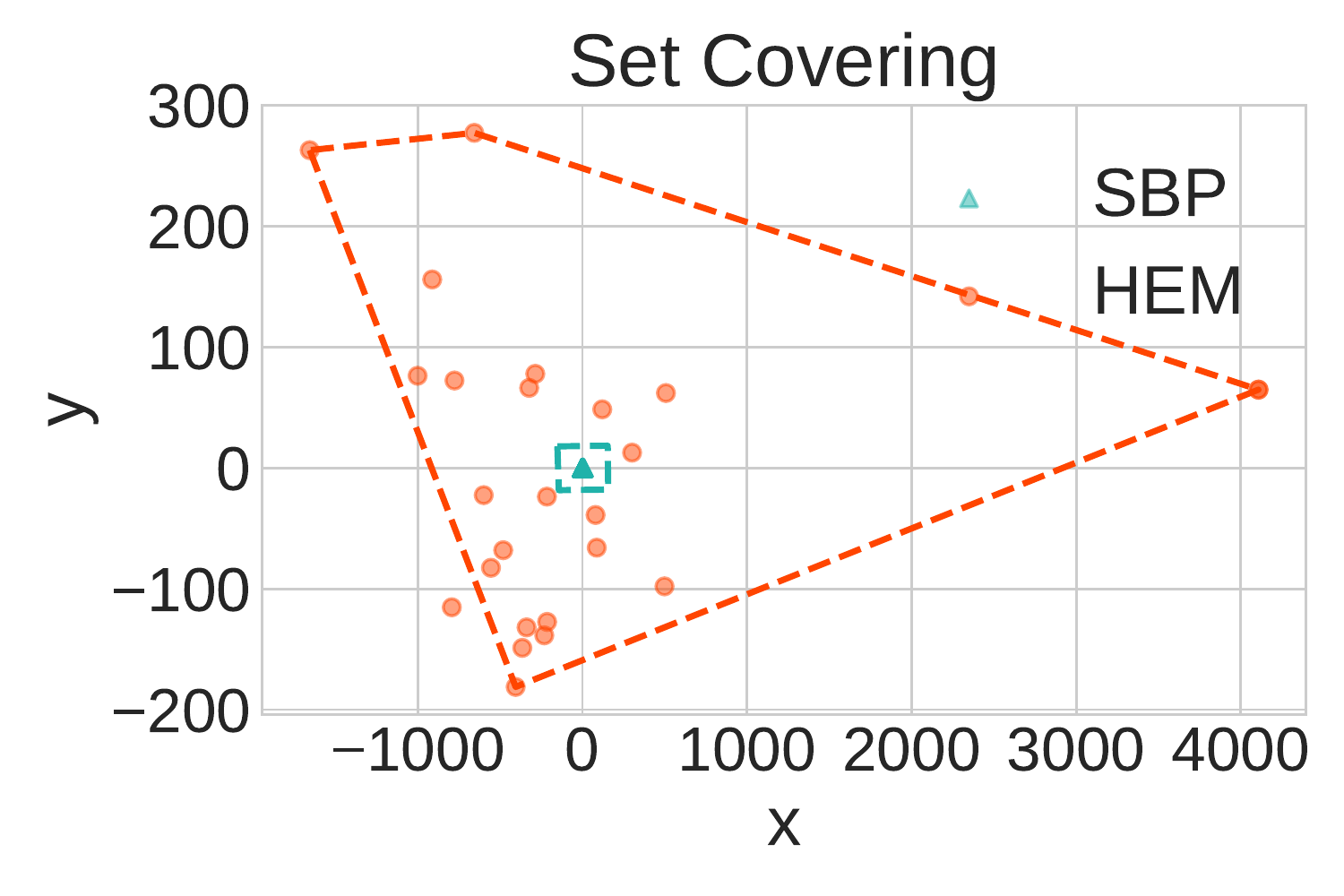}
        \vspace{-3mm}
        \caption{We perform principal component analysis on cuts selected by HEM-ratio/HEM and SBP. Each colored point illustrates a reduced cut feature. To visualize the diversity of selected cuts, we use dashed lines to connect the points with the smallest and largest x,y coordinates.}
        \label{fig:visualization_more_results}
    \end{figure}  
    
\subsection{More visualization results}\label{appendix_results_visualization}
    In this part, we provide more visualization results on Set Covering and MIK. On MIK, we visualize the cuts selected by HEM-ratio and SBP on a randomly sampled instance. We perform principal component analysis \citep{machine_learning} on selected cuts to reduce the cut features to two-dimensional space. Colored points illustrate reduced cut features. 
    To visualize the diversity of selected cuts, we use dashed lines to connect the points with the smallest and largest x,y coordinates.
    The results in Figure \ref{fig:visualization_more_results} still show that HEM-ratio selects much more diverse cuts than SBP on MIK.
    However, HEM-ratio performs poorer than SBP on Set Covering (see Appendix \ref{appendix_analysis_hem_ratio} for a detailed analysis). Therefore, we visualize the cuts selected by HEM and SBP on a randomly sampled instance from Set Covering. Although HEM learns the number of cuts that should be selected, we find that HEM selects much fewer cuts than SBP. Specifically, HEM selects 25 cuts, while SBP selects 158 cuts. Interestingly, the results in Figure \ref{fig:visualization_more_results} show that SBP selects 158 similar cuts with high scores, while HEM selects much more diverse cuts than SBP. The results show that HEM tends to select cuts that complement each other nicely.

\begin{table}[t]
\caption{Policy evaluation on MIPLIB mixed neos and MIPLIB mixed supportcase with a time limit of \textit{three hours}.}
\vspace{-2mm}
\label{evaluation_3h}
\centering
\resizebox{0.96\textwidth}{!}{
\begin{tabular}{@{}ccccccccccc@{}}
\toprule
\toprule
 & \multicolumn{5}{c}{MIPLIB mixed neos} & \multicolumn{5}{c}{MIPLIB mixed   supportcase} \\ \midrule
Method & Time (s) & Nodes & PD gap & PD integral & \begin{tabular}[c]{@{}c@{}}Improvement\\      (PD integral, \%)\end{tabular} & Time (s) & Nodes & PD gap & PD integral & \begin{tabular}[c]{@{}c@{}}Improvement\\      (PD integral, \%)\end{tabular} \\ \cmidrule(r){1-6} \cmidrule(l){7-11} 
NoCuts & 8,126.15 (4631.32) & 5,058,282.42 (5067358) & 0.97 (1.14) & 369,180.71 (368274) & NA & 3,616.09 (4703.68) & 258,543.71 (692821) & 0.00 (0.0096) & 37,858.65 (59578.6) & NA \\
Default & 8,129.27 (4625.92) & 4,800,372.67 (4815222) & 0.94 (1.07) & 392,324.09 (393266) & -6.27 & 3,997.69 (4720.2) & 188,844.71 (599665) & 0.01 (0.012) & 56,375.64 (106694) & -48.91 \\
Random & 8,125.44 (4632.53) & 5,107,627.25 (5114152) & 0.95 (1.09) & 395,694.74 (393445) & -7.18 & 2,605.78 (3993.84) & \textbf{8,458.96 (14785.9)} & 0.01 (0.029) & 44,222.39 (70937.6) & -16.81 \\
NV & 8,133.64 (4618.37) & 4,930,498.92 (4930971) & 0.91 (1.03) & 326,315.04 (391163) & 11.61 & \textbf{2,561.10 (3980.44)} & 34,851.08 (142069) & 0.00 (0.0095) & 27,908.00 (37742.2) & 26.28 \\
Eff & 8,130.78 (4623.31) & 4,788,568.75 (4804443) & 0.94 (1.07) & 395,640.44 (393572) & -7.17 & 3,105.98 (4308.57) & 182,187.63 (543167) & 0.01 (0.028) & 39,932.37 (58307.3) & -5.48 \\ \midrule
SBP & 8,130.63 (4626.62) & 4,763,378.00 (4829037) & 0.91 (1.03) & 388,564.35 (398176.3) & -5.25 & 3,014.63 (4082.19) & 146,180.96 (543662.45) & 0.00 (0.0084) & 24,447.37 (55984.9) & 35.42 \\
HEM(Ours) & \textbf{8,124.32 (4634.54)} & \textbf{4,748,599.50 (4816327)} & \textbf{0.78 (0.82)} & \textbf{194,557.87 (345035)} & \textbf{47.30} & 3,465.48 (4485.11) & 52,750.08 (140410) & \textbf{0.00 (0.0087)} & \textbf{17,885.98 (24045.9)} & \textbf{52.76} \\ \bottomrule
\end{tabular}
}
\end{table}

\begin{figure*}
    \centering
    \includegraphics[width=0.3\textwidth]{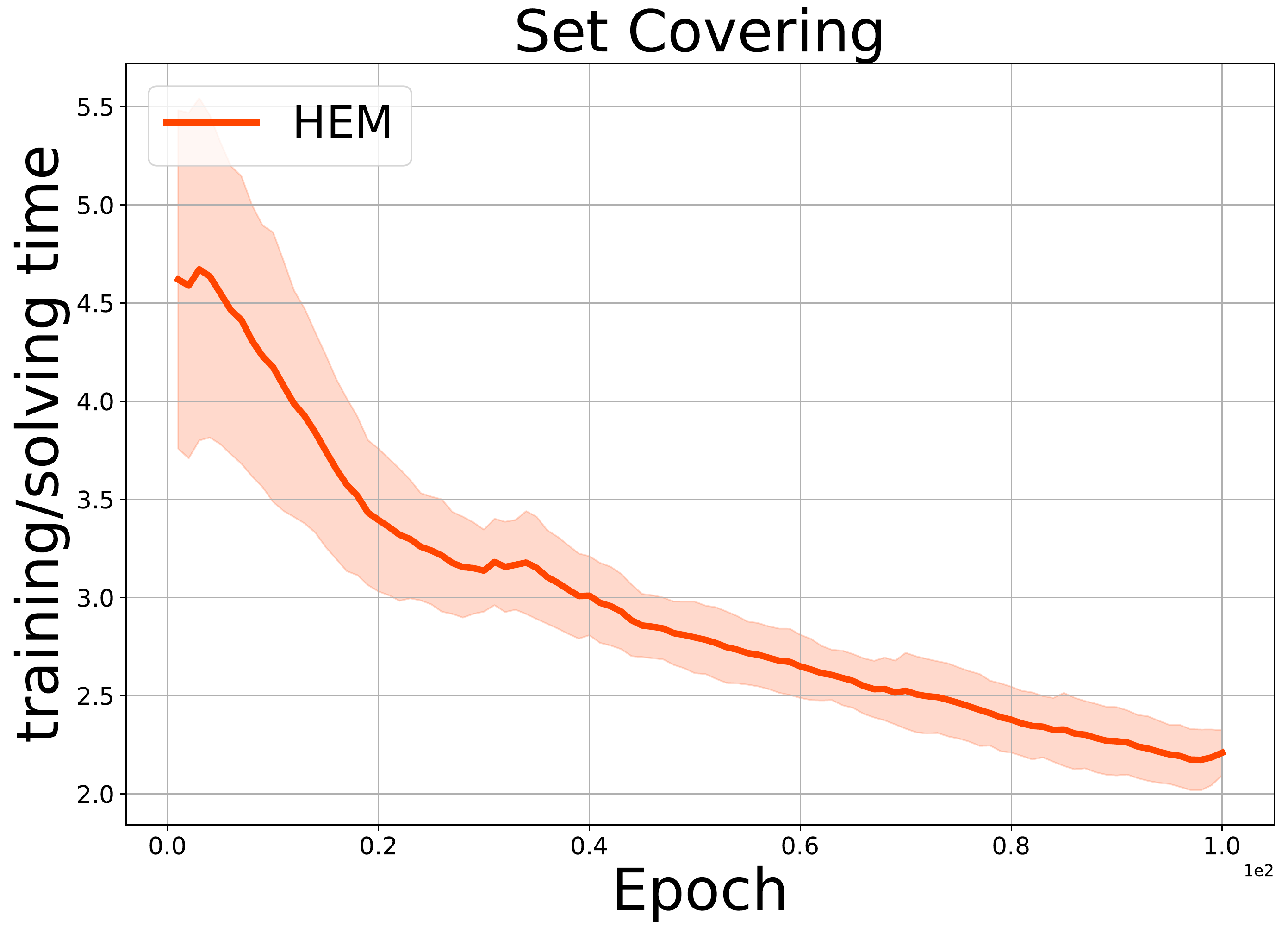}
        \includegraphics[width=0.3\textwidth]{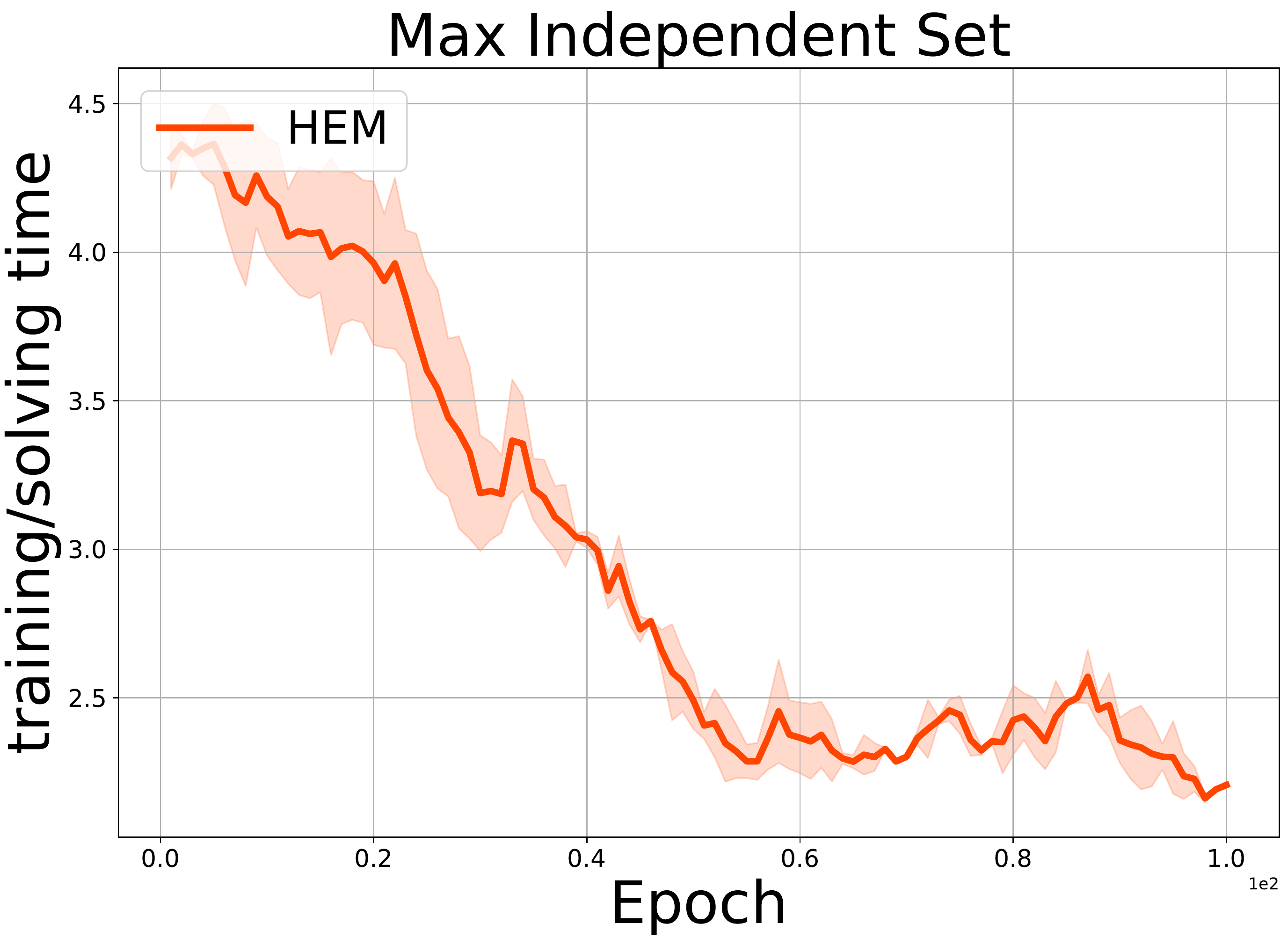}
        \includegraphics[width=0.3\textwidth]{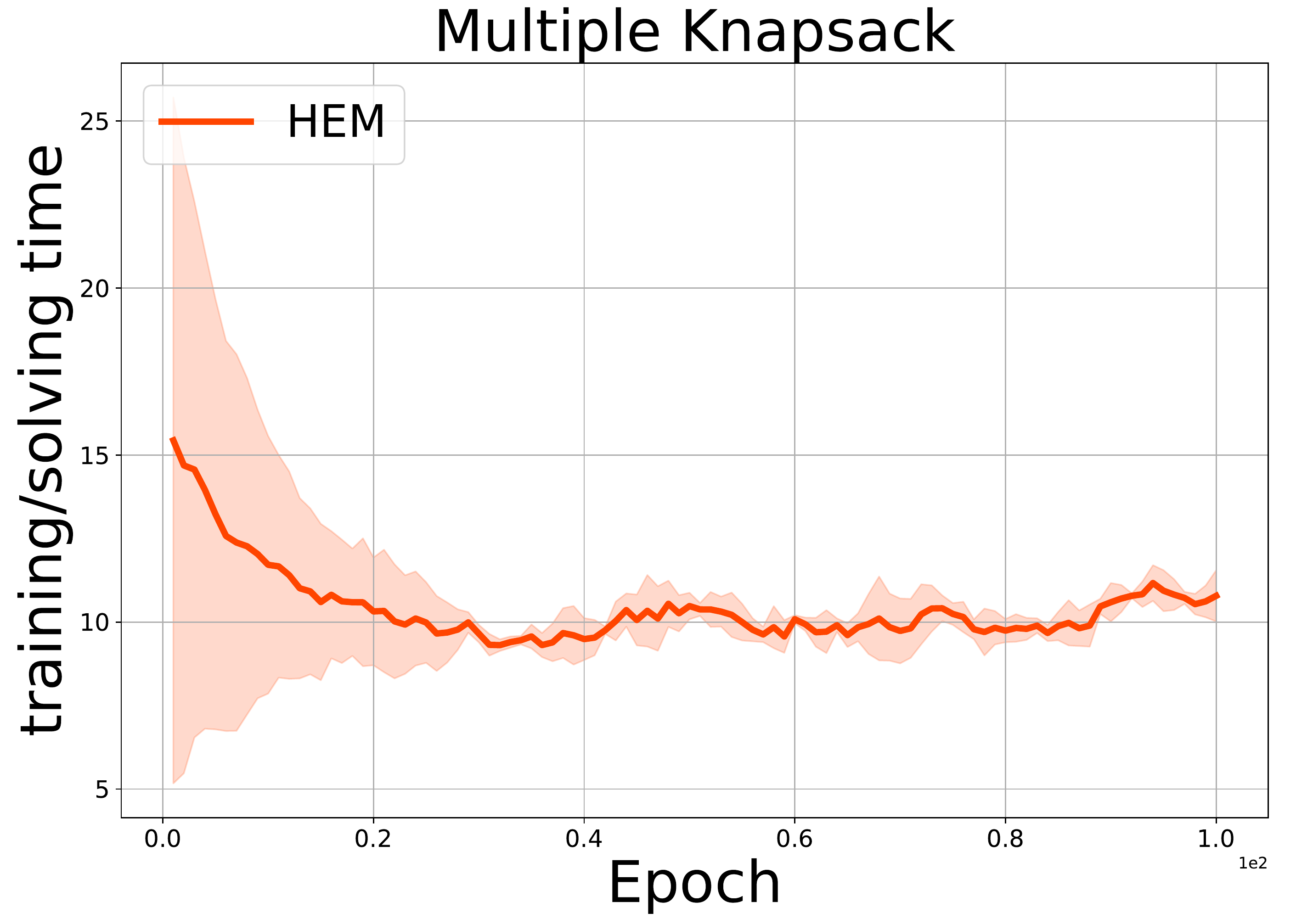}
        \includegraphics[width=0.3\textwidth]{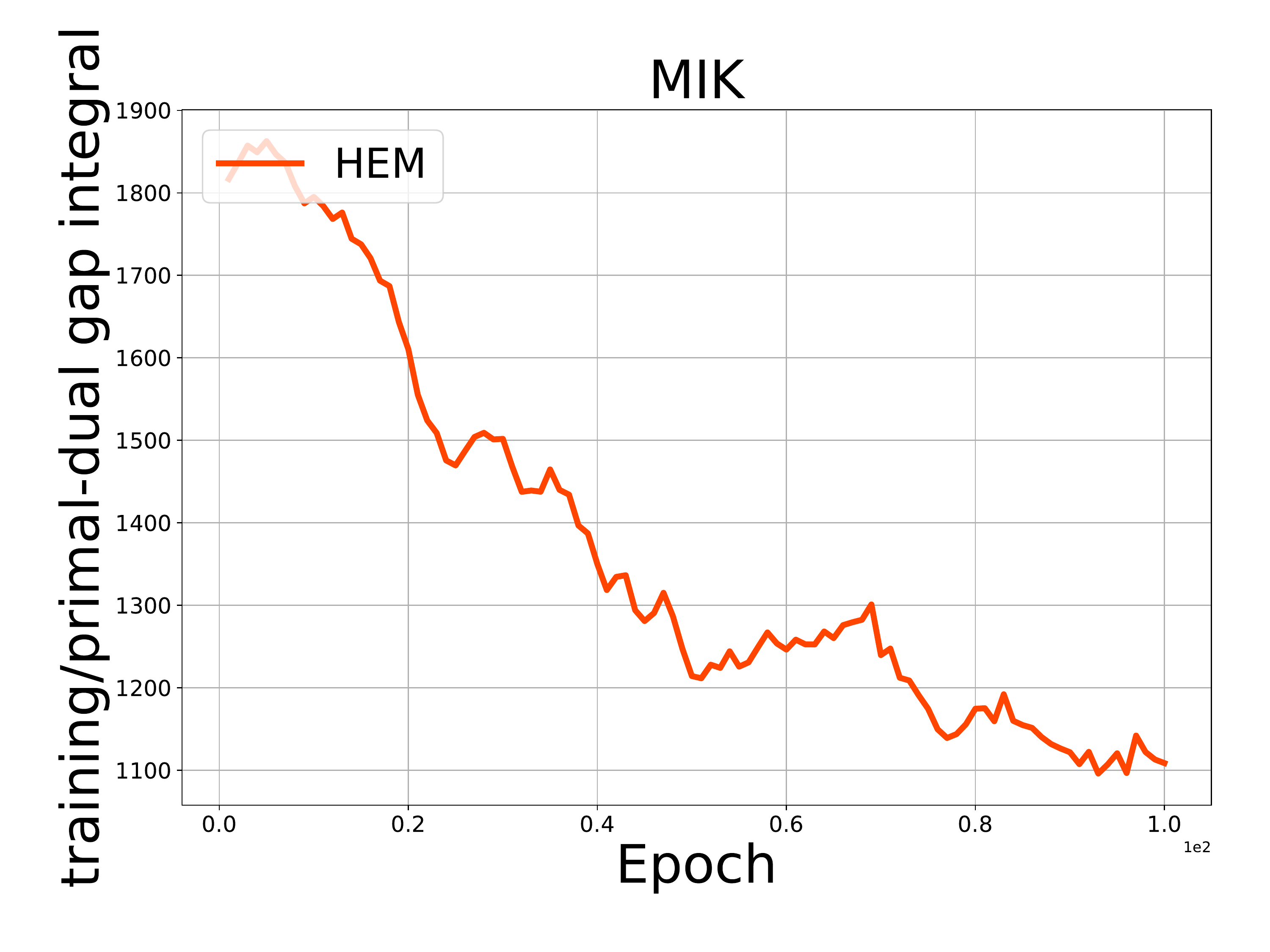}
        \includegraphics[width=0.3\textwidth]{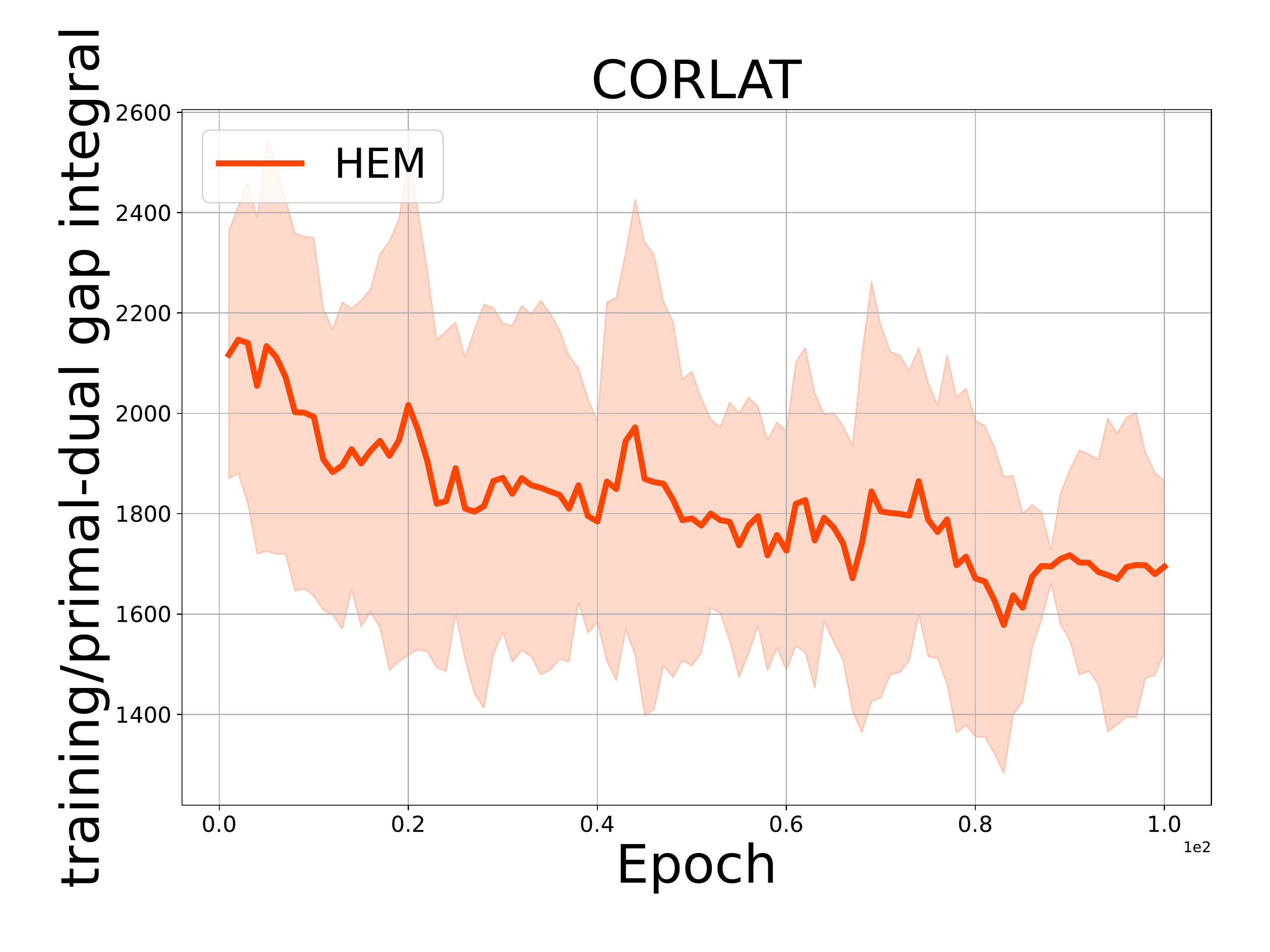}
        \includegraphics[width=0.3\textwidth]{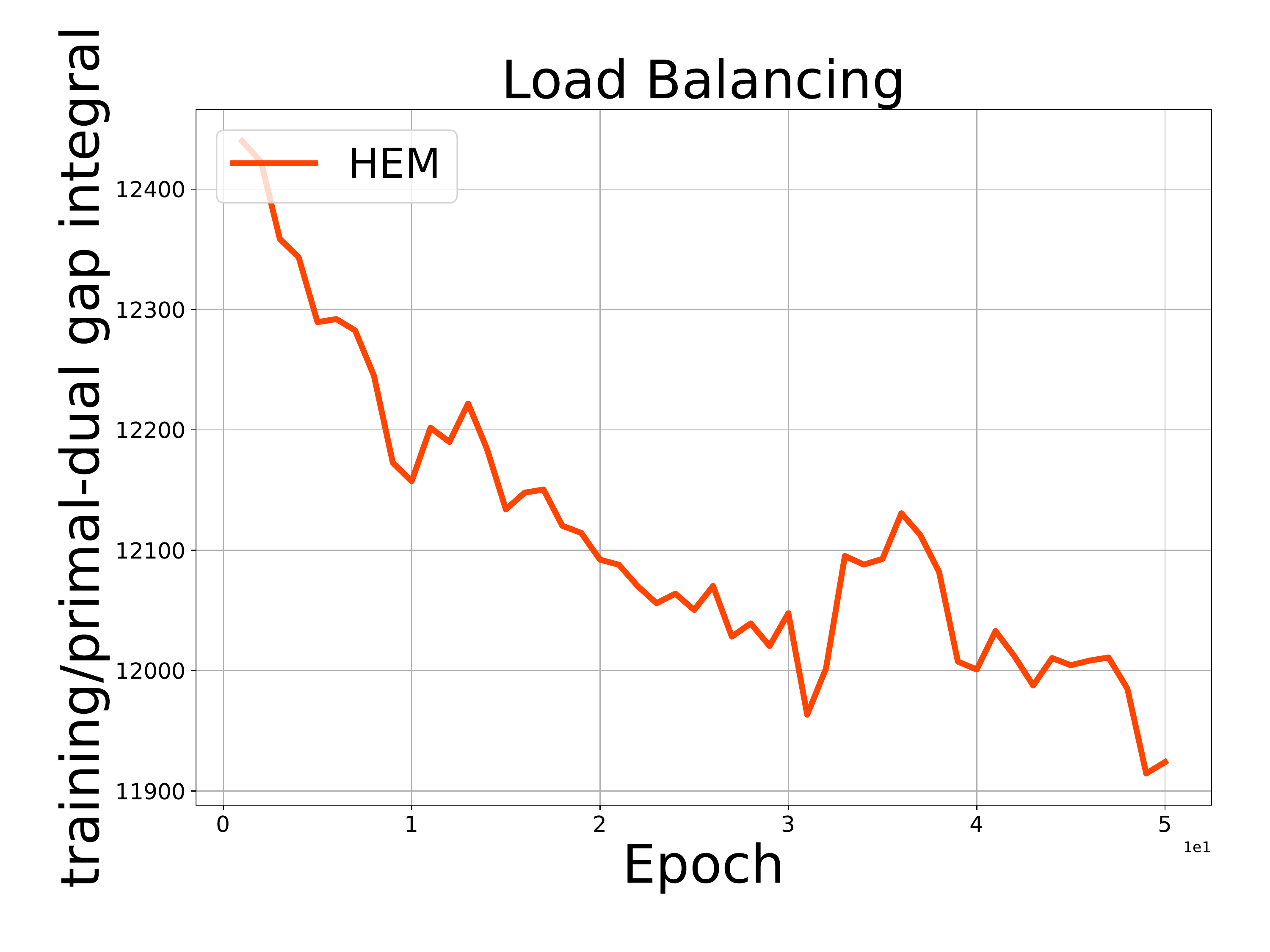}
        \includegraphics[width=0.3\textwidth]{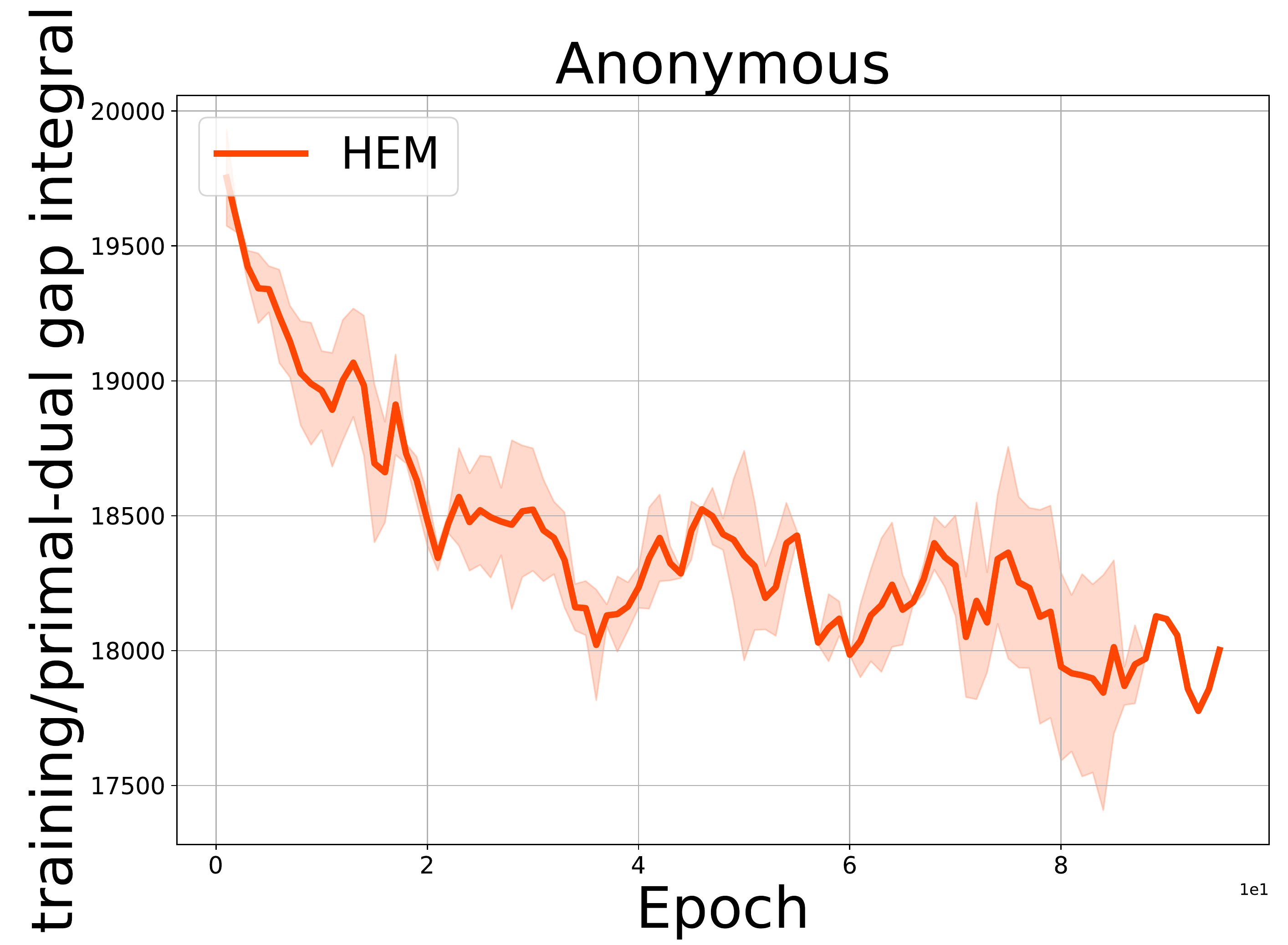}
        \includegraphics[width=0.3\textwidth]{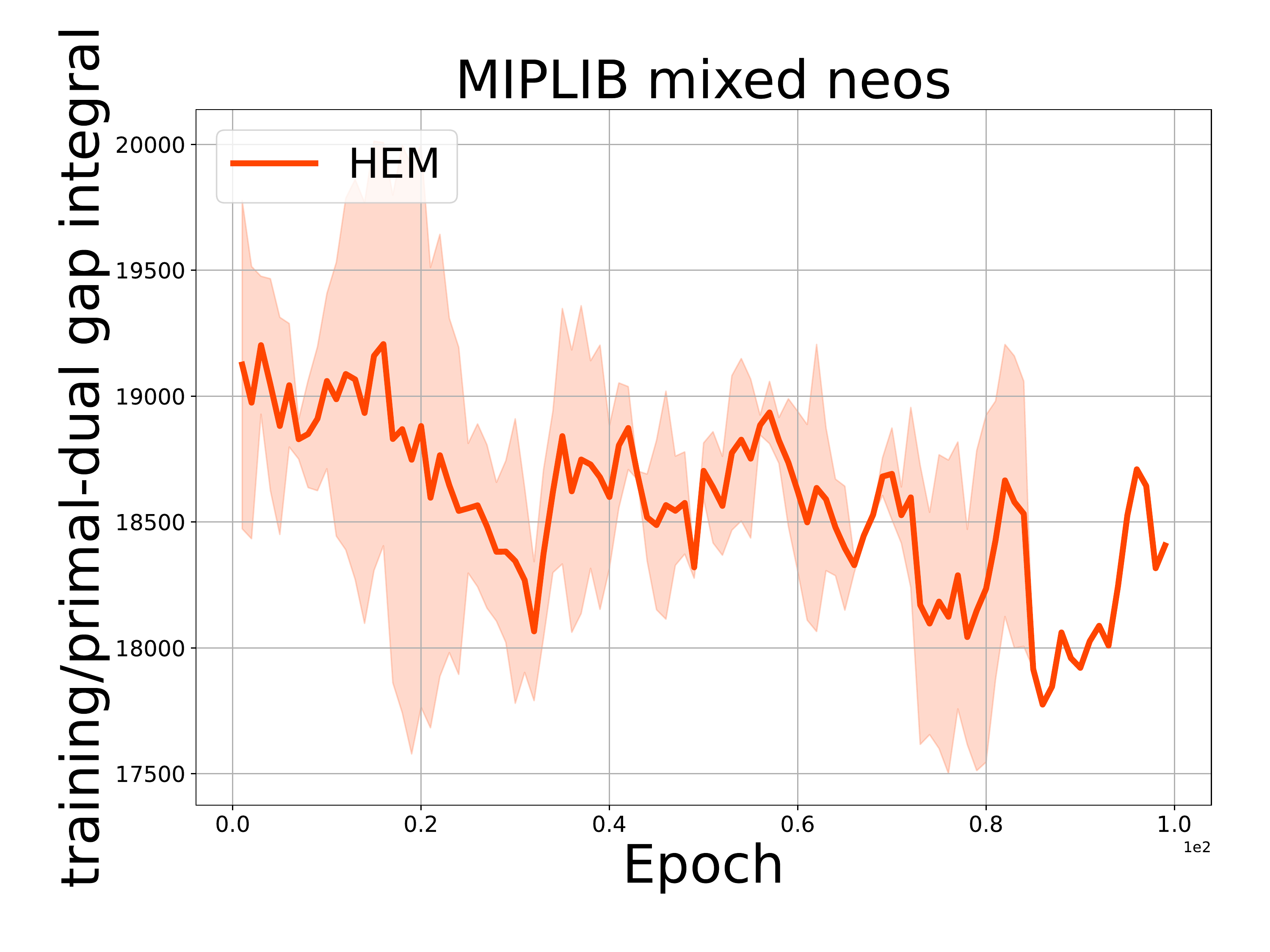}
        \includegraphics[width=0.3\textwidth]{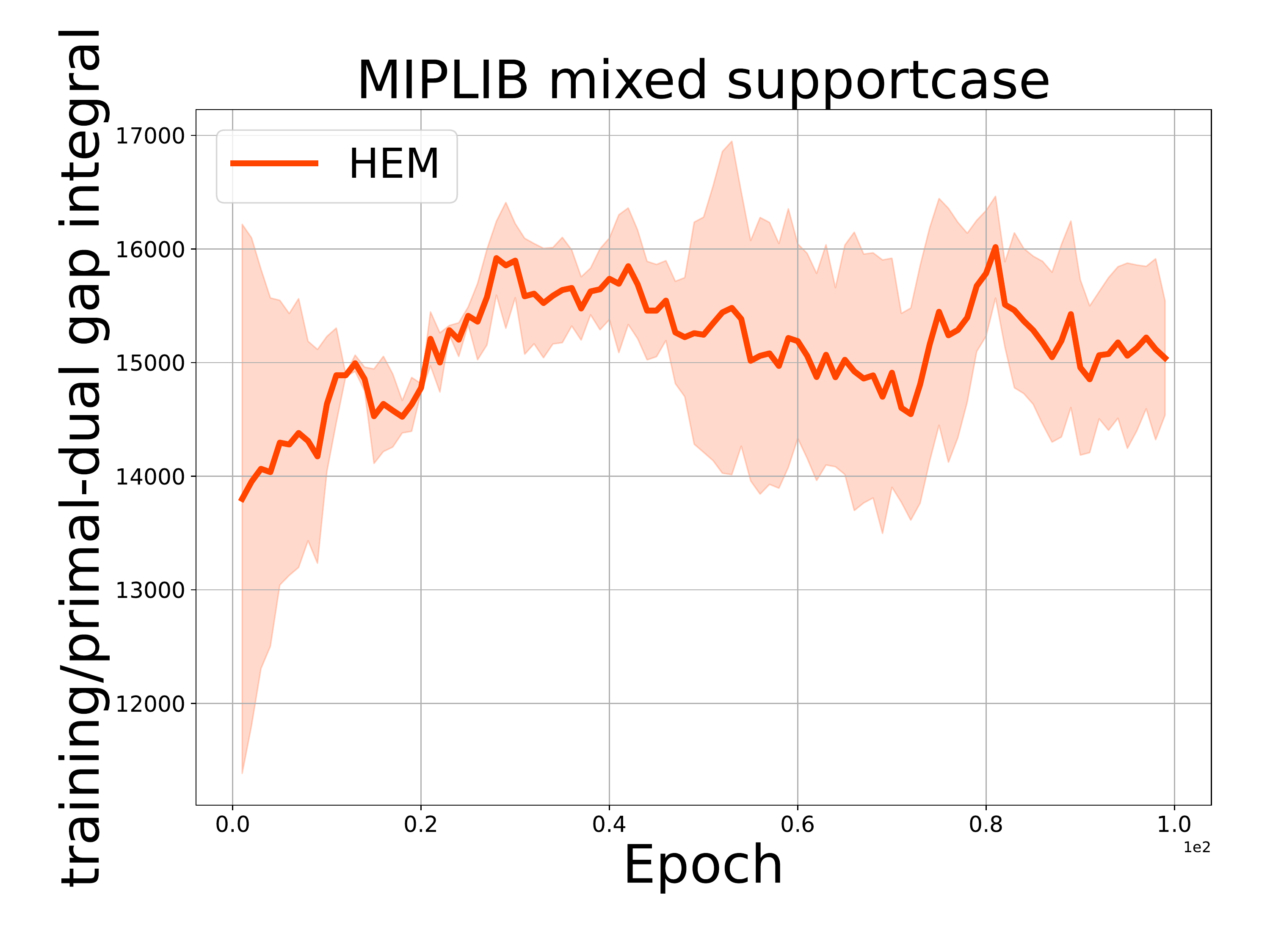}
    \caption{Training curves of HEM on all nine datasets. The x-axis corresponds to the training epochs. The y-axis corresponds to the average solving time on easy datasets and the primal-dual gap integral on the other datasets. The solid curves correspond to the mean and the shaded region to the standard deviation over three random seeds. For visual clarity, we smooth curves.}
    \label{fig:training_curves}
\end{figure*}

\subsection{Evaluation with a time limit of three hours}\label{appendix_results_long_test_time}
    In this section, we aim to evaluate whether HEM can generalize well to solving problems within a much longer time limit. Specifically, we evaluate HEM on two extremely challenging MIPLIB datasets within a time limit of \textit{three hours}. Note that we still train HEM with a time limit of 300 seconds, while we test HEM with a time limit of three hours.
    The results in Table \ref{evaluation_3h} show that HEM still significantly outperforms all the baselines, especially in terms of the primal-dual gap integral on MIPLIB mixed neos and MIPLIB mixed supportcase. In terms of the primal-dual gap, HEM also outperforms the baselines. Moreover, HEM performs better than baselines in terms of the solving time on MIPLIB mixed neos, but HEM performs poorly in terms of the solving time on MIPLIB mixed supportcase. Interestingly, the primal-dual gap integral is not always consistent with the solving time. We emphasize that we train with the negative primal-dual gap integral reward. To further improve the performance of HEM in terms of the solving time, we can set the reward as the negative solving time instead of the negative primal-dual gap integral. 

\subsection{Training curves}
    In this section, we provide the training curves of HEM on all nine datasets. The results in Figure \ref{fig:training_curves} show that the performance of our learned policies in terms of the solving time or the primal-dual gap integral drops with the training epochs, demonstrating the effectiveness of our learning process.

\end{document}